\documentclass[a4paper]{article}[11pts]

\usepackage[english]{babel}
\usepackage[utf8x]{inputenc}
\usepackage[T1]{fontenc}

\normalsize

\usepackage[a4paper,top=1in,bottom=1in,left=1in,right=1in,marginparwidth=1in]{geometry}
\usepackage{setspace}

\usepackage{appendix}
\usepackage{color}
\definecolor{strcolor}{rgb}{0.6, 0.2, 0.6}
\definecolor{commentcolor}{rgb}{0.3125, 0.5, 0.3125}
\definecolor{keycol}{rgb}{0, 0, 1}

\usepackage{amssymb}
\usepackage{amsmath}
\usepackage{amsthm}
\usepackage{amsfonts, mathtools}
\usepackage{bbm}
\usepackage{tablefootnote}
\usepackage{booktabs}
\usepackage{multirow}

\DeclareMathOperator*{\argmax}{arg\,max}

\newtheorem{lemma}{Lemma}

\newtheorem{theorem}{Theorem}
\newtheorem{corollary}{Corollary}
\newtheorem{definition}{Definition}

\newcommand{\train}{\text{train}}
\newcommand{\test}{\text{test}}
\newcommand{\trans}{\text{trans}}

\newcommand{\tb}[1]{\textbf{#1}}
\newcommand{\1}{\mathbbm{1}}
\newcommand{\mP}{\mathbb{P}}

\newcommand{\bs}[1]{\boldsymbol{#1}}

\usepackage{listings}
\usepackage{url}
\usepackage{hyperref}
\hypersetup{
    hidelinks,
    colorlinks=true,
    linkcolor=red,
    citecolor=magenta
} % delete the box around the link

\newcommand {\bea}{\begin{eqnarray}}
	\newcommand {\eea}{\end{eqnarray}}

\newtheorem{algorithm}{Algorithm}
\newtheorem{remark}{Remark}

\def\blot{\quad \mbox{$\vcenter{ \vbox{ \hrule height.4pt
				\hbox{\vrule width.4pt height.9ex \kern.9ex \vrule width.4pt}
				\hrule height.4pt}}$}}

\usepackage{natbib}
 \bibpunct[, ]{(}{)}{,}{a}{}{,}%
\usepackage{xspace}

\usepackage{subfig}
\usepackage{adjustbox}
\usepackage{algorithm}
\usepackage{algpseudocode}
\usepackage{tikz}
\usepackage{pgfplots}
\pgfplotsset{compat=1.5}
\usepackage{multirow}
\usepgfplotslibrary{fillbetween, groupplots}
\usepackage{makecell}

\makeatletter
\newenvironment{breakablealgorithm}
  {% \begin{breakablealgorithm}
   \begin{center}
     \refstepcounter{algorithm}% New algorithm
     \hrule height.8pt depth0pt \kern2pt% \@fs@pre for \@fs@ruled
     \renewcommand{\caption}[2][\relax]{% Make a new \caption
       {\raggedright\textbf{\fname@algorithm~\thealgorithm} ##2\par}%
       \ifx\relax##1\relax % #1 is \relax
         \addcontentsline{loa}{algorithm}{\protect\numberline{\thealgorithm}##2}%
       \else % #1 is not \relax
         \addcontentsline{loa}{algorithm}{\protect\numberline{\thealgorithm}##1}%
       \fi
       \kern2pt\hrule\kern2pt
     }
  }{% \end{breakablealgorithm}
     \kern2pt\hrule\relax% \@fs@post for \@fs@ruled
   \end{center}
  }
\makeatother

\pgfplotsset{
  every nth point/.style={
    x filter/.code={
      \pgfmathtruncatemacro{\keep}{mod(\coordindex,#1)}
      \ifnum\keep=0
      \else
        \def\pgfmathresult{nan}
      \fi
    }
  }
}

\begin{document}
	%%%%%%%%%%%%%%%%

	\title{From Small to Large: A Graph Convolutional Network Approach for Solving Assortment Optimization Problems}

	\author{Guokai Li$^1$ \and Pin Gao$^2$ \and Stefanus Jasin$^3$ \and Zizhuo Wang$^2$} %
    \date{\small
        1. Smith School of Business, Queen's University, Kingston, Ontario, Canada\\
        2. School of Data Science, The Chinese University of Hong Kong, Shenzhen, Guangdong, China\\
        3. Stephen M. Ross School of Business, University of Michigan, Ann Arbor, MI, USA\\
        guokai.li@queensu.ca, gaopin@cuhk.edu.cn, sjasin@umich.edu, wangzizhuo@cuhk.edu.cn}
     \maketitle

     \onehalfspacing

    \begin{abstract}
Assortment optimization seeks to select a subset of substitutable products, 
subject to constraints, to maximize expected revenue.  
The problem is NP-hard due to its combinatorial and nonlinear nature and arises 
frequently in industries such as e-commerce, where platforms must solve 
thousands of such problems each minute.  
We propose a \emph{graph convolutional network} (GCN) framework to efficiently 
solve constrained assortment optimization problems.  
Our approach constructs a graph representation of the problem, trains a GCN to 
learn the mapping from problem parameters to optimal assortments, and develops 
three inference policies based on the GCN's output.  
Owing to the GCN's ability to generalize across instance sizes, patterns learned 
from small-scale samples can be transferred to large-scale problems. 
Theoretical results are established to show the expressive power of the proposed GCN, and explain the underlying mechanism of the size generalization ability. 
Numerical experiments show that a GCN trained on instances with 20~products 
achieves over 85\% of the optimal revenue on problems with up to 
2{,}000~products within seconds, outperforming existing heuristics in both 
accuracy and efficiency.  
We further extend the framework to settings with an unknown choice model using 
transaction data and demonstrate similar performance and scalability.  

\end{abstract}

\section{Introduction}

Assortment optimization is a fundamental problem in revenue management with broad applications across various industries, including airline, hospitality, retailing, and online advertisement. The objective is to select a subset of substitutable products in order to maximize the expected revenue. In practice, there are often constraints on the set of products that can be selected. For example, given a query keyword, an e-commerce platform needs to choose a subset of substitutable products to display on the user interface with limited space. In some other cases, a product can be displayed only if another product is displayed.  The assortment optimization problem is usually computationally challenging due to its combinatorial and non-linear nature. For instance, \cite{rusmevichientong2014assortment} show that the unconstrained assortment problem under the mixed multinomial logit model is NP-hard even when there are only two customer types, and \cite{desir2022capacitated} demonstrate that the assortment problem under the multinomial logit model is NP-hard when there exists a capacity constraint.  However, real-world applications often require frequently solving (personalized) assortment problems in response to customer visits. For instance, in order to determine personalized product recommendations for each customer visit, Alibaba needs to solve more than 25 million constrained assortment optimization problems per week (\citealt{feldman2022customer}), implying that each assortment problem needs to be solved in seconds. Therefore, developing fast and reliable algorithms for assortment optimization is of paramount importance for practical deployment, e.g., deriving high-quality solutions within seconds. 

In recent years, neural networks have demonstrated remarkable success across a 
variety of combinatorial optimization problems, including scheduling 
(\citealt{khalil2016learning}), vehicle routing 
(\citealt{nazari2018reinforcement}), and matching 
(\citealt{li2019graph}).  
Motivated by these advances, we investigate how neural networks can be 
leveraged to efficiently solve \emph{constrained assortment optimization} 
problems.  
We focus on practical settings in which a platform repeatedly solves numerous 
assortment problems with similar parameter structures—for example, instances 
drawn from a common distribution or within specific parameter ranges.  
In such cases, a latent mapping exists between problem parameters and optimal 
assortments, and solving each instance from scratch becomes computationally 
inefficient.  
The key idea of our approach is to use neural networks to learn this mapping 
from data and to design efficient, high-performing policies that exploit the 
network's output.  

Specifically, we aim to address the following research questions:
\begin{enumerate}
    \item \textit{Given knowledge of the underlying choice model, how can we 
    efficiently derive high-quality solutions using neural networks?}
    \item \textit{How does the performance of the neural-network-based approach 
    scale with problem size?}
    \item \textit{When the choice model is unknown, how can we adapt the policy 
    to leverage transaction data for effective assortment optimization?}
\end{enumerate}

To address the first research question, we consider the case where the 
underlying choice model and its parameters are known.  
We focus on the mixed multinomial logit (MMNL) model as the representative 
choice model for demonstrating our results.  
This model is selected because it can approximate a broad class of random 
utility models with arbitrary accuracy under mild regularity conditions 
(\citealt{mcfadden2000mixed}).  In this setting, our goal is to train a neural network to learn the structural 
patterns of the optimal assortment and subsequently design inference policies 
based on the trained model.  
Specifically, we employ a \emph{graph convolutional network} (GCN), which has 
been shown to perform effectively across various combinatorial optimization 
tasks (see \citealt{cappart2023combinatorial} for a comprehensive review).  
Because a GCN operates on graph-structured data, we construct a graph 
representation of the constrained assortment optimization problem under the 
MMNL model.  
We then generate a training dataset by solving a set of representative 
instances and using their optimal assortments as supervision labels.  
Since the goal is to predict whether each product belongs to the optimal 
assortment, the training task can be viewed as a classification problem.  
Once trained, the GCN takes as input the graph representation of a new 
problem instance and outputs the predicted probability that each product is 
included in the optimal assortment.  

Given the predicted inclusion probabilities, we next design inference policies 
to construct feasible assortments.  
We first propose an \emph{index policy} that ranks all products in descending 
order of their indices and evaluates all feasible assortments consisting of 
the top-$n$ products for $n = 1, 2, \ldots, N$, where $N$ is the total number 
of products.  
This approach is inspired by the well-known \emph{revenue-ordered policy} 
(e.g., \citealt{talluri2004revenue}), which can be viewed as a special case of 
an index policy based on product prices.  
In our setting, the indices are determined by the inclusion probabilities 
predicted by the GCN, leading to what we refer to as the 
\emph{GCN-based Index (GI) policy}.  
The GI policy is highly efficient computationally, as it evaluates at most 
$N$ candidate assortments—orders of magnitude fewer than the 
$2^N - 1$ possible subsets.  To further enhance performance, we incorporate the local-search procedure of 
\citet{gallego2024efficient} to refine the solution obtained by the GI policy, 
yielding the \emph{GCN-based Index with Local Search (GILS) policy}.  
Finally, we introduce the \emph{GCN-based Integer Programming (GIP) policy}, 
which formulates an integer linear program in which the nonlinear objective of 
the original assortment problem is replaced by the linear sum of the GCN's 
predicted probabilities.  
This reformulation substantially reduces computation time while maintaining 
high solution quality.

To address the second research question, we leverage a key advantage of the 
GCN—its inherent ability to generalize across problem instances of different 
sizes.  
Because the GCN employs a weight-sharing mechanism across all nodes in the 
graph, a model trained on small-scale instances can naturally be applied to 
larger instances without retraining.  
This property enables a \emph{from-small-to-large} learning paradigm, in which 
the GCN first learns structural patterns from tractable small-scale samples 
and is then used to design policies capable of solving large-scale assortment 
problems efficiently.  We also provide theoretical foundation for this size generalization ability in a special case. 

To demonstrate the effectiveness of this idea, we conduct extensive numerical 
experiments.  
Specifically, we train the GCN on 2{,}000 instances with 20~products and test 
it on large-scale instances with up to 2{,}000~products.  
The results show that the GI policy achieves excellent performance, with a 
competitive ratio exceeding 85\% within milliseconds, outperforming most 
existing heuristic methods in both quality and speed.  
The GILS and GIP policies achieve even higher performance levels 
(approximately 95\% competitive ratio) while requiring only a few seconds of 
computation.  
These results indicate that the structural patterns of optimal assortments in 
small-scale problems closely resemble those in large-scale problems, and that 
the GCN effectively captures and transfers these shared patterns.  To further examine this generalization ability, we evaluate the proposed 
policies on test sets with varying numbers of products, customer types, and 
constraints.  
We observe that performance improves as the size of the training instances 
increases.  
Motivated by this, we experiment with training on larger instances 
($N = 200$).  
Because solving these larger instances optimally is computationally intensive, 
we use suboptimal solutions as training labels and find that a GCN trained on 
such suboptimal samples can still enhance the overall performance of the 
proposed policies.

To address the third research question, we consider the 
\emph{unknown-choice-model} setting, where the underlying choice model is fixed 
but unobserved, and abundant historical transaction data are available.  
In this case, we first train a GCN—referred to as the \emph{choice-GCN}—using 
the transaction data to approximate the underlying choice probabilities.  
Next, we generate small-scale constrained assortment instances and compute 
their approximate optimal solutions by evaluating the expected revenues of all 
feasible assortments based on the choice-GCN's predictions.  
These approximate solutions serve as labels for training a second GCN, termed 
the \emph{solution-GCN}, which is constructed analogously to the network used 
in the known-choice-model case.  
Using the probabilities predicted by the solution-GCN, we then apply the same 
inference policies (GI, GILS, and GIP) as before, except that the expected 
revenues of candidate assortments are estimated via the choice-GCN rather than 
the true choice model.  Numerical experiments show that the proposed policies achieve over 85\% of the 
optimal expected revenue on large-scale instances (with up to 
2{,}000~products) within seconds, confirming both their effectiveness and 
computational efficiency.  
Furthermore, the choice-GCN component can be replaced by other state-of-the-art 
choice-probability estimators—such as those proposed by 
\citet{aouad2022representing}, \citet{wang2023transformer}, and 
\citet{liu2025beyond}—without altering the overall framework.  
As long as the estimation model provides accurate choice probability 
predictions for assortments of varying sizes, the resulting approach can still 
produce high-quality solutions.  

In summary, to the best of our knowledge, this study presents the first 
\emph{learning-to-optimize} framework for constrained assortment optimization.  
The proposed GCN-based approach effectively addresses the scalability 
challenge by exploiting its weight-sharing mechanism, which naturally 
accommodates inputs of varying sizes.  
Moreover, the framework applies seamlessly to both known and unknown choice 
model settings. 
Our theoretical and numerical results demonstrate that the GCN-based approach provides 
scalable, fast, and high-quality solutions, highlighting its potential as a 
promising direction for future research in data-driven assortment optimization. Our contributions primarily lie in the proposed graph representations for constrained assortment problems, the design of GCN-based inference policies, and the theoretical justification of the GCN's expressiveness and size-generalization ability.

The remainder of this paper is organized as follows. In the rest of this section, we review literature related to our work. In Section~\ref{sec:model}, we formally introduce the constrained assortment optimization problem under the MMNL model. In Section~\ref{sec:gcn_policy}, we present the GCN framework, based on which we propose three policies. In Section~\ref{sec:theory}, we provide theoretical results to explain GCNs' capabilities. In Section~\ref{sec:numerical}, we compare our proposed policies with several existing policies through extensive numerical experiments. In Section~\ref{sec:other_choice}, we consider other commonly used choice models.  In Section~\ref{sec:model-free}, we adapt our policies to fit the case where the underlying choice model is fixed but unknown, and then test their performance through numerical experiments. We conclude the paper in Section~\ref{sec:conclusion}. 

\subsection{Literature Review}
This work is broadly related to three streams of research: choice modeling, assortment optimization, and GCN for optimization. In the following, we review these streams respectively.

\tb{Choice modeling.} Modeling customers' choice behaviors is fundamental and crucial in revenue management. In order to depict the substitution between products, researchers have proposed various discrete choice models. We refer readers to \cite{gallego2019revenue} and \cite{wang2021discrete} for comprehensive reviews of choice models. In the following, we only review some related works. One of the most well-studied choice models is the multinomial logit (MNL) model proposed in \cite{mcfadden1972conditional}. It has gained significant attention from researchers in economics and operations research. For instance, \cite{feldman2022customer} conduct field experiments on Alibaba platform and show that the policy based on the MNL model can significantly outperform the existing machine learning policy ignoring the substitution effect. Despite its broad applications, the MNL model has also been criticized due to the unrealistic independence of irrelevant alternatives (IIA) property (\citealt{hausman1984specification}). In order to address this challenge, \cite{mcfadden2000mixed} propose the mixed MNL (MMNL) model, which assumes that the market consists of customers with different latent types, and the behavior of each type is characterized by an MNL model. It is also shown in \cite{mcfadden2000mixed} that the MMNL can approximate a large class of choice models to any degree of accuracy under mild conditions. Other popular choice models include the nested logit (NL) model (\citealt{mcfadden1980econometric}), the Markov chain (MC) model (\citealt{blanchet2016markov}), and the exponomial model (\citealt{alptekinouglu2016exponomial}). 
In this work, we present how to construct graph representations for these choice models, and then use GCN to solve corresponding assortment optimization problems.

In recent years, there have been some works using neural networks to approximate customers' choice probabilities. For example, in order to mitigate the IIA property of the MNL model, \cite{rosenfeld2020predicting} adopt a permutation-invariant neural network, which can be seen as an edge-free counterpart of GCN, to extract features related to the provided product set such that the resulting choice probabilities can depend on the product set. Following this idea, \cite{aouad2022representing} incorporate a sample-average approximation scheme such that the resulting approximation is consistent with the random utility theory.  \cite{wang2023transformer}, \cite{zhang2024modeling}, and \cite{liu2025beyond} adopt the attention mechanism or a variant of GCN such that the neural networks can better capture cross-product interactions. In this work, we provide a general framework to use neural networks to represent various choice models but our focus is not on estimating the choice probabilities but rather on the corresponding assortment optimization problems.

\tb{Assortment optimization. } Assortment optimization is a classic problem in revenue management, and many researchers are devoted to proposing efficient policies for the problem under different choice models. Among many others, the revenue-order policy has gained significant attention from both researchers and practitioners due to its efficiency. For many \textit{unconstrained} assortment problems, researchers have provided performance guarantees for the revenue-order policy. For example, \cite{talluri2004revenue} show that the revenue-order policy is optimal for the unconstrained assortment optimization problem under the MNL model. Under the MMNL model and the MNL model with network effects, \cite{rusmevichientong2014assortment} and \cite{wang2017consumer} identify some conditions under which the revenue-order policy is optimal, respectively. Furthermore, \cite{berbeglia2020assortment}, \cite{gallego2021bounds} and \cite{gallego2024assortment} provide performance guarantees for the policy when the choice model satisfies some mild conditions. 

As mentioned, the revenue-order policy is a special index policy, and the performance can be further improved if we can design suitable indices. For example, \cite{rusmevichientong2010dynamic} establish the optimality of an index policy for the cardinality-constrained problem under the MNL model, and provide an efficient algorithm to compute the indices. \cite{gallego2015general} show that an index policy based on modified revenues is optimal for the unconstrained problem under the general attraction model (GAM). In this work, we show that with a properly-designed machine learning approach, the corresponding index policy can perform quite well even for large-scale problems. The resulting solution can be further refined by the local-search algorithm proposed in \cite{gallego2024efficient}. In addition, we also propose another GCN-based policy that replaces the nonlinear objective in the initial integer programming with the summation of indices, resulting in an integer linear programming.

A recent and closely related paper by \cite{guo2025solving} proposes a label-free, instance-specific neural optimization framework for assortment optimization. For each problem instance, their method trains feedforward neural networks \textit{from scratch} and uses straight-through estimators to optimize binary assortment decisions. In contrast, our framework follows a train-once-solve-many paradigm: the GCN is trained offline on labeled small-scale instances and then directly deployed to solve new, much larger instances without retraining. This design substantially reduces online computation time and the burden of training process, making it more suitable for repeated and time-sensitive assortment decisions.

There are also two recent works that utilize machine learning approaches to solve unknown-choice-model assortment problems. First, \cite{chen2023machine} propose a variant of GCN to approximate the choice probabilities, and then directly implement a local-search algorithm similar to \cite{gallego2024efficient} based on the trained network to solve the assortment problem under one cardinality constraint. In contrast, based on the approximated choice model (i.e., choice-GCN), we generate solution samples and train another GCN (i.e., solution-GCN) to learn the pattern of the optimal solution, resulting in computationally efficient and well-performing policies for the assortment problem under general linear constraints. Even though we also use the local-search algorithm in some policies, the starting solution predicted by the solution-GCN is much better than an empty set, resulting in a better performance and a shorter running time.
Second, \cite{wang2023neural} train a fully connected neural network to approximate the choice probabilities, and then formulate the assortment problem as a non-linear integer programming based on the closed-form expression of the neural network. However, as the number of products increases, the number of learnable weights of the neural network increases linearly, and the computation time of the non-linear integer programming will become prohibitive. In contrast, the number of weights in our proposed GCNs does not increase in the product number, and the inference policies can output high-quality solutions efficiently.

\tb{GCN for optimization.} As an emerging topic in computer science, learning to optimize typically refers to training neural networks to imitate the behaviors of a (near-)optimal policy. In this area, the efficiency of GCNs has already been demonstrated in several related tasks. Readers are referred to \cite{cappart2023combinatorial} for a comprehensive review. In the following, we review some closely related works. One of the pioneering works is \cite{gasse2019exact} which adopts the GCN to guide the branch-and-bound algorithm for mixed integer linear programming. In particular, the authors first construct a bipartite graph to represent the constraints and the objective, and then train a GCN to imitate the behaviors of a strong branching rule, which is time-consuming but usually results in fewer branching operations. Their numerical experiments show that a GCN trained on small-scale instances can speed up the computation of large-scale instances. Following this idea, \cite{fan2023smart} train a GCN to determine the initial basis in the simplex method for large-scale linear programs. Rather than relying on existing heuristics, they directly learn from training samples with the optimal basis as the label. Similarly, \cite{liu2024learning} utilize a GCN to determine the pivoting rule of the simplex method. They first propose two prophet policies based on the knowledge of the optimal solution, and then train a GCN to imitate their behaviors based on only the problem parameters and local information related to the current basis. In this work, we also represent constraints by a bipartite graph, and train a GCN to directly learn from optimal solutions. However, since the objective function of the assortment problem is non-linear and non-monotone, we need to modify the bipartite graph based on the problem's features, and design new inference policies accordingly. 

Moreover, some researchers develop theoretical results to explain the capability of GCNs (see \citealt{jegelka2022theory} and \citealt{zhang2024expressive} for comprehensive reviews). Some researchers investigate the expressive power of GCNs. For example, \cite{xu2018how} and \cite{azizian2021expressive} develop theoretical frameworks to analyze the expressiveness of GCNs by deriving approximation guarantees. Then, \cite{chen2023gnn-lp,chen2023on,chen2024rethinking,chen2025expressive} investigate GCNs' expressive power in optimization problems, such as mixed integer linear programming. Another stream of works investigate the size generalization ability. For example, \cite{yehudai2021local} caution that size generalization is not always guaranteed, especially when small- and large-scale instances do not share similar local or structural patterns. Then, researchers identify some scenarios where the phenomenon happens. \cite{ruiz2020graphon} establish size generalization guarantees when graphs are generated from a graphon model. \cite{maskey2022generalization} provide generalization guarantees for large random graphs. In this work, we investigate GCNs' expressiveness in constrained assortment problems, and theoretically establish its size generalization ability in this setting.

\section{Model}\label{sec:model}

In this paper, we study the assortment optimization problem with linear constraints. Consider $N$ products with fixed price vector $\bs{r} \in \mathbb{R}_+^N$. 
Without loss of generality, we assume zero product costs. Let $[N] := \{1, 2, \dots, N\}$ denote the set of all products, and let $0$ represent the no-purchase option. 

For any assortment $S \subseteq [N]$, define the binary vector 
$
\bs{x}_S := \{x_{S,j} = \1_{\{j \in S\}}\}_{j=1}^N,
$
where $x_{S,j} = 1$ if product $j$ is included in the assortment $S$ and 
$x_{S,j} = 0$ otherwise. Conversely, for any binary vector 
$\bs{x} \in \{0,1\}^N$, we let 
$
S_{\bs{x}} := \{j \in [N] \mid x_j = 1\}
$
denote the corresponding assortment represented by $\bs{x}$. For simplicity, we 
will often write $\bs{x}_S$ simply as $\bs{x}$ when the underlying assortment 
$S$ is clear from context. 

The choice probability of product $j$ under assortment $S$ is denoted by 
$\phi_j(\bs{x}_S)$ and follows a given choice model. To capture practical 
business constraints, we consider $M$ linear inequalities of the form 
$\bs{A}\bs{x} \le \bs{b}$, where $\bs{A} \in \mathbb{R}^{M \times N}$ and 
$\bs{b} \in \mathbb{R}^M$. The resulting constrained assortment optimization 
problem can be formulated as the following nonlinear integer program:
\begin{align}
\max_{\bs{x} \in \{0,1\}^N} \quad 
& \mathcal{R}(\bs{x}) := \sum_{j=1}^N \phi_j(\bs{x})\, r_j, 
\label{eq:main} \\
\text{s.t.} \quad 
& \bs{A}\bs{x} \le \bs{b}. \nonumber
\end{align}

In this paper, we first focus on a specific choice model, the Mixed Multinomial 
Logit (MMNL) model; other choice models will be discussed in 
Section~\ref{sec:other_choice}. The MMNL model generalizes the well-known 
Multinomial Logit (MNL) model and can approximate a broad class of random 
utility models to arbitrary accuracy under mild conditions 
(\citealt{mcfadden2000mixed}). 

Under the MMNL model, customers are assumed to belong to one of $K$ latent 
types, which are unobservable to the seller. Let $\alpha_k \in \mathbb{R}_+$ 
denote the proportion of type-$k$ customers, satisfying 
$\sum_{k=1}^K \alpha_k = 1$. Each customer type follows an MNL model. 
Specifically, the random utility of product $j$ for a type-$k$ customer is 
\begin{align*}
U_{kj} = u_{kj} - \eta r_j + \xi_{kj},
\end{align*}
where $\eta \in \mathbb{R}_+$ represents the price sensitivity, 
$u_{kj} - \eta r_j$ is the deterministic utility, and $\xi_{kj}$ is an 
i.i.d.\ standard Gumbel error term capturing idiosyncratic preferences. 
Without loss of generality, we normalize $u_{k0} = 0$ and $r_0 = 0$, so the 
utility of the no-purchase option for a type-$k$ customer is 
$U_{k0} = \xi_{k0}$.

Define the attraction matrix $\tb{V}\in \mathbb{R}^{K\times N}$ with $v_{kj} := \exp(u_{kj} - \eta r_j)$. Given an assortment vector $\bs{x} \in \{0,1\}^N$, the probability that a type-$k$ customer chooses product $j$ is 
\begin{align*}
\psi_{kj}(\bs{x}) 
&= \mP\!\left(U_{kj} \ge U_{k\ell}, \ \forall\, \ell \in S_{\bs{x}} \cup \{0\}\right) 
   \cdot x_j \\[2mm]
&= \frac{x_j \exp(u_{kj} - \eta r_j)}
        {1 + \sum_{\ell=1}^N x_\ell \exp(u_{k\ell} - \eta r_\ell)} 
 = \frac{x_j v_{kj}}
        {1 + \sum_{\ell=1}^N x_\ell v_{k\ell}}.
\end{align*} 
Thus, the overall choice probability of product $j$ across all customer 
types is 
\begin{align}\label{eq:choice_prob}
\phi_j(\bs{x}) 
= \sum_{k=1}^K \alpha_k \, \psi_{kj}(\bs{x}) 
= \sum_{k=1}^K 
  \frac{\alpha_k x_j v_{kj}}
       {1 + \sum_{\ell=1}^N x_\ell v_{k\ell}}.
\end{align}

According to \citet{rusmevichientong2014assortment}, the assortment optimization 
problem~\eqref{eq:main} under the MMNL model is NP-hard, even in the special 
case with only two customer types ($K = 2$) and no constraints 
($M = 0$). Given this computational hardness, our goal in this paper is to 
develop an efficient algorithm that can produce high-quality solutions to 
problem~\eqref{eq:main} within a short computation time.

\section{GCN-Based Policies}\label{sec:gcn_policy}

In this section, we introduce the graph convolutional network (GCN) and explain 
how it can be applied to solve the constrained assortment optimization problem. 
Section~\ref{sec:highlevel-GCN} outlines the high-level intuition behind the GCN 
approach, emphasizing its contrast with standard neural networks. 
Section~\ref{subsec:graph_represent} then describes a graph-based representation 
of the problem. Section~\ref{sec:GCN} presents our proposed GCN framework, and 
Section~\ref{subsec:infer} discusses how to use the GCN output to construct a 
feasible solution for the constrained assortment optimization problem.

\subsection{High-Level Idea of GCN} \label{sec:highlevel-GCN}

Before presenting the technical details of our approach, we first discuss the 
limitations of standard \textit{feed-forward networks} (FFNs) and explain how 
graph convolutional networks (GCNs) address these challenges. This comparison 
highlights why GCNs are particularly well suited for solving constrained 
assortment optimization problems.

\begin{figure}[ht]
    \centering
    \includegraphics[width=0.8\linewidth]{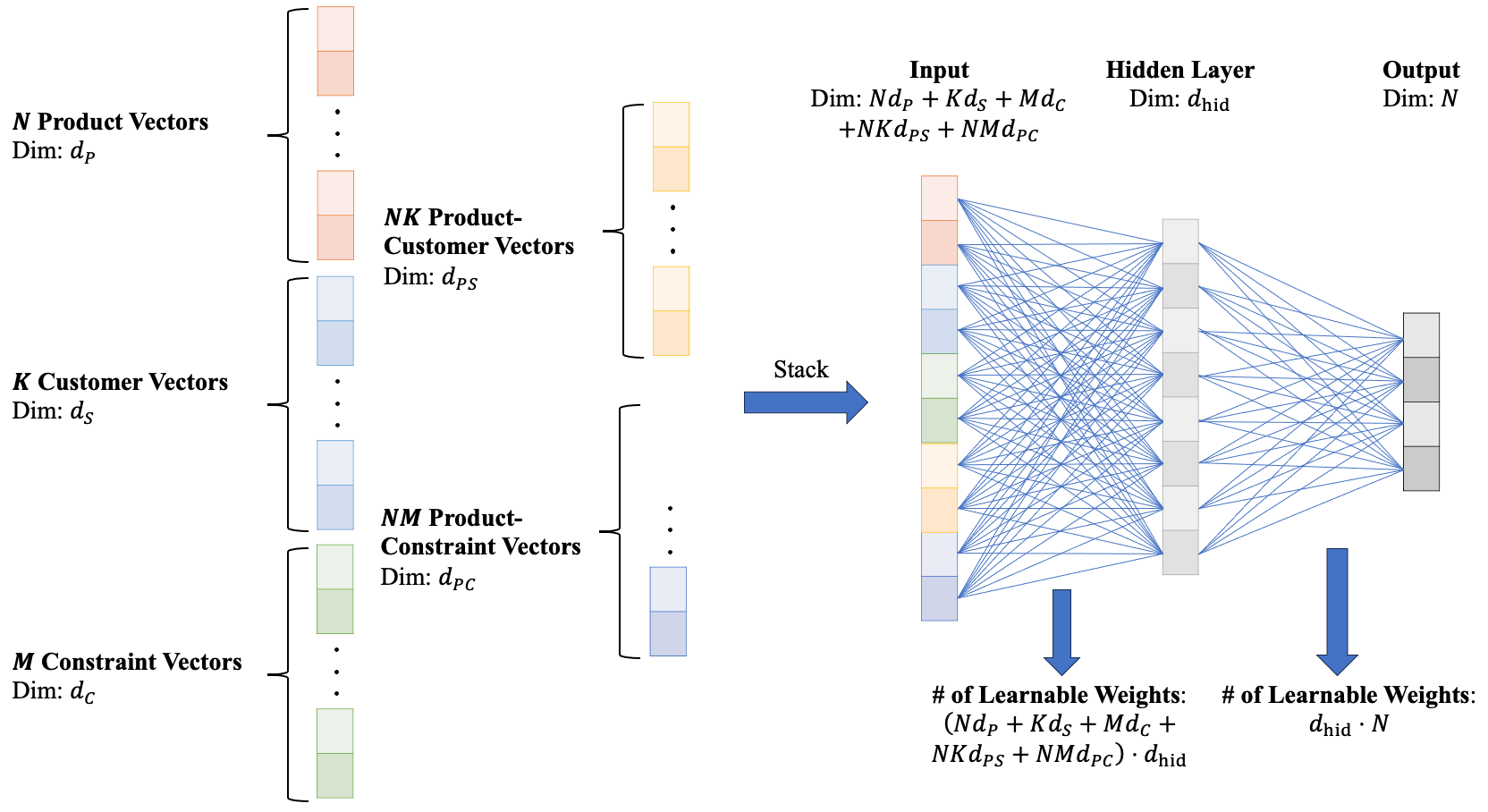}
    \caption{Illustration of feed-forward neural network with one hidden layer; each blue line represents a learnable weight.}
    \label{fig:ffn}
\end{figure}

In FFNs, both inputs and outputs are typically represented by 
\textit{fixed-size} vectors. FFNs process inputs through a sequence of dense 
layers and nonlinear activations, learning complex representations via 
hierarchical transformations. In the context of assortment optimization, each 
product is characterized by a feature vector (e.g., price, utility score, and 
other attributes), and the network input is constructed by stacking these 
vectors into a single input vector 
(see, e.g., \citealt{wang2023neural, wang2024neural}). 
Customer-type information and constraint parameters can also be incorporated 
into the input, enabling the network to learn a mapping from problem parameters 
to binary decisions that indicate which products to include in the assortment. 
An illustration is provided in Figure~\ref{fig:ffn}.

While powerful, FFNs face two main limitations in the context of assortment 
optimization:
\vspace{1mm}
\begin{enumerate}

\item \textbf{Limited generalization across problem sizes.} 
As illustrated in Figure~\ref{fig:ffn}, the number of learnable parameters in an 
FFN grows linearly with the number of products, customer types, or constraints. 
Consequently, an FFN must be trained for fixed input and output dimensions. 
Although a trained FFN can handle smaller instances by padding unused entries 
with zeros, the architecture must be redesigned and retrained if the problem 
size increases (e.g., from 20 to 500 products). 

\vspace{1mm}
\item \textbf{Lack of structural awareness.} 
FFNs treat all inputs as flat feature vectors and do not naturally capture 
the relational structure among products, customer types, and constraints. 
Without an architecture that reflects this inherent structure, learning becomes 
less efficient and its generalization capability becomes weaker—especially in data-sparse or dynamic 
environments. For instance, if we permute the product indices, the optimal 
assortment (after index recovery) should remain unchanged. However, in a stacked 
input representation, the FFN's output (after index recovery) typically differs 
from the original one because the learned weights depend on feature locations in the 
input vector.
\end{enumerate}

\begin{figure}[ht]
    \centering
    \includegraphics[width=0.8\textwidth]{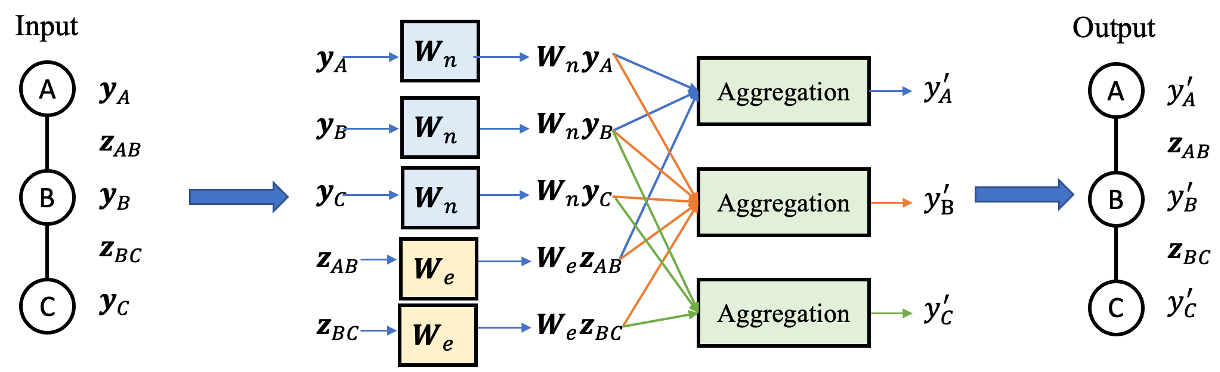}
    \caption{Illustration of GCN computation in a toy example. Each node (A, B, 
    C) has a feature vector $y$, and each edge (A--B, B--C) has a feature vector 
    $z$. Blue boxes represent learnable weight matrices for nodes 
    ($\bs{W}_n$), and yellow boxes represent learnable weight matrices for edges 
    ($\bs{W}_e$). Boxes of the same color share the same weights, reflecting the 
    \emph{weight-sharing} mechanism that makes GCNs scalable. Each node updates 
    its representation by aggregating messages from its neighbors through a 
    parameter-free aggregation function.}
    \label{fig:gcn_illustrate}
\end{figure}

\vspace{1mm}
In contrast, graph convolutional networks (GCNs) are naturally suited for 
problems whose sizes and structures vary across instances—for example, 
assortment problems with different numbers of products, customer types, or 
constraints. GCNs consist primarily of \textit{message-passing} layers and 
operate directly on graph-based representations, where nodes correspond to 
entities (e.g., products, customers, or constraints) and edges encode their 
relationships or interactions. This design enables GCNs to process inputs of 
arbitrary size using a common set of shared parameters, allowing them to learn 
more efficiently and to generalize across instances with different scales and 
structures. For clarity, we illustrate the computation process of a GCN with a 
single message-passing layer in Figure~\ref{fig:gcn_illustrate}. More details on 
the specific GCN architecture used for our assortment optimization problem will 
be provided in the next subsection.

\vspace{1mm}
Figure~\ref{fig:gcn_illustrate} provides a simplified view of how a single GCN 
layer operates. Each node (A, B, and C) represents an entity—such as a product, 
a customer type, or a constraint—and each edge (e.g., A--B or B--C) represents 
a relationship or interaction between two entities. In the GCN, information 
flows along these edges: for example, node~A sends a message to node~B using 
their respective node and edge features, and node~B aggregates messages from all 
connected neighbors before updating its own representation. 

From a practical standpoint, an edge encodes a quantitative relationship between 
two elements of the assortment problem. For instance, an edge between a 
\textit{customer node} and a \textit{product node} captures how much that 
customer type values the product (i.e., its utility or choice probability 
weight), whereas an edge between a \textit{product node} and a 
\textit{constraint node} represents how the product contributes to a capacity 
or display constraint (e.g., shelf space, budget, or compatibility rule). These 
edges define the channels through which information is exchanged in the 
network—allowing the GCN to learn how customer preferences and business 
constraints jointly shape the optimal assortment. 

According to Figure~\ref{fig:gcn_illustrate}, each layer of a GCN performs two 
main operations. First, the feature vectors on nodes and edges are transformed 
through matrices with shared weights. Then, each node aggregates information 
from its connected neighbors and edges and combines it with its own feature 
vector. In this way, every node updates its representation based on both its 
local attributes and the surrounding network structure. In what follows, we 
explain how these operations allow GCNs to overcome the limitations of FFNs.

\vspace{1mm}
\begin{enumerate}
\item \textbf{Scalability through weight sharing.} 
A key feature of GCNs is that the same weight matrices are shared across all 
nodes and edges, so the number of learnable parameters does not increase with 
the size of the graph. For example, in Figure~\ref{fig:gcn_illustrate}, the GCN 
only needs to learn two weight matrices, $\bs{W}_n$ and $\bs{W}_e$, whose 
dimensions are fixed regardless of how many nodes or edges are present. This 
weight-sharing mechanism makes GCNs highly scalable and allows a network trained 
on small instances (e.g., with 20 products) to generalize effectively to much 
larger instances (e.g., with 2{,}000 products) without retraining or modifying 
the architecture. Conceptually, the weight-sharing mechanism can be viewed as a 
low-rank approximation of the large weight matrix used in a standard FFN.

\vspace{1mm}
\item \textbf{Context-aware representations.} 
Through the message-passing process, each node incrementally builds a 
representation that reflects not only its own features but also its structural 
context within the graph. In our setting, this means that each product node 
learns an embedding that captures its price and utility, along with demand 
signals from customer nodes and feasibility information from constraint nodes. 
By stacking multiple GCN layers, the network can capture higher-order 
dependencies—such as interactions among products that share customers or 
constraints—leading to more accurate predictions of which products should be 
included in the optimal assortment.
\end{enumerate}

\subsection{Graph Representation for Assortment Optimization Problem}
\label{subsec:graph_represent}

To apply a GCN to the assortment optimization problem, we first construct a 
graph that captures the key structural relationships in the problem—namely, 
customer preferences, product characteristics, and linear constraints. 
The construction follows the same principle as the toy example in 
Figure~\ref{fig:gcn_illustrate}, where nodes exchanged information through 
edges representing pairwise relationships. 
Here, the same idea is extended to model how customers, products, and 
constraints interact in the assortment optimization setting. 
Each node represents an entity (customer type, product, or constraint), and 
edges specify quantitative relationships—such as how strongly a customer 
values a product or how much a product consumes a limited resource—along which 
information flows during message passing. 

\begin{figure}[h]
    \centering
    \includegraphics[width=0.525\textwidth]{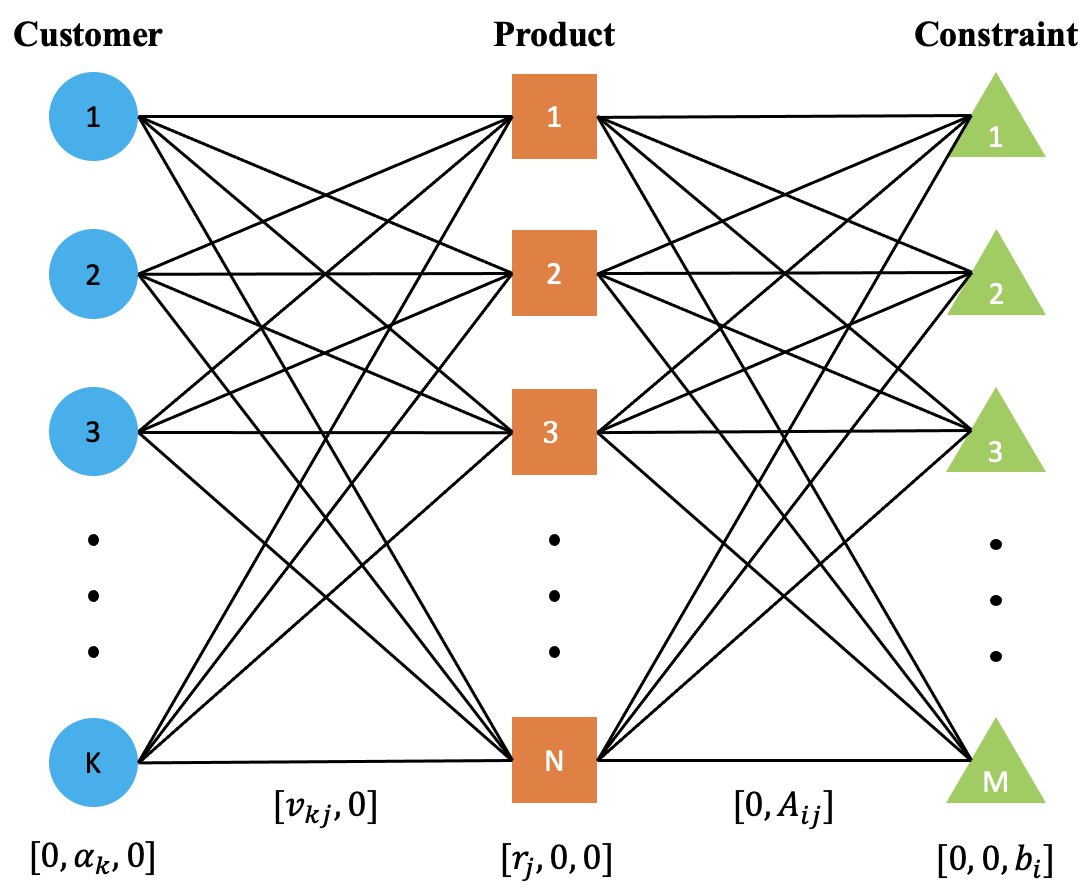}
    \caption{Graph representation of the constrained assortment problem under 
    the MMNL choice model. Customer nodes (blue circles) are connected to 
    product nodes (orange squares) through edges weighted by utilities 
    $v_{kj}$, while product nodes are connected to constraint nodes (green 
    triangles) through edges weighted by resource coefficients $A_{ij}$.}
    \label{fig:MMNL_graph}
\end{figure}

As illustrated in Figure~\ref{fig:MMNL_graph}, the resulting graph contains 
three types of nodes:
\begin{itemize}
    \item \textbf{Customer nodes} ($\bigcirc$): One node for each customer type 
    $k \in [K]$.
    \item \textbf{Product nodes} ($\square$): One node for each product 
    $j \in [N]$.
    \item \textbf{Constraint nodes} ($\triangle$): One node for each linear 
    constraint $i \in [M]$.
\end{itemize}

\vspace{1mm}
\noindent
Formally, let $\mathcal{N} = \mathcal{N}^S \cup \mathcal{N}^P \cup 
\mathcal{N}^C$ denote the full set of nodes, where $\mathcal{N}^S = [K]$ is 
the set of customer nodes, $\mathcal{N}^P = [N]$ is the set of product nodes, 
and $\mathcal{N}^C = [M]$ is the set of constraint nodes. 
The graph includes two types of undirected edges:
\begin{itemize}
    \item Between customer node $k$ and product node $j$, capturing the utility 
    weight $v_{kj}$;
    \item Between product node $j$ and constraint node $i$, capturing the 
    resource consumption $A_{ij}$.
\end{itemize}
We denote the edge set by 
$\mathcal{E} = \mathcal{E}^S \cup \mathcal{E}^C$, where 
$\mathcal{E}^S = \mathcal{N}^S \times \mathcal{N}^P$ and 
$\mathcal{E}^C = \mathcal{N}^P \times \mathcal{N}^C$.

Each node in the graph is associated with a feature vector of dimension 
$d^N = 3$ that encodes problem-specific information. 
We define the node feature matrices 
$\bs{Y}^S \in \mathbb{R}^{K \times d^N}$ for customer nodes, 
$\bs{Y}^P \in \mathbb{R}^{N \times d^N}$ for product nodes, and 
$\bs{Y}^C \in \mathbb{R}^{M \times d^N}$ for constraint nodes. 
The specific features are defined as follows:
\begin{itemize}
    \item Customer node $k$: $\bs{Y}^S_{k,:} = [0, \alpha_k, 0]$;
    \item Product node $j$: $\bs{Y}^P_{j,:} = [r_j, 0, 0]$;
    \item Constraint node $i$: $\bs{Y}^C_{i,:} = [0, 0, b_i]$.
\end{itemize}

\vspace{1mm}
\noindent
Each edge is also associated with a feature vector of dimension $d^E = 2$. 
We define the edge feature tensors 
$\bs{Z}^S \in \mathbb{R}^{N \times K \times d^E}$ for edges between product 
and customer nodes, and 
$\bs{Z}^C \in \mathbb{R}^{M \times N \times d^E}$ for edges between constraint 
and product nodes. 
Each edge feature is specified as follows:
\begin{itemize}
    \item Edge $(k, j)$ between customer $k$ and product $j$: 
    $\bs{Z}^S_{j,k,:} = [v_{kj}, 0]$;
    \item Edge $(j, i)$ between product $j$ and constraint $i$: 
    $\bs{Z}^C_{i,j,:} = [0, A_{ij}]$.
\end{itemize}

\vspace{1mm}
If additional contextual information is available—such as customer demographics 
or product metadata—it can be incorporated by expanding the node feature 
matrices $\bs{Y}^S$, $\bs{Y}^P$, or the edge feature tensors 
$\bs{Z}^S$, $\bs{Z}^C$ accordingly. 
This graph-based formulation captures both the demand and constraint structures 
of the assortment problem in a form that can be directly processed by a GCN. 

Our proposed graph representation is inspired by \cite{gasse2019exact}, which 
constructs a bipartite graph of product and constraint nodes to represent mixed 
integer linear programs. 
Different from their linear objective, the objective in problem~\eqref{eq:main} 
is nonlinear and depends on choice probabilities. 
To model this nonlinear coupling, we introduce additional customer nodes and 
their connecting edges to represent how customer preferences interact with 
product and constraint features.

\subsection{The Proposed GCN Framework} \label{sec:GCN}

\begin{figure}[h]
    \centering
    \includegraphics[width=0.95\textwidth]{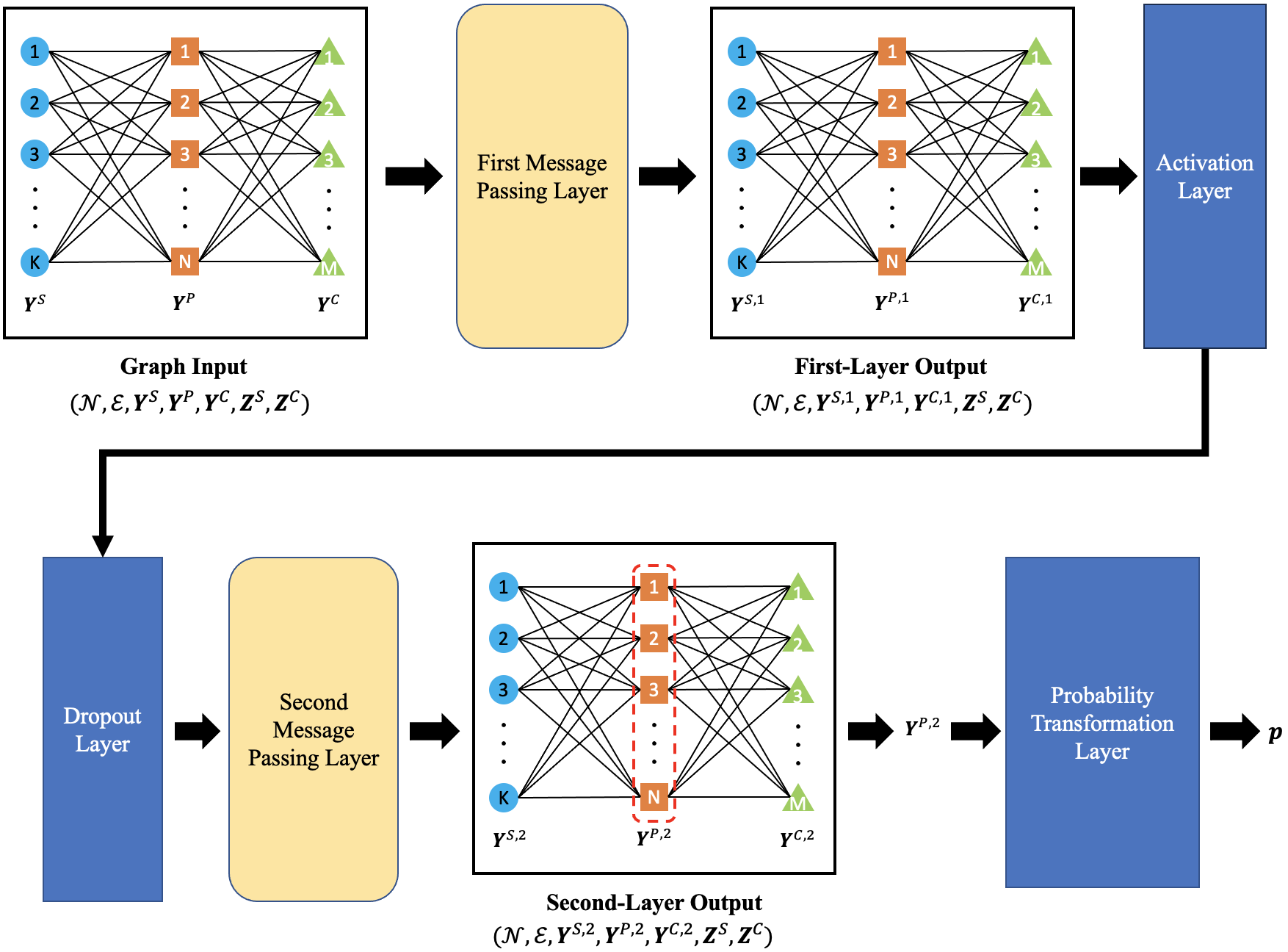} 
    \caption{Illustration of GCN network.}
    \label{fig:GCN_network}
\end{figure}

With the graph representation introduced above, we construct a GCN comprising two message-passing layers, one activation layer, one dropout layer, and one probability transformation layer. This architecture enables the model to iteratively aggregate and refine node features through the graph structure. The message-passing layers extract relational patterns by allowing nodes to exchange information with their neighbors. The activation layer enhances the expressive power of the GCN, and the dropout layer serves as a regularization mechanism to prevent overfitting during training. Finally, the probability transformation layer maps the learned representations of product nodes into selection probabilities. To facilitate understanding, we provide a schematic illustration of 
the GCN architecture in Figure~\ref{fig:GCN_network}. Below, we describe the computations performed in each layer.

\vspace{1mm}
\textbf{First message-passing layer.} 
The first message-passing layer in Figure~\ref{fig:GCN_network} follows the methodology proposed in \cite{li2020deepergcn}. It maps the graph in Figure~\ref{fig:MMNL_graph} into a new graph with updated node features while keeping the edge features unchanged. This layer enables each node to incorporate information from its neighbors through a structured computation process involving message construction, aggregation, and node update. To be self-contained, we describe each step in details below.
\vspace{1mm}
\begin{enumerate}
\item \tb{Information extraction and message construction.} 
In this step, the feature vectors on nodes and edges are transformed by some feature extraction matrices as follows:
\begin{align*}
    &\bs{l}_k^S:= \bs{Y}^S_{k, :}\bs{W}_n^{(1)} \in \mathbb{R}^{1\times d_{\text{mid}}}\qquad \bs{l}_j^P:= \bs{Y}^P_{j, :}\bs{W}_n^{(1)}\in \mathbb{R}^{1\times d_{\text{mid}}} \qquad \bs{l}_i^C:= \bs{Y}^C_{i,:}\bs{W}_n^{(1)}\in \mathbb{R}^{1\times d_{\text{mid}}}\\
    &\bs{q}_{j, k}^{S} = \bs{Z}_{j, k, :}^S\bs{W}_e^{(1)}\in \mathbb{R}^{1\times d_{\text{mid}}} \qquad \bs{q}_{i, j}^C:=\bs{Z}_{i, j, :}^C \bs{W}_e^{(1)}\in \mathbb{R}^{1\times d_{\text{mid}}},
\end{align*}
where $d_{\text{mid}}$ is a hyperparameter (we use $d_{\text{mid}} = 32$ in this work). 
Note that all nodes share the learnable weight matrix $\bs{W}_n^{(1)}\in \mathbb{R}^{d^N\times d_{\text{mid}}}$ and all edges share the learnable weight matrix $\bs{W}_e^{(1)}\in \mathbb{R}^{d^E\times d_{\text{mid}}}$. 

Then, each node prepares messages to send to its neighbors, encoding both its own features and those of the connecting edge. For example, the message from customer node $k$ to product node $j$ is:
\begin{eqnarray*}
\bs{m}_{kj}^{SP} := \text{ReLU}(\bs{l}_k^S 
+ \bs{q}_{j, k}^S) + \epsilon\bs{1} \in \mathbb{R}^{1\times d_{\text{mid}}},
\end{eqnarray*}
where $\text{ReLU}(\bs{x}) = \max\{\bs{x}, \bs{0}\}$ is applied element-wise. The small positive constant $\epsilon$ (e.g., $10^{-7}$) ensures all messages are strictly positive, which can help numerical stability. Moreover, the transformed feature vectors $\bs{l}$'s also serve as local information of nodes.

Similarly, the messages passed from product node $j$ to customer node $k$, from product node $j$ to constraint node $i$, and from constraint node $i$ to product node $j$ are as follows:
\begin{align*}
\bs{m}_{jk}^{PS} &:= \text{ReLU}(\bs{l}^P_j 
+ \bs{q}_{j,k}^S) + \epsilon \bs{1}\in \mathbb{R}^{1\times d_{\text{mid}}}, \\
\bs{m}_{ji}^{PC} &:= \text{ReLU}(\bs{l}^P_j 
+ \bs{q}_{i, j}^C) + \epsilon\bs{1}\in \mathbb{R}^{1\times d_{\text{mid}}}, \\
\bs{m}_{ij}^{CP} &:= \text{ReLU}(\bs{l}_i^C
+ \bs{q}^C_{i, j}) + \epsilon\bs{1}\in \mathbb{R}^{1\times d_{\text{mid}}}.
\end{align*}

\vspace{1mm}
\item \tb{Message aggregation.} 
Once all messages are constructed, each node aggregates the messages received from its neighbors. Unlike simple averaging, we use a softmax-based weighted summation that adjusts the weight of each incoming message based on its magnitude. Define the aggregation operator as:
\begin{align*}
\mathcal{F}\left(\{\bs{m}_1, \dots, \bs{m}_S\}\right) 
:= \bs{m}' \in \mathbb{R}^{1 \times d_{\text{mid}}}, \,\, \mbox{ where } \,\,
m'_\ell := \sum_{s=1}^S 
\frac{e^{m_{s,\ell}}}{\sum_{s'=1}^S e^{m_{s',\ell}}} 
\cdot m_{s,\ell}.
\end{align*}
We use softmax-based aggregation to allow the model to assign higher importance to stronger messages. This weighted scheme helps the network focus on the most relevant signals, amplifying useful patterns and suppressing noise, thereby improving the expressiveness of the learned node representations.

Using this operator, the aggregated hidden representations become:
\begin{align*}
\bs{h}^S_k &:= \mathcal{F}\big(\{\bs{m}^{PS}_{jk}\}_{j=1}^N\big) 
+ \bs{l}_k^S
\in \mathbb{R}^{1 \times d_{\text{mid}}}, \\
\bs{h}^P_j &:= \mathcal{F}\big(\{\bs{m}^{SP}_{kj}\}_{k=1}^K \cup \{\bs{m}^{CP}_{ij}\}_{i=1}^M\big) 
+ \bs{l}_j^P \in \mathbb{R}^{1 \times d_{\text{mid}}}, \\
\bs{h}^C_i &:= \mathcal{F}\big(\{\bs{m}^{PC}_{ji}\}_{j=1}^N\big) 
+ \bs{l}_i^C 
\in \mathbb{R}^{1 \times d_{\text{mid}}}.
\end{align*}

Intuitively, each vector $\bs{h}$ serves an intermediate hidden representation for a node (customer, product, or constraint) after receiving and processing messages from its neighbors. It combines both contextual signals from connected nodes (via softmax-weighted messages) and the node's own local information (via the transformed feature vector). In our setting, $\bs{h}^S_k$ captures how customer type $k$ is influenced by the available products, $\bs{h}^P_j$ reflects product $j$'s relevance, incorporating both demand-side feedback from customers and supply-side constraints, and $\bs{h}^C_i$ summarizes the burden on constraint $i$ based on its connections to products. These representations are then passed through the final transformation step to produce the updated node features for the next layer.

\vspace{1mm}
\item \tb{Node update.} 
After aggregating information from their neighbors, each node holds an intermediate vector $\bs{h}$ 
that summarizes both incoming signals and local context. To convert this into a new feature representation for the next layer, we apply a (shared) two-layer feed-forward neural network. This allows the model to capture non-linear interactions among the aggregated features, increasing the expressive power 
of the node representation. Using two layers with an expansion factor of 2 is a standard design 
that enables the network to approximate a broader class of functions (see, e.g., \citealt{hornik1989multilayer}). 

The updated node features are given by:
\begin{align*}
\bs{Y}^{S, 1}_{k, :} &:= 
\sigma\big(\bs{h}^S_k \bs{T}_1^{(1)} + \bs{b}_1^{(1)}\big) \bs{T}_2^{(1)} + \bs{b}_2^{(1)} \in \mathbb{R}^{1\times d_{\text{mid}}}, \\
\bs{Y}^{P, 1}_{j, :} &:= 
\sigma\big(\bs{h}^P_j \bs{T}_1^{(1)} + \bs{b}_1^{(1)}\big) \bs{T}_2^{(1)} + \bs{b}_2^{(1)} \in \mathbb{R}^{1\times d_{\text{mid}}}, \\
\bs{Y}^{C, 1}_{i, :} &:= 
\sigma\big(\bs{h}^C_i \bs{T}_1^{(1)} + \bs{b}_1^{(1)}\big) \bs{T}_2^{(1)} + \bs{b}_2^{(1)} \in \mathbb{R}^{1\times d_{\text{mid}}},
\end{align*}
where $\sigma(\bs{x}) = 1 / (1 + e^{-\bs{x}})$ is the element-wise sigmoid activation, $\bs{T}_1^{(1)} \in \mathbb{R}^{d_{\text{mid}} \times 2d_{\text{mid}}}$, $\bs{b}_1^{(1)} \in \mathbb{R}^{1 \times 2d_{\text{mid}}}$, $\bs{T}_2^{(1)} \in \mathbb{R}^{2d_{\text{mid}} \times d_{\text{mid}}}$, and 
$\bs{b}_2^{(1)} \in \mathbb{R}^{1 \times d_{\text{mid}}}$. The output of this step is denoted by $\left(\mathcal{N}, \mathcal{E}, \bs{Y}^{S, 1}, \bs{Y}^{P, 1}, \bs{Y}^{C, 1}, \bs{Z}^S, \bs{Z}^C\right)$.
\end{enumerate}

\vspace{1mm}
To summarize, the message-passing layer updates the feature vector of each node based on its local information and its neighbor messages. All nodes share the same weight matrix $\bs{W}_n^{(1)}$, and all edges share the same weight matrix $\bs{W}_e^{(1)}$. Similarly, 
the transformation weights used in the final step (i.e., $\bs{T}_1^{(1)}$, $\bs{T}_2^{(1)}$, $\bs{b}_1^{(1)}$, and $\bs{b}_2^{(1)}$) are also shared across all nodes. This architecture embodies a \emph{weight-sharing mechanism}, under which the number of learnable parameters does not grow with the number of 
nodes or edges in the graph. Instead, the number of learnable parameters depends only on a few model-level quantities: the node feature dimension $d^N$, the edge feature dimension $d^E$, and the hidden dimension $d_{\text{mid}}$. The first two quantities ($d^N$ and $d^E$) reflect how many features are used to describe each node and edge, and are fixed once the feature encoding is specified. The hidden 
dimension $d_{\text{mid}}$ is a tunable hyperparameter that trades off model expressiveness and computational cost.

\vspace{2mm}
\tb{Activation layer.} The activation layer in Figure~\ref{fig:GCN_network} incorporates more nonlinear patterns to the neural network to improve its representation power. We choose the ReLU function as the activation function (see \citealt{glorot2011deep}), and compute the following terms:
\begin{align*}
    \tilde{\bs{Y}}^{S, 1} := \mbox{ReLU}(\bs{Y}^{S, 1}), \quad \tilde{\bs{Y}}^{P, 1} := \mbox{ReLU}(\bs{Y}^{P, 1}), \quad \tilde{\bs{Y}}^{C, 1} := \mbox{ReLU}(\bs{Y}^{C, 1}).
\end{align*}

\vspace{2mm}
\tb{Dropout layer.} The dropout layer in Figure~\ref{fig:GCN_network} is introduced to reduce overfitting by encouraging generalization during training (\citealt{srivastava2014dropout}). 
Given a dropout probability $p_D = 0.5$, each entry in the intermediate feature matrices 
$\tilde{\bs{Y}}^{S, 1}$, $\tilde{\bs{Y}}^{P, 1}$, and $\tilde{\bs{Y}}^{C, 1}$ is independently set to zero with probability $p_D$ or scaled up proportionally during training. The shapes of these matrices remain unchanged, and we use $\bar{\bs{Y}}^{S, 1}$, $\bar{\bs{Y}}^{P, 1}$, and $\bar{\bs{Y}}^{C, 1}$ to denote the updated matrices. Specifically, we update the matrix $\bar{Y}^{S, 1}_{k, \ell}$ as follows:
\begin{align*}
    \bar{Y}^{S, 1}_{k, \ell} := \begin{cases}
        \tilde{Y}^{S, 1}_{k, \ell} \cdot \mathcal{X}^S_{k, \ell}/ (1-p_D) & \text{during training}\\
        \tilde{Y}^{S, 1}_{k, \ell} & \text{during validation or testing,}
    \end{cases}
\end{align*}
where $\mathcal{X}^S_{k, \ell}$ is a random variable and follows a Bernoulli distribution with rate $(1-p_D)$. This stochastic masking discourages the network from over-relying on specific features or neurons, forcing it to learn redundant and robust representations. By doing so, dropout acts as a regularization mechanism 
that helps reduce the risk of overfitting. When computing the validation loss or testing the model, dropout is turned off, and the full network is used without masking.

\vspace{2mm}
\tb{Second message-passing layer.} 
This layer performs another round of message passing similar to the first layer. Given the input $\left(\mathcal{N}, \mathcal{E}, \bar{\bs{Y}}^{S, 1}, \bar{\bs{Y}}^{P, 1}, \bar{\bs{Y}}^{C, 1}, \bs{Z}^S, \bs{Z}^C\right)$, it again aggregates signals from neighbors and combines them with local information, and updates a feature vector of size $d_{\text{mid}}$ for each node. The learnable weight matrices are as follows:
\begin{align*}
    &\bs{W}_n^{(2)}\in \mathbb{R}^{d_{\text{mid}}\times d_{\text{mid}}} \quad \bs{W}_e^{(2)}\in \mathbb{R}^{d^E\times d_{\text{mid}}} \quad \bs{T}_1^{(2)} \in \mathbb{R}^{d_{\text{mid}} \times 2d_{\text{mid}}} \\
    &\bs{b}_1^{(2)} \in \mathbb{R}^{1 \times 2d_{\text{mid}}} \quad \bs{T}_2^{(2)} \in \mathbb{R}^{2d_{\text{mid}} \times d_{\text{mid}}} \quad \bs{b}_2^{(2)} \in \mathbb{R}^{1 \times d_{\text{mid}}},
\end{align*}
and the updated node feature matrices are denoted by $\bs{Y}^{S, 2}\in \mathbb{R}^{K \times d_{\text{mid}}}$, $\bs{Y}^{P, 2}\in \mathbb{R}^{N \times d_{\text{mid}}}$, and $\bs{Y}^{C, 2}\in \mathbb{R}^{M \times d_{\text{mid}}}$. Then, we extract the feature matrix of product nodes, $\bs{Y}^{P, 2}$, for the following computation.

\vspace{2mm}
\tb{Probability transformation layer.} The final probability transformation layer converts the feature matrix $\bs{Y}^{P, 2}$ into a vector of probabilities. Specifically, we compress the feature vector of each product $j$ into a probability as follows:
\begin{align*}
    p_j := \sigma\big(\bs{Y}^{P, 2}_{j, :} \bs{W}^{(3)}\big) \in (0, 1),
\end{align*}
where $\bs{W}^{(3)} \in \mathbb{R}^{d_{\text{mid}}\times 1}$ is the learnable weight matrix shared by all product nodes, and $\sigma(x) = 1 / (1 + e^{-x})$ is the sigmoid function. This reduction serves a purpose: By the above layers, the model has already built rich, context-aware representations for each node. The role of this layer is to distill those representations into a probability $p_j$ for each product $j$, which can then be interpreted as the predicted likelihood that product $j$ should be included in the assortment. These probabilities are then used in the downstream decision-making step, which we discuss in the next subsection.

\vspace{2mm}
\tb{Data augmentation and training procedure.} To train the GCN, we recommend practitioners to first collect historical problem instances, and then use the conic integer programming method proposed in \citet{csen2018conic} to obtain the corresponding optimal solutions \( \bs{x}^*_\ell \)'s. Then, each sample $\ell$ is characterized by \( (N_\ell, \bs{\alpha}_\ell, \bs{r}_\ell, \bs{V}_\ell, \bs{A}_\ell, \bs{b}_\ell, \bs{x}^*_\ell) \).  Given that historical instances can be scarce and the running time of the conic programming can be relatively long, we propose an approach to augment the dataset by generating additional samples from the current samples (see Algorithm~\ref{alg:data_augment}). Specifically, in order to generate a new sample from a given sample $\ell$, we randomly drop a few products that are not in the optimal assortment, and the corresponding optimal assortment does not change due to the property of the problem. Such an operation can increase the diversity of the dataset, and hence enhance the generalization ability of the trained GCN. 

\begin{algorithm}[ht]
    \caption{Data augmentation algorithm}\label{alg:data_augment}
    \small
    \begin{algorithmic}
    \State Input: $(N, \bs{\alpha}, \bs{r}, \bs{V}, \bs{A}, \bs{b}, \bs{x}^*)$.
    \State Set $\Omega_+\gets \{j\in [N_\ell]\mid x^*_j=1\}$, and $\Omega_-\gets \{j\in [N_\ell]\mid x^*_j=0\}$. 
    \State Generate a random number $n_-$ uniformly from $[|\Omega_-|]$, and generate a random subset $\Omega_-'$ of size $n_-$ from the set $\Omega_-$. 
    \State Set $\Omega\gets \Omega_+\cup \Omega_-'$.
    \State Return: $(|\Omega|, \bs{\alpha}, \bs{r}_{\Omega}, \bs{V}_{\Omega}, \bs{A}_{\Omega}, \bs{b}, \bs{x}^*_{\Omega})$, where the subscript $\Omega$ means selecting data corresponding to the set $\Omega$. 
    \end{algorithmic}
\end{algorithm}

We then convert each problem instance into its graph representation, construct the input tensors \( \bs{Y} \) and \( \bs{Z} \) and the labels \( \bs{x}^*\), and split the dataset into 80\% \textit{training} and 20\% \textit{validation} sets. Let $L_T$ denote the number of instances in the training set, and $\kappa_+:=\sum_{\ell=1}^{L_T} \sum_{j=1}^{N_\ell} x_{\ell, j}^*$ and $\kappa_-:=\sum_{\ell=1}^{L_T} \sum_{j=1}^{N_\ell} (1-x_{\ell, j}^*)$ denote the number of different labels. During each training epoch, we first pass the graphs in the training set to the GCN to obtain the predicted product inclusion probabilities \( \bs{p} \)'s. Then, we compute the weighted cross-entropy loss between \( \bs{p} \) and the ground-truth labels \( \bs{x}^* \) as follows:
\begin{align}
    \mathcal{L}_{CE}:= -\frac{\sum_{\ell=1}^L \sum_{j=1}^{N_\ell} \left(\log\left(p_{\ell, j}\right) \cdot w_+\cdot x^*_{\ell, j} + \log\left(1-p_{\ell, j}\right)\cdot w_-\cdot  (1-x^*_{\ell, j}) \right)}{\sum_{\ell=1}^L  \sum_{j=1}^{N_\ell} \left(w_+ \cdot x^*_{\ell, j} + w_- \cdot (1-x_{\ell, j}^*)\right)},\label{eq:CE_loss}
\end{align} 
where the weights $w_+:=(\kappa_++\kappa_-)/\kappa_+$ and $w_-:=(\kappa_++\kappa_-)/\kappa_-$ are introduced to tackle sample imbalance (see \citealt{lin2017focal}). We use the Adam optimizer (see \citealt{kingma2014adam}) to update the model weights to minimize the training loss. At the end of each epoch, in order to monitor generalization performance, we compute the weighted cross-entropy loss of the validation set as the validation loss. If the validation loss of the new model is smaller than that of the retained model, then we replace the retained model with the new model. Note that the validation set is not used in the Adam optimization procedure but is used to compare and select trained neural networks because the validation loss reflects the generalization ability from the training set to the validation set. 

In order to further mitigate overfitting, we adopt the early stopping rule to determine the number of epochs. Specifically, if the validation loss does not improve for a few epochs, then the training process terminates early; otherwise, the training process continues for a predefined number of epochs. After the training process, the final retained model will be used in inference policies, which are specified in the next subsection, to derive solutions for testing samples. 

\begin{remark}[Potential improvement of GCN framework]\label{remark:improvement}
    In this work, for ease of exposition, we consider a shallow network with two message-passing layers and utilize the simplest message-passing operation. The reason is that we aim to illustrate the potential of GCN for assortment problems. In order to improve the expressive power of the GCN, practitioners can increase the number of message-passing layers, and incorporate more advanced message-passing operations. For example, inspired by \cite{gasse2019exact}, the GCN can be improved by assigning a direction-specific edge-feature extraction matrix to each message-passing direction, such as customer-to-product and product-to-customer directions. This design allows the model to capture the fact that edge information may play different roles for different types of nodes.
\end{remark}

\subsection{Inference Policies}\label{subsec:infer}

Given the output probabilities from the GCN, we must still determine which 
products to include in the final assortment. A naive approach would be to 
select all products whose predicted inclusion probabilities exceed 0.5. 
However, this method can yield infeasible assortments that violate the linear 
constraints and perform poorly because the objective function is 
non-monotonic—adding a product can reduce the expected revenue. 
To address these challenges, we design several inference policies that translate 
the GCN outputs into feasible and high-quality assortments. 

\vspace{1mm}
\textbf{(1) GCN-based Index (GI) Policy.} 
Our first approach is an \textit{index policy} (Algorithm~\ref{alg:index_policy}). 
This policy sorts all products in descending order of their predicted 
probabilities $p_j$ and evaluates all feasible assortments formed by the top-$n$ 
products for $n \in [N]$. 
Among these candidates, it selects the feasible assortment with the highest 
objective value. 
The design is inspired by the well-known \textit{revenue-ordered policy} 
(\citealt{talluri2004revenue}) and related approaches in business practice 
(\citealt{feldman2022customer}), which can be viewed as index policies based on 
product prices or individual expected revenues. 
Other index-based heuristics also appear in the literature: under the 
cardinality-constrained MNL model, \citet{rusmevichientong2010dynamic} show that 
the optimal policy ranks products by $v_j(r_j - \lambda)$ for some $\lambda$, 
and under the unconstrained GAM model, \citet{gallego2015general} prove the 
optimality of ranking by $r_j v_j / (v_j - w_j)$, where $w_j$ is the shadow 
attraction of product $j$. 
In our setting, the probabilities $p_j$ serve as \emph{GCN-based 
indices} that integrate both demand-side and constraint-side information. 
We refer to the resulting policy as the GCN-based Index (\textbf{GI}) policy.

\vspace{4mm}
\begin{breakablealgorithm}
    \caption{Index policy}\label{alg:index_policy}
    \small
    \begin{algorithmic}
    \State Input: Indices $\tau_i$ for $i=1, 2, \dots, N$.
    \State Initialize: $\bs{x}\gets \bs{0}$, $\bs{x}^*\gets \bs{0}$, $\mathcal{R}^*\gets 0$.
    \State Sort $\tau_i$'s in a descending order $\delta$. 
    \For{$i = 1, 2, \dots, N$}
        \State Set $x_j \gets \1\{\tau_j \ge \tau_{\delta(i)}\}$ for $j=1, 2, \dots, N$;
        \If{$\bs{A}\bs{x} \le \bs{b}$ is violated}
        \State Continue;
        \EndIf
        \State Compute the expected revenue $\mathcal{R}\gets \mathcal{R}(\bs{x})$;
        \If{$\mathcal{R}> \mathcal{R}^*$}
        \State Set $\mathcal{R}^*\gets \mathcal{R}$ and $\bs{x}^*\gets \bs{x}$;
        \EndIf
    \EndFor
    \State Return: $\bs{x}^*$. 
    \end{algorithmic}
\end{breakablealgorithm}
\vspace{4mm}

\vspace{1mm}
\textbf{(2) GCN-based Index with Local Search (GILS) Policy.} 
To further enhance performance, we build on the local search algorithm of 
\citet{gallego2024efficient} (Algorithm~\ref{alg:local_search}) and propose a 
policy that begins with the GI solution and refines it through local addition, 
deletion, and swap operations. 
This hybrid approach, referred to as the GCN-based Index with Local Search 
(\textbf{GILS}) policy (Algorithm~\ref{alg:GCN_LS}), improves the GI solution by 
iteratively exploring neighboring assortments and retaining modifications that 
increase expected revenue.

\vspace{4mm}
\begin{breakablealgorithm}
    \caption{GCN-based index with local search (GILS) policy}\label{alg:GCN_LS}
    \small
    \begin{algorithmic}
    \State Input: Indices $p_j$ for $j=1, 2, \dots, N$.
    \State Implement the index policy in Algorithm~\ref{alg:index_policy} by plugging in $p_j$'s. The output is denoted by $\bs{x}_1$. 
    \State Implement the LS policy in Algorithm~\ref{alg:local_search} by plugging in $\bs{x}_1$. The output is denoted by $\bs{x}_2$
    \State Return: $\bs{x}_2$. 
    \end{algorithmic}
\end{breakablealgorithm}
\vspace{4mm}

\vspace{1mm}
\textbf{(3) GCN-based Integer Programming (GIP) Policy.} 
The previous two policies perform line or local search within the feasible 
space and may not fully exploit the problem structure under complex constraints. 
To address this, we propose an \textit{integer-programming-based} policy guided 
by the GCN indices (Algorithm~\ref{alg:IP_policy}). 
Specifically, the policy solves the following integer linear program:
\begin{align}\label{eq:ip_new}
    \max_{\bs{x}\in \{0, 1\}^N} &\quad \sum_{j\in [N]} p_j x_j \\
    \text{s.t.} &\quad \bs{A}\bs{x} \le \bs{b}. \nonumber
\end{align}
Compared with the original nonlinear integer program~\eqref{eq:main}, this 
formulation replaces the nonlinear objective with a linear one that can be 
solved efficiently using standard optimization solvers. 
Although \eqref{eq:ip_new} does not explicitly incorporate the choice model, the 
information learned by the GCN is embedded in the indices $p_j$. 
Moreover, if the constraint matrix $\bs{A}$ is totally unimodular, the integer 
constraints in \eqref{eq:ip_new} can be relaxed, and the problem can be solved 
as a linear program (see \citealt{wolsey1999integer}), resulting in a highly 
efficient policy. 
We refer to this approach as the GCN-based Integer Programming (\textbf{GIP}) 
policy.

\vspace{4mm}
\begin{breakablealgorithm}
    \caption{GCN-based integer programming (GIP) policy}\label{alg:IP_policy}
    \small
    \begin{algorithmic}
    \State Input: Probabilities $p_j$ for $j=1, 2, \dots, N$.
    \State Solve problem \eqref{eq:ip_new} with $p_j$'s and obtain the optimal solution $\bs{x}^*$.
    \State Return: $\bs{x}^*$. 
    \end{algorithmic}
\end{breakablealgorithm}
\vspace{4mm}

\begin{remark}
    If the constraints are not totally unimodular and the number of products is extremely large, we can further speed up the GIP policy by filtering out products with predicted probabilities less than 0.5 because these products are predicted to be out of the optimal assortment with strong confidence.
\end{remark}

\section{Theoretical Foundation}\label{sec:theory}

In this section, we provide some theoretical results to explain why the proposed GCN policies work well. We first show that the proposed GCN framework can represent the constrained MMNL assortment problem, and then establish its size generalization ability in a special case. 

Given an instance $\bs{I}=(\bs{\alpha}, \bs{r}, \bs{V}, \bs{A}, \bs{b})$, the graph input is the same as that in Section~\ref{subsec:graph_represent}, i.e., a weighted tripartite graph $G(\bs{I})=(\mathcal{N}, \mathcal{E}; \bs{Z})$, and let  $\mathcal{G}_{\mathrm{GCN}}$ denote the collection of all such tripartite graphs. The mapping from $\bs{I}$ to $G(\bs{I})$ is injective. As for the GCN, for ease of theoretical analysis, we generalize the network in Section~\ref{sec:GCN} and consider a family $\mathcal{F}_{\text{GCN}}^\sigma$ of finite-depth GCN. 

We start with a family $\mathcal{F}_{\text{GCN}}$ of finite-depth GCN without the last sigmoid layer. The computation of such a GCN with $L$ layers is as follows:
\begin{enumerate}
    \item Let $\bs{Y}^{S, 0}=\bs{Y}^S$, $\bs{Y}^{P, 0}=\bs{Y}^P$ and $\bs{Y}^{C, 0}=\bs{Y}^C$. The dimension of the $l$-th layer is $d_l$ with $d_0=d_N$ and $d_l=d_{\text{mid}}$ for $l=1, 2, \dots, L$. For each layer $l=1, 2, \dots, L$, we choose learnable functions $g_S^{(l)}, g_P^{(l)}, g_C^{(l)}: \mathbb{R}^{1\times d_{l-1}} \times \mathbb{R}^{1\times d_{l}} \to \mathbb{R}^{1\times d_{l}}$, and $f_{SP}^{(l)},f_{PS}^{(l)}, f_{PC}^{(l)}, f_{CP}^{(l)}: \mathbb{R}^{1 \times d_E} \times \mathbb{R}^{1 \times d_{l-1}}\to \mathbb{R}^{1 \times d_{l}}$ and update the feature vector of each node as
    \begin{align}\label{eq:finite_GCN_compute}
        &\bs{Y}^{S, l}_{k, :} = g_S^{(l)}\bigg(\bs{Y}^{S, l-1}_{k, :},\; \sum_{j=1}^N f_{PS}^{(l)}\big(\bs{Z}^S_{j, k,:}, \bs{Y}^{P, l-1}_{j, :} \big) \bigg) && \forall\; k\in [K] \nonumber\\
        &\bs{Y}^{P, l}_{j, :} = g_P^{(l)}\bigg(\bs{Y}^{P, l-1}_{j, :},\; \sum_{k=1}^K f_{SP}^{(l)}\big(\bs{Z}^S_{j, k,:}, \bs{Y}^{S, l-1}_{k, :} \big) + \sum_{i=1}^M f_{CP}^{(l)}\big(\bs{Z}^C_{i, j,:}, \bs{Y}^{C, l-1}_{i, :} \big)\bigg) && \forall\; j\in [N]\\
        &\bs{Y}^{C, l}_{i, :} = g_C^{(l)}\bigg(\bs{Y}^{C, l-1}_{i, :},\; \sum_{j=1}^M f_{PC}^{(l)}\big(\bs{Z}^C_{i, j, :}, \bs{Y}^{P, l-1}_{j, :} \big) \bigg) && \forall\; i\in [M].\nonumber
    \end{align}
    \item Then, we choose a learnable function $f_{\text{out}}: \mathbb{R}^{1\times d_\text{mid}}\to \mathbb{R}$ and compute the output of each product node as $s_j = f_{\text{out}}\left(\bs{Y}^{P, l}_{j, :}\right)$ for $j\in [N]$. 
\end{enumerate}

Following the setting in \cite{azizian2021expressive}, all learnable functions, $\{g_S^{(l)}, g_P^{(l)}, g_C^{(l)}\}_{l=1}^L$, $\{f_{SP}^{(l)}, $ $ f_{PS}^{(l)}, f_{PC}^{(l)}, f_{CP}^{(l)}\}_{l=1}^L$ and $f_{\text{out}}$, may take all continuous functions on given domains. According to \cite{cybenko1989approximation} and \cite{hornik1989multilayer}, they can be parametrized by multi-layer perceptrons (MLPs), i.e., multi-layer feedforward neural networks, which are capable of approximating any continuous function when the network is sufficiently dense e.g., either sufficiently deep or sufficiently wide. Let $\mathcal{F}_{\text{GCN}}$ denote the resulting family of GCN mappings $\Phi:\mathcal{G}_{\text{GCN}}\to \mathbb{R}^N$. The depth $L$ can be any finite positive integer. 

Let $\mathfrak{G}_S$, $\mathfrak{G}_P$ and $\mathfrak{G}_C$ denote all permutation functions of customer nodes, product nodes and constraint nodes, respectively. Due to the permutation-invariant property of \eqref{eq:finite_GCN_compute}, given $\gamma=(\pi_S, \pi_P, \pi_C)\in \Gamma:=\mathfrak{G}_S\times \mathfrak{G}_P\times \mathfrak{G}_C$, we have 
\begin{align*}
    \Phi(G)_j = \Phi\left(\gamma\cdot G \right)_{\pi_P(j)} \qquad \forall \; G\in \mathcal{G}_{\mathrm{GCN}}, \; \Phi\in \mathcal{F}_{\mathrm{GCN}}. 
\end{align*}

Then, we can define the family $\mathcal{F}_{\text{GCN}}^\sigma$ of finite-layer GCNs with a sigmoid layer as
\begin{align*}
    \mathcal{F}_{\text{GCN}}^\sigma = \left\{\sigma \odot \Phi: \Phi\in \mathcal{F}_{\text{GCN}} \right\},
\end{align*}
where $\odot$ implies item-wise computations. In the theoretical analysis, we first analyze the property of $\mathcal{F}_{\text{GCN}}$ and then establish the property of $\mathcal{F}_{\text{GCN}}^\sigma$. 

\subsection{Expressiveness}
In this subsection, following the idea of \cite{azizian2021expressive} and \cite{chen2023on}, given a finite set of instances, we establish that GCN can approximate the optimal solution to the constrained MMNL assortment problem under some conditions. 

In order to analyze the expressive power of $\mathcal{F}_{\text{GCN}}$, the literature typically uses the \textit{Weisfeiler-Lehman} (WL) test (\citealt{weisfeiler1968reduction}), where each node is labeled with an embedding in a real linear space (which is referred to as ``color''). In the following, we propose a modified WL test algorithm for the constrained MMNL assortment problem based on the computations in \eqref{eq:finite_GCN_compute}. 

\vspace{4mm}
\begin{breakablealgorithm}
    \caption{WL Test for Constrained MMNL Assortment Problem (denoted by $\text{WL}_{\text{MMNL}}$)}\label{alg:WL_test}
    \small
    \begin{algorithmic}
    \State Input: A graph instance $\bs{I}$ and the number of GCN layers $L$.
    \State Initialize: $C^{S, 0}_k = \text{HASH}_{S, 0}\big(\bs{Y}_{k,:}^S\big)$, $C^{P, 0}_j = \text{HASH}_{P, 0}\big(\bs{Y}_{j,:}^P\big)$, and $C^{C, 0}_i = \text{HASH}_{C, 0}\big(\bs{Y}_{i,:}^C\big)$.
    \For{$l=1, 2, \dots, L$}
        \State $C^{S, l}_k = \text{HASH}_{S, l}\big( C^{S, l-1}_k, \sum_{j=1}^N\text{HASH}^E_{PS, l}\big(\bs{Z}^S_{j, k, :}, C^{P, l-1}_j\big)\big)$.
        \State $C^{P, l}_j = \text{HASH}_{P, l}\big( C^{P, l-1}_j, \sum_{k=1}^K\text{HASH}^E_{SP, l}\big(\bs{Z}^S_{j, k, :}, C^{S, l-1}_k \big)+ \sum_{i=1}^M\text{HASH}^E_{CP, l}\big(\bs{Z}^C_{i, j, :}, C^{C, l-1}_i\big)\big)$.
        \State $C^{C, l}_i = \text{HASH}_{C, l}\big( C^{C, l-1}_i, \sum_{j=1}^N\text{HASH}^E_{ PC, l}\big(\bs{Z}^C_{i, j, :}, C^{P, l-1}_j\big)\big)$.
    \EndFor
    \State Return: $\left(\big\{C_k^{S, L} \big\}_{k=1}^K, \big\{C_j^{P, L} \big\}_{j=1}^N, \big\{C_i^{C, L} \big\}_{i=1}^M \right)$. 
    \end{algorithmic}
\end{breakablealgorithm}
\vspace{4mm}

In each iteration of Algorithm~\ref{alg:WL_test}, nodes are assigned the same color if and only if they share identical current colors, neighborhood color distributions, and edge information. This refinement process is strictly monotonic, meaning nodes assigned distinct colors in previous iterations will remain distinct. The coloring has not yet stabilized if at least two nodes sharing the same previous color are assigned different colors in the current step. The algorithm is guaranteed to stabilize after at most $K+N+M$ iterations (\citealt{berkholz2017tight}) and hence we can restrict our focus to GCNs with $L\le K+N+M$ layers. In Algorithm~\ref{alg:WL_test}, the hash functions, $\left\{\text{HASH}_{\tau, l}:\tau\in \{S, P, C\}, l\in [L]\right\}$ and $\big\{ \text{HASH}^E_{\rho, l}: \rho\in\{PS, SP, CP, PC\}, l\in [L] \big\}$, are injective on their corresponding domains. Moreover, the edge hash functions, $\big\{ \text{HASH}^E_{\rho, l}: \rho\in\{PS, SP, CP, PC\}, l\in [L]\big\}$, should map the inputs (whose number is finite) into linearly independent outputs such that the summation of the hash colors is injective (see Appendix~\ref{append:hash_function} for details). Since the output of Algorithm~\ref{alg:WL_test} is a tuple of three multisets, the WL test is permutation-invariant. 

If $\text{WL}_{\text{MMNL}}(\bs{G})=\text{WL}_{\text{MMNL}}(\bs{G}')$ for any depth $L\in \mathbb{N}$ and any hash function choice, then $\bs{G}$ and $\bs{G}'$ are indistinguishable, which is denoted by $\bs{G}\sim \bs{G}'$. 
Following \cite{chen2023on}, we introduce the notion of ``unfoldability'', meaning that the graph input does not contain any perfectly symmetrical or structurally equivalent nodes. In this case, the WL test can assign a unique color to each node. In the following, we provide the specific definition. 
\begin{definition}[unfoldable instance]\label{def:unfoldable}
    $\bs{G}\in \mathcal{G}_{\text{GCN}}$ is unfoldable if there exists an admissible choice of hash functions and a depth $L$, such that for any $\tau\in \{S, P, C\}$ and any distinct nodes $u\neq u'$, we have $C_u^{\tau, L}\neq C_{u'}^{\tau, L}$. In other words, there exists a WL test that can separate all nodes simultaneously. Let $\mathcal{G}_{\text{GCN}}^U$ denote the set of unfoldable graphs. 
\end{definition}

In the following theorem, we establish that $\mathcal{F}_{\text{GCN}}^\sigma$ can approximate any finite set of unfoldable graphs to any accuracy. 
\begin{theorem}[Finite-sample representation]\label{thm:finite_represent}
    Let $\mathcal{D}\subseteq \mathcal{G}_{\text{GCN}}^U$ be any finite set of unfoldable MMNL instances. Let $\Phi_{\text{assort}}$ denote the mapping from graph inputs to canonical optimal assortments (defined in Appendix~\ref{append:canonical_assort}). For any $\varepsilon \in (0, 1/2)$, there exists $\Phi^\sigma \in \mathcal{F}_{\text{GCN}}^\sigma$ such that for any $\bs{G}\in \mathcal{D}$ and every $j\in [N]$, 
    \[
    \left|\Phi^\sigma(\bs{G})_j - \Phi_{\text{assort}}(\bs{G})_j \right| \le \epsilon,
    \]
    and consequently
    \[
    \1\left[\Phi^\sigma(\bs{G})_j>1/2\right] = \Phi_{\text{assort}}(\bs{G})_j.
    \]
\end{theorem}

By Theorem~\ref{thm:finite_represent}, given any finite set of unfoldable MMNL instances, there exists a finite-depth GCN in $\mathcal{F}_{\text{GCN}}^\sigma$ that can approximate the optimal solutions to any precision. It provides theoretical foundations for the expressiveness of the proposed GCN framework.

\subsection{Size Generalization}
In this subsection, we prove the optimality of the GI policy on varying-size problems for the unconstrained MNL assortment problem, i.e., $K=1$ and $M=0$. According to \cite{rusmevichientong2010dynamic}, the optimal solution can be depicted as 
\begin{align}\label{eq:opt_mnl}
    \bs{x}^*(\bs{G})_j = \1[r_j > \mathcal{R}^*(\bs{G})],
\end{align}
where $\mathcal{R}^*(\bs{G})=\max_{\bs{x}\in \{0, 1\}^N} \mathcal{R}_{\bs{G}}(\bs{x})$ is the optimal revenue given the graph input $\bs{G}$. The optimal solution is unique and hence the mapping $\Phi_{assort}:\mathcal{G}_{GCN}\to \{0, 1\}^N$ is well-defined.
In the following, we consider $\delta$-separable graph instances:
\begin{align*}
    \mathcal{G}_\delta = \left\{\bs{G}\in \mathcal{G}_{\text{GCN}}: K=1, M=0, \min_j\; |r_j-\mathcal{R}^*(\bs{G})|\ge \delta \right\}.
\end{align*}

\begin{theorem}\label{thm:mnl_prob_learning}
    Let $\mathcal{D}\subseteq \mathcal{G}_\delta$ be any finite set of $\delta$-separable graph instances.
    For any $\epsilon\in (0, 1/2)$, there exists a GCN $\Phi^\sigma\in \mathcal{F}^\sigma_{\text{GCN}}$ with $L=2$ whose output is 
    \begin{align}\label{eq:rev_prob}
        p_j = \sigma\bigg(\beta\cdot \bigg(r_j-\rho\bigg(\sum_{j=1}^N\zeta(r_j, v_j)\bigg)\bigg)\bigg),
    \end{align}
    where $\beta\ge (1/\delta)\log(1/\epsilon-1)$ and $\zeta(\cdot)$ and $\rho(\cdot)$ are some continuous functions such that $\rho(\sum_{j=1}^N\zeta(r_j, v_j))=\mathcal{R}^*(\bs{G})$ for $\bs{G}\in \mathcal{D}$. For any $\bs{G}\in \mathcal{D}$ and every $j\in [N]$, we have 
    \begin{align*}
        \left|\Phi^\sigma_{\text{GCN}}(\bs{G})_j - \Phi_{assort}(\bs{G})_j \right| \le \epsilon. 
    \end{align*}
    Moreover, the training loss tends to zero as $\beta$ tends to infinity.
\end{theorem}

By Theorem~\ref{thm:mnl_prob_learning}, there exists a two-layer GCN whose predicted probabilities are increasing in the product revenues, and its corresponding training loss can be made arbitrarily close to zero. This result should be interpreted as an expressiveness result rather than a convergence guarantee. Specifically, Theorem~\ref{thm:mnl_prob_learning} shows that the proposed GCN architecture is sufficiently expressive to represent a revenue-order index. Due to the nonconvexity of the GCN structure, the network obtained in practice may not exactly satisfy \eqref{eq:rev_prob}. However, due to the structure of the optimal solution $\eqref{eq:opt_mnl}$, the obtained network's output may approximately increase in product revenues when the training dataset is large and the training loss is small.

In the following corollary, we show the optimality of the GI policy in this case. 
\begin{corollary}\label{coro:GI_opt}
    Given the GCN in Theorem~\ref{thm:mnl_prob_learning}, the GI policy is optimal to unconstrained MNL assortment problems of any size. 
\end{corollary}

Given the GCN in Theorem~\ref{thm:mnl_prob_learning}, although $\rho(\sum_{j=1}^N\zeta(r_j, v_j))$ can be far from the optimal revenue $\mathcal{R}^*(\bs{G})$ when $\bs{G}\notin \mathcal{D}$, the GI policy is the same as the revenue-order policy, which is optimal for the unconstrained MNL assortment problem (\citealt{rusmevichientong2010dynamic}). Furthermore, as long as a GCN's predicted probabilities increase in product revenues, the GI policy is optimal to all unconstrained MNL assortment problems. 

To summarize, for unconstrained MNL assortment problems, the proposed GCN framework has the potential to learn the revenue-order pattern from small-scale instances. The GI policy can then exploit this learned pattern to recover optimal solutions for large-scale instances. This provides a theoretical foundation for the ``from small to large'' idea in this special case. For more general assortment problems, we conjecture that small- and large-scale instances still share similar structural patterns, which can also be learned by the proposed GCN framework and exploited by the proposed inference policies.

\section{Numerical Experiments}\label{sec:numerical}

In this section, we present the numerical experiments. All computations are 
performed on a machine equipped with a 2.0\,GHz Quad-Core Intel Core~i5 CPU 
and an NVIDIA Tesla~V100 GPU,\footnote{We use NVIDIA Tesla V100 GPUs because they are cost-effective and readily available in corporate environments. More advanced GPUs can substantially reduce the running times.} running Ubuntu~22.04. The implementation uses 
Gurobi~12.0, Python~3.10, PyTorch~2.8, and CUDA~12.8. Each message-passing 
layer in Figure~\ref{fig:GCN_network} is implemented using the 
\texttt{GENConv} layer from the \texttt{torch\_geometric} package. 

\subsection{Base Simulations}
We compare the proposed GCN-based policies with five benchmark policies 
described below (the details of the algorithms can be found in the Appendix). All policies are initialized with the empty assortment, which is always feasible, and return the empty set if no additional feasible solution is identified.
\vspace{1mm}
\begin{enumerate}
    \item \textsc{Revenue-order (RO) policy.} 
    This policy implements the index rule based on product prices 
    $r_j$. 

\vspace{1mm}
    \item \textsc{Local-search (LS) policy.} 
    Following Algorithm~\ref{alg:local_search}, we implement the local-search 
    procedure of \citet{gallego2024efficient} starting from an empty set. 
    The maximum number of removals per product is set to one, and an assortment 
    is updated only if the relative revenue improvement exceeds $0.1\%$. 
    The algorithm explores $\Theta(N^3)$ candidate assortments.  

\vspace{1mm}
    \item \textsc{Revenue-order with local search (ROLS) policy.} 
    As described in Algorithm~\ref{alg:rols}, we first apply the RO policy to 
    generate an initial assortment and then refine it using the LS algorithm.  

\vspace{1mm}
    \item \textsc{Revenue-based integer programming (RP) policy.} 
    As specified in Algorithm~\ref{alg:rp}, we replace the GCN-derived indices 
    in Algorithm~\ref{alg:IP_policy} with the revenues $r_j$ and solve the 
    resulting integer linear program.  

\vspace{1mm}
    \item \textsc{Conic integer programming (CONIC) policy.} 
    Based on the approach of \citet{csen2018conic}, this method is 
    significantly faster than the standard MILP formulation of 
    \citet{mendez2014branch}. For each instance, we first solve $N$ linear 
    programs to compute McCormick estimators that tighten the feasible space. 
    We then solve the resulting conic integer program with a MIPGap threshold 
    of $0.1\%$, so that the algorithm terminates once the relaxed gap is below 
    this level. To prevent excessive computation time, we also impose a time 
    limit of 600~seconds on the conic solver. Although the Gurobi time limit is set to 600 seconds, the reported runtime may exceed this limit because Gurobi performs additional computations required to terminate the optimization and report solution attributes.
\end{enumerate}

\begin{remark}[Role of GCN-based indices]
The policies can be grouped into three pairs: (GI, RO), (GILS, ROLS), and (GIP, RP). The two policies in each pair share the same algorithmic structure but differ in the indices used for decision making: one uses GCN-based indices, whereas the other uses revenue-based indices. Therefore, comparisons within each pair isolate the value of the GCN-based indices while holding the inference approach fixed.
\end{remark}
\vspace{0.1in}

\tb{Solution sample generation.} 
Let $\mathcal{U}[a, b]$ denote the uniform distribution over the interval 
$[a, b]$, and let $\mathcal{U_D}[N]$ denote the discrete uniform distribution 
over the set $[N]$. In the numerical experiments, given the number of products 
$N$, the number of constraints $M$, and the number of customer types $K$, we 
generate each problem instance as follows:
\vspace{1mm}
\begin{itemize}
    \item \textsc{Customer proportions.}  
    We first sample $K$ values $z_k$ from $\mathcal{U}[0, 1]$, and normalize 
    them to obtain the type-share vector 
    $\bs{\alpha} = (\alpha_1, \dots, \alpha_K)$, where 
    $\alpha_k = z_k / \sum_{k'=1}^K z_{k'}$. 

\vspace{1mm}
    \item \textsc{Product revenues and utilities.}  
    For each product $j \in [N]$ and customer type $k \in [K]$, we draw 
    revenues $r_j \sim \mathcal{U}[1, 2]$ and utilities 
    $u_{kj} \sim \mathcal{U}[0, 1]$, and compute
    $
        v_{kj} = \exp(u_{kj} - \eta r_j),
    $
    where $\eta = 3$ denotes the price-sensitivity parameter.  

\vspace{1mm}
    \item \textsc{Constraints.}  
    We include two types of linear constraints:
    \begin{itemize}
        \item \emph{$\lfloor M/2\rfloor$ capacity constraints:}  
        For each capacity constraint $i$, we sample 
        $A_{ij} \sim \mathcal{U}[0, 1]$ for all $j \in [N]$, and set 
        $b_i \sim \mathcal{U}[5, 10]$.  
        
        \item \emph{$\lceil M/2\rceil$ precedence constraints:}  
        For each precedence constraint $i$, we set $b_i = 0$, and randomly draw 
        two products $j_1, j_2 \sim \mathcal{U_D}[N]$.  
        If $j_1 \ne j_2$, set $A_{ij_1} = 1$, $A_{ij_2} = -1$, and 
        $A_{ij'} = 0$ for all $j' \in [N] \setminus \{j_1, j_2\}$;  
        otherwise, set $A_{ij} = 0$ for all $j \in [N]$.  
    \end{itemize}

\vspace{1mm}
    \item \textsc{Label generation.}  
    Lastly, for each generated instance, we compute the optimal assortment 
    $\bs{x}^*$ using the CONIC policy described in 
    Algorithm~\ref{alg:conic}.  
    The resulting sample is represented as
    $(\bs{\alpha}, \bs{r}, \bs{V}, \bs{A}, \bs{b}, \bs{x}^*)$, 
    where $\bs{V} = (v_{kj})_{k,j}$.
\end{itemize}

\vspace{1mm}
Before training, we first generate $2{,}000$ solution samples with 
$K_{\train}=10$ customer types, $N_{\train}=20$ products, and 
$M_{\train}=10$ constraints. The total sample generation time is 
808~seconds. Using Algorithm~\ref{alg:data_augment}, we augment the dataset 
to $10{,}000$ samples by generating four additional samples from each 
original instance. The augmented dataset is then randomly divided into 
a training set (80\%) and a validation set (20\%), ensuring no overlap 
between them.  

We train the GCN using the training set and select the final model based on 
its validation performance. The model is optimized using the Adam optimizer 
\citep{kingma2014adam} with a learning rate of $10^{-4}$ and a batch size of 
128. To mitigate randomness of training, we conduct ten independent trials with random 
seeds ranging from~1 to~10 
(see Appendix~\ref{append:ind_trial} for details). In each trial, the GCN is 
trained for up to 100~epochs with an early-stopping rule of patience~50. 
Specifically, if the best validation loss does not improve by at least~0.001 
for 50~epochs, training is terminated early. The model achieving the lowest 
validation loss is retained as the final version. The average training time 
across the ten trials is 114~seconds.  

For all proposed GCN-based policies, we report the average competitive ratios 
and standard deviations across both the ten trials and the testing dataset. 
Hence, the reported standard deviation reflects variation from both the 
training process and the randomness of the test instances.

\tb{Testing.} 
To evaluate the performance of different policies, we generate 
$100$ testing instances with $K_{\test}=10$ customer types, 
$N_{\test}=500$ or $2{,}000$ products, and $M_{\test}=10$ constraints. 
The performance of each policy is measured by its 
\emph{competitive ratio}, defined as the ratio between the expected 
revenue achieved by the policy and that obtained by the 
CONIC benchmark. The numerical results are summarized in 
Table~\ref{tab:comparison_general}.

\begin{table}[ht]
    \centering
    \caption{Performance of proposed policies and existing policies when $K_\train=K_\test=10$, $M_\train=M_\test=10$, $N_\train=20$ and $N_\test=500\mbox{ or } 2,000$.}
    \label{tab:comparison_general} 
    \begin{tabular}{@{} *{5}{c} @{}}
    \toprule
       \multicolumn{1}{c}{\multirow{2}{*}{Policies}} & \multicolumn{2}{c@{}}{$N_\test=500$} & \multicolumn{2}{c@{}}{$N_\test=2,000$} \\
       \cmidrule(l){2-3}  \cmidrule(l){4-5} 
         & Avg. Ratio (std) & Avg. Time (s) & Avg. Ratio (std) & Avg. Time (s)\\
        \midrule 
        GI & 87.8\% (5.2\%) & 0.02 & 86.4\% (5.6\%) & 0.15\\
        GILS & 95.1\% (1.9\%) & 11.37 & 93.9\% (2.1\%) & 114.5\\
        GIP & 96.1\% (1.9\%) & 2.80 & 96.0\% (1.8\%)& 7.07\\
        RO & 12.3\% (3.7\%) & 0.04 & 11.6\% (1.6\%) & 0.20\\
        LS & 83.2\% (4.7\%) & 5.71 & 78.7\% (4.7\%) & 36.23\\
        ROLS & 92.9\% (3.2\%) & 47.00 & 91.1\% (2.1\%) & 366.85\\
        RP & 62.3\% (6.1\%) & 2.37 & 65.3\% (5.2\%) & 4.62\\
        CONIC & 100.0\% (0.0\%) &  142.01 & 100.0\% (0.0\%) & 752.52\\
        \bottomrule
    \end{tabular}
\end{table}

According to Table~\ref{tab:comparison_general}, although the GCN is trained 
on small-scale data with only 20~products, the proposed policies achieve 
strong performance on large-scale instances with 500 or 2{,}000~products, 
demonstrating the remarkable generalization ability of the GCN. 
Specifically, the GI policy attains a competitive ratio above~86\% within one 
second, and its performance can be further improved to about~94\% when 
combined with local search, albeit with longer runtime. In comparison, the 
GIP policy delivers even better performance (over~96\%) within a few seconds, 
striking an effective balance between solution quality and computational 
efficiency. This advantage arises because the GIP policy explicitly accounts 
for feasibility constraints, while modern integer programming solvers (e.g., 
Gurobi) can solve such linear formulations efficiently. Therefore, when strict 
response-time requirements are imposed (e.g., one~second), the GI policy is 
recommended; when slightly more time is acceptable (e.g., up to one~minute), 
the GILS or GIP policy offers superior results. Moreover, in Appendix~\ref{append:shift}, we evaluate the performance of the proposed policies under a distribution shift.

Next, we assess the effectiveness of the GCN-generated indices by comparing 
groups of related policies. First, both the RO and GI policies perform line 
searches over the feasible space, yet the GI policy consistently achieves 
significantly better outcomes. This indicates that the GCN-derived indices 
effectively encode constraint information and thus guide the search along a 
more promising path. In contrast, the RO policy may select only a few large 
items or even fail to identify a feasible assortment under capacity and 
precedence constraints, leading to poor performance.  
Second, both the RP and GIP policies ignore the exact structure of the choice 
model and solve an integer linear program, but the GIP policy performs 
substantially better. This shows that the GCN indices also capture key demand 
patterns embedded in the choice model.  
Finally, the LS, ROLS, and GILS policies all adopt the local-search procedure 
of \citet{gallego2024efficient}, yet GILS outperforms the others because the 
GI policy provides a high-quality initial solution that accelerates and 
stabilizes the local improvement process.

\subsection{Impact of Instance Configuration}

In this subsection, we examine how different instance configurations affect 
policy performance, focusing on four key parameters: the number of products 
in training and testing samples ($N_{\train}$ and $N_{\test}$), the number of 
customer types in testing samples ($K_{\test}$), and the number of constraints 
($M_{\test}$). Unless otherwise stated, the default configuration is 
$K_{\train}=K_{\test}=10$, $M_{\train}=M_{\test}=10$, 
$N_{\train}=20$, and $N_{\test}=500$. For each configuration, we evaluate 
both the proposed and benchmark policies on 100~independent testing instances.

\vspace{1mm}
\noindent
\tb{Impact of product numbers $N_{\train}$ and $N_{\test}$.} 
We first study how the number of products in the training set, $N_{\train}$, 
affects the performance of the proposed policies. 
Figure~\ref{fig:training_N} reports the average competitive ratios as 
$N_{\train}$ varies. As shown in the figure, increasing $N_{\train}$ improves 
the performance of all GCN-based policies, though with diminishing returns. 
This pattern arises because larger training samples capture a richer variety 
of optimal-assortment structures, but beyond a certain point, most of the 
relevant patterns are already learned from moderate-size instances, leading 
to smaller incremental gains. 

\begin{figure}[ht]
    \centering
        \begin{tikzpicture}
            \begin{axis}[
                width=0.7\textwidth, height=6cm,     % size of the image
                xmin = 10,     % start the diagram at this x-coordinate
                xmax = 50,    % end   the diagram at this x-coordinate
                scaled x ticks = false, 
                xticklabel style={
                    /pgf/number format/fixed,          
                    /pgf/number format/precision=2,    
                },
                % ymin = 0,     % start the diagram at this y-coordinate
                ymax = 1,   % end   the diagram at this y-coordinate
                % /pgfplots/xtick = {0,5,...,60}, % make steps of length 5
                % xticks = {50, 100, 150, 200, 250, 300, 350, 400, 450, 500},
                % extra y ticks = {0.507297},
                axis background/.style = {fill=white},
                ylabel = {Competitive Ratio},
                xlabel = {$N_\train$},
                legend pos= outer north east,
                tick align = outside,
                yticklabel={\pgfmathparse{\tick*100}\pgfmathprintnumber{\pgfmathresult}\%}
                ]
        
                % import the correct data from a CSV file
                \addplot[orange, mark=o,thick] table[x=Ntrain, y=GI, col sep=comma]{GCN_different_Ntrain.csv};
                \addlegendentry{GI}
                \addplot[name path=U1, draw=none]table[x=Ntrain, y expr=\thisrow{GI}+\thisrow{GI-std}, col sep=comma, forget plot]{GCN_different_Ntrain.csv};
                \addplot[name path=L1, draw=none, forget plot]table[x=Ntrain, y expr=\thisrow{GI}-\thisrow{GI-std}, col sep=comma]{GCN_different_Ntrain.csv};
                \addplot[fill=orange!60, fill opacity=0.25, forget plot] fill between[of=U1 and L1];
                
                \addplot[blue, mark=square,thick] table[x=Ntrain, y=GILS, col sep=comma]{GCN_different_Ntrain.csv};
                \addlegendentry{GILS}
                \addplot[name path=U2, draw=none]table[x=Ntrain, y expr=\thisrow{GILS}+\thisrow{GILS-std}, col sep=comma, forget plot]{GCN_different_Ntrain.csv};
                \addplot[name path=L2, draw=none, forget plot]table[x=Ntrain, y expr=\thisrow{GILS}-\thisrow{GILS-std}, col sep=comma]{GCN_different_Ntrain.csv};
                \addplot[fill=blue!60, fill opacity=0.25, forget plot] fill between[of=U2 and L2];
                
                \addplot[red, mark=triangle,thick] table[x=Ntrain, y=GIP, col sep=comma]{GCN_different_Ntrain.csv};
                \addlegendentry{GIP}
                \addplot[name path=U3, draw=none]table[x=Ntrain, y expr=\thisrow{GIP}+\thisrow{GIP-std}, col sep=comma, forget plot]{GCN_different_Ntrain.csv};
                \addplot[name path=L3, draw=none, forget plot]table[x=Ntrain, y expr=\thisrow{GIP}-\thisrow{GIP-std}, col sep=comma]{GCN_different_Ntrain.csv};
                \addplot[fill=red!60, fill opacity=0.25, forget plot] fill between[of=U3 and L3];
                
            \end{axis} 
        \end{tikzpicture}
    \caption{Average competitive ratios of proposed policies as functions of training instance size $N_\train$ when $K_\train=K_\test=10$, $M_\train=M_\test=10$, and $N_\test=500$. The shaded areas represent the standard deviation.}
    \label{fig:training_N}
    \end{figure}

\begin{figure}[ht]
    \centering
        \begin{tikzpicture}

  \begin{groupplot}[
    group style={group size=2 by 1, horizontal sep=1.6cm},
    width=0.48\textwidth, height=6cm,
    xmin=500, xmax=2000,
    scaled x ticks=false,
    xticklabel style={/pgf/number format/fixed,/pgf/number format/precision=2},
    xtick={500, 750, 1000, 1250, 1500, 1750, 2000},
    axis background/.style={fill=white},
    xlabel={$N_\test$},
    tick align=outside,
    yticklabel={\pgfmathparse{\tick*100}\pgfmathprintnumber{\pgfmathresult}\%},
    legend to name=seglegend,          
    legend columns=7,
    legend style={/tikz/every even column/.append style={column sep=0.8em}}
  ]

    \nextgroupplot[
      title={(a) GCN policies},
      title style={at={(axis description cs:0.5, -0.35)}, anchor=north},
      ylabel={Competitive Ratio},
      ymin=0.5,
      ymax=1.0
    ]
      % GI
      \addplot[orange, mark=o, thick]
        table[x=Nt, y=GI, col sep=comma]{GCN_different_N.csv};
      \addlegendentry{GI}
      \addplot[name path=U1, draw=none, forget plot]
        table[x=Nt, y expr=\thisrow{GI}+\thisrow{GI-std}, col sep=comma]{GCN_different_N.csv};
      \addplot[name path=L1, draw=none, forget plot]
        table[x=Nt, y expr=\thisrow{GI}-\thisrow{GI-std}, col sep=comma]{GCN_different_N.csv};
      \addplot[fill=orange!60, fill opacity=0.25, forget plot]
        fill between[of=U1 and L1];

      % GILS
      \addplot[blue, mark=square, thick]
        table[x=Nt, y=GILS, col sep=comma]{GCN_different_N.csv};
      \addlegendentry{GILS}
      \addplot[name path=U2, draw=none, forget plot]
        table[x=Nt, y expr=\thisrow{GILS}+\thisrow{GILS-std}, col sep=comma]{GCN_different_N.csv};
      \addplot[name path=L2, draw=none, forget plot]
        table[x=Nt, y expr=\thisrow{GILS}-\thisrow{GILS-std}, col sep=comma]{GCN_different_N.csv};
      \addplot[fill=blue!60, fill opacity=0.25, forget plot]
        fill between[of=U2 and L2];

      % GIP
      \addplot[red, mark=triangle, thick]
        table[x=Nt, y=GIP, col sep=comma]{GCN_different_N.csv};
      \addlegendentry{GIP}
      \addplot[name path=U3, draw=none, forget plot]
        table[x=Nt, y expr=\thisrow{GIP}+\thisrow{GIP-std}, col sep=comma]{GCN_different_N.csv};
      \addplot[name path=L3, draw=none, forget plot]
        table[x=Nt, y expr=\thisrow{GIP}-\thisrow{GIP-std}, col sep=comma]{GCN_different_N.csv};
      \addplot[fill=red!60, fill opacity=0.25, forget plot]
        fill between[of=U3 and L3];

    \nextgroupplot[
      title={(b) Benchmark policies},
      title style={at={(axis description cs:0.5, -0.35)}, anchor=north},
      ylabel={},
      ymin=0.5, ymax=1
    ]

    \addlegendimage{orange, mark=o, thick}
      \addlegendentry{GI}
      \addlegendimage{blue, mark=square, thick}
      \addlegendentry{GILS}
      \addlegendimage{red, mark=triangle, thick}
      \addlegendentry{GIP}

      % LS
      \addplot[magenta, mark=diamond, thick]
        table[x=N, y=LS, col sep=comma]{different_N.csv};
      \addlegendentry{LS}
      \addplot[name path=U4, draw=none, forget plot]
        table[x=N, y expr=\thisrow{LS}+\thisrow{LS-std}, col sep=comma]{different_N.csv};
      \addplot[name path=L4, draw=none, forget plot]
        table[x=N, y expr=\thisrow{LS}-\thisrow{LS-std}, col sep=comma]{different_N.csv};
      \addplot[fill=magenta!60, fill opacity=0.25, forget plot]
        fill between[of=U4 and L4];

      % ROLS
      \addplot[cyan, mark=star, thick]
        table[x=N, y=ROLS, col sep=comma]{different_N.csv};
      \addlegendentry{ROLS}
      \addplot[name path=U5, draw=none, forget plot]
        table[x=N, y expr=\thisrow{ROLS}+\thisrow{ROLS-std}, col sep=comma]{different_N.csv};
      \addplot[name path=L5, draw=none, forget plot]
        table[x=N, y expr=\thisrow{ROLS}-\thisrow{ROLS-std}, col sep=comma]{different_N.csv};
      \addplot[fill=cyan!60, fill opacity=0.25, forget plot]
        fill between[of=U5 and L5];

      % RP
      \addplot[violet, mark=pentagon, thick]
        table[x=N, y=RP, col sep=comma]{different_N.csv};
      \addlegendentry{RP}
      \addplot[name path=U6, draw=none, forget plot]
        table[x=N, y expr=\thisrow{RP}+\thisrow{RP-std}, col sep=comma]{different_N.csv};
      \addplot[name path=L6, draw=none, forget plot]
        table[x=N, y expr=\thisrow{RP}-\thisrow{RP-std}, col sep=comma]{different_N.csv};
      \addplot[fill=violet!60, fill opacity=0.25, forget plot]
        fill between[of=U6 and L6];

  \end{groupplot}

  %======= 底部共享 legend（整组图下方居中）=======
  \node at ([yshift=-5mm]current bounding box.south)
    {\pgfplotslegendfromname{seglegend}};

\end{tikzpicture}
    \caption{Average competitive ratios of different policies as functions of product number $N_\test$ when $K_\train=K_\test=10$, $M_\train=M_\test=10$, and $N_\train=20$. The shaded areas represent the standard deviation.}
    \label{fig:product_num}
    \end{figure}

Next, we study the impact of the number of products in the testing samples, 
denoted by $N_{\test}$, as shown in Figure~\ref{fig:product_num}. 
The figure shows that the performance of both the proposed and benchmark 
policies remains relatively stable across different values of $N_{\test}$. 
This suggests that optimal assortments across instances with varying product 
sizes exhibit similar structural patterns, which the GCN successfully captures. 
The strong generalization ability observed here can be attributed to the 
diversity of training samples with different $N_{\train}$ values, enabling 
the network to learn representations that transfer effectively to larger 
instances.  

\vspace{1mm}
\noindent
\tb{Impact of number of customer types $K_{\test}$.} 
We next examine how policy performance changes as the number of customer types 
$K_{\test}$ in the testing samples varies, as illustrated in 
Figure~\ref{fig:segment_num}.  

\begin{figure}[ht]
\centering
\begin{tikzpicture}

  \begin{groupplot}[
    group style={group size=2 by 1, horizontal sep=1.6cm},
    width=0.48\textwidth, height=6cm,
    xmin=2, xmax=18,
    scaled x ticks=false,
    xticklabel style={/pgf/number format/fixed,/pgf/number format/precision=2},
    xtick={2,4,6,8,10,12,14,16,18},
    axis background/.style={fill=white},
    xlabel={$K_\test$},
    tick align=outside,
    yticklabel={\pgfmathparse{\tick*100}\pgfmathprintnumber{\pgfmathresult}\%},
    legend to name=seglegend,          
    legend columns=7,
    legend style={/tikz/every even column/.append style={column sep=0.8em}}
  ]

    \nextgroupplot[
      title={(a) GCN policies},
      title style={at={(axis description cs:0.5, -0.35)}, anchor=north},
      ylabel={Competitive Ratio},
      ymin=0.5,
      ymax=1
    ]
      % GI
      \addplot[orange, mark=o, thick]
        table[x=Kt, y=GI, col sep=comma]{GCN_different_K.csv};
      \addlegendentry{GI}
      \addplot[name path=U1, draw=none, forget plot]
        table[x=Kt, y expr=\thisrow{GI}+\thisrow{GI-std}, col sep=comma]{GCN_different_K.csv};
      \addplot[name path=L1, draw=none, forget plot]
        table[x=Kt, y expr=\thisrow{GI}-\thisrow{GI-std}, col sep=comma]{GCN_different_K.csv};
      \addplot[fill=orange!60, fill opacity=0.25, forget plot]
        fill between[of=U1 and L1];

      % GILS
      \addplot[blue, mark=square, thick]
        table[x=Kt, y=GILS, col sep=comma]{GCN_different_K.csv};
      \addlegendentry{GILS}
      \addplot[name path=U2, draw=none, forget plot]
        table[x=Kt, y expr=\thisrow{GILS}+\thisrow{GILS-std}, col sep=comma]{GCN_different_K.csv};
      \addplot[name path=L2, draw=none, forget plot]
        table[x=Kt, y expr=\thisrow{GILS}-\thisrow{GILS-std}, col sep=comma]{GCN_different_K.csv};
      \addplot[fill=blue!60, fill opacity=0.25, forget plot]
        fill between[of=U2 and L2];

      % GIP
      \addplot[red, mark=triangle, thick]
        table[x=Kt, y=GIP, col sep=comma]{GCN_different_K.csv};
      \addlegendentry{GIP}
      \addplot[name path=U3, draw=none, forget plot]
        table[x=Kt, y expr=\thisrow{GIP}+\thisrow{GIP-std}, col sep=comma]{GCN_different_K.csv};
      \addplot[name path=L3, draw=none, forget plot]
        table[x=Kt, y expr=\thisrow{GIP}-\thisrow{GIP-std}, col sep=comma]{GCN_different_K.csv};
      \addplot[fill=red!60, fill opacity=0.25, forget plot]
        fill between[of=U3 and L3];

    \nextgroupplot[
      title={(b) Benchmark policies},
      title style={at={(axis description cs:0.5, -0.35)}, anchor=north},
      ylabel={},
      ytick={},
      ymin=0.5, ymax=1
    ]

    \addlegendimage{orange, mark=o, thick}
      \addlegendentry{GI}
      \addlegendimage{blue, mark=square, thick}
      \addlegendentry{GILS}
      \addlegendimage{red, mark=triangle, thick}
      \addlegendentry{GIP}

      % LS
      \addplot[magenta, mark=diamond, thick]
        table[x=K, y=LS, col sep=comma]{different_K.csv};
      \addlegendentry{LS}
      \addplot[name path=U4, draw=none, forget plot]
        table[x=K, y expr=\thisrow{LS}+\thisrow{LS-std}, col sep=comma]{different_K.csv};
      \addplot[name path=L4, draw=none, forget plot]
        table[x=K, y expr=\thisrow{LS}-\thisrow{LS-std}, col sep=comma]{different_K.csv};
      \addplot[fill=magenta!60, fill opacity=0.25, forget plot]
        fill between[of=U4 and L4];

      % ROLS
      \addplot[cyan, mark=star, thick]
        table[x=K, y=ROLS, col sep=comma]{different_K.csv};
      \addlegendentry{ROLS}
      \addplot[name path=U5, draw=none, forget plot]
        table[x=K, y expr=\thisrow{ROLS}+\thisrow{ROLS-std}, col sep=comma]{different_K.csv};
      \addplot[name path=L5, draw=none, forget plot]
        table[x=K, y expr=\thisrow{ROLS}-\thisrow{ROLS-std}, col sep=comma]{different_K.csv};
      \addplot[fill=cyan!60, fill opacity=0.25, forget plot]
        fill between[of=U5 and L5];

      % RP
      \addplot[violet, mark=pentagon, thick]
        table[x=K, y=RP, col sep=comma]{different_K.csv};
      \addlegendentry{RP}
      \addplot[name path=U6, draw=none, forget plot]
        table[x=K, y expr=\thisrow{RP}+\thisrow{RP-std}, col sep=comma]{different_K.csv};
      \addplot[name path=L6, draw=none, forget plot]
        table[x=K, y expr=\thisrow{RP}-\thisrow{RP-std}, col sep=comma]{different_K.csv};
      \addplot[fill=violet!60, fill opacity=0.25, forget plot]
        fill between[of=U6 and L6];

  \end{groupplot}

  \node at ([yshift=-5mm]current bounding box.south)
    {\pgfplotslegendfromname{seglegend}};

\end{tikzpicture}

\caption{Average competitive ratios of different policies as functions of $K_\test$ when $K_\train=10$, $M_\train=M_\test=10$, $N_\train=20$, and $N_\test=500$. The shaded areas represent the standard deviation.}
\label{fig:segment_num}
\end{figure}

According to Figure~\ref{fig:segment_num}, the proposed policies maintain 
strong performance, indicating that the GCN effectively captures the 
underlying choice model and can provide high-quality predictions even when 
the number of customer types changes. The performance deteriorates only 
slightly as the difference $|K_{\test} - K_{\train}|$ increases.  

\vspace{1mm}
\noindent
\tb{Impact of constraint number $M_{\test}$.} 
Figure~\ref{fig:constraint_num} shows how policy performance varies with the 
number of constraints in the testing samples. Both the GI and GILS policies 
are somewhat sensitive to changes in $M_{\test}$, particularly when the number 
of constraints decreases. Two factors explain this behavior.  
First, since the training instances are generated with a fixed 
$M_{\train}$, the GCN does not learn patterns corresponding to different 
constraint counts.  
Second, a smaller $M_{\test}$ enlarges the feasible region, making it harder 
for the line-search-based policies to locate high-quality solutions and more 
likely to encounter local optima. A similar pattern is also observed for the 
LS and ROLS benchmarks.  

In contrast, the GIP policy exhibits much greater stability because it 
explicitly incorporates the constraints while still leveraging GCN-generated 
indices that capture demand-side information. Although the RP policy also 
accounts for the constraints, its performance deteriorates markedly as 
$N_{\test}$ increases, since it ignores the interaction between the choice 
model and the constraint structure.

\begin{figure}[ht]
    \centering
        \begin{tikzpicture}

  \begin{groupplot}[
    group style={group size=2 by 1, horizontal sep=1.6cm},
    width=0.48\textwidth, height=6cm,
    xmin=2, xmax=18,
    scaled x ticks=false,
    xticklabel style={/pgf/number format/fixed,/pgf/number format/precision=2},
    xtick={2,4,6,8,10,12,14,16,18},
    axis background/.style={fill=white},
    xlabel={$M_\test$},
    tick align=outside,
    yticklabel={\pgfmathparse{\tick*100}\pgfmathprintnumber{\pgfmathresult}\%},
    legend to name=seglegend,          
    legend columns=7,
    legend style={/tikz/every even column/.append style={column sep=0.8em}}
  ]

    \nextgroupplot[
      title={(a) GCN policies},
      title style={at={(axis description cs:0.5, -0.35)}, anchor=north},
      ylabel={Competitive Ratio},
      ymin=0.45,
      ymax=1.0
    ]
      % GI
      \addplot[orange, mark=o, thick]
        table[x=Mt, y=GI, col sep=comma]{GCN_different_M.csv};
      \addlegendentry{GI}
      \addplot[name path=U1, draw=none, forget plot]
        table[x=Mt, y expr=\thisrow{GI}+\thisrow{GI-std}, col sep=comma]{GCN_different_M.csv};
      \addplot[name path=L1, draw=none, forget plot]
        table[x=Mt, y expr=\thisrow{GI}-\thisrow{GI-std}, col sep=comma]{GCN_different_M.csv};
      \addplot[fill=orange!60, fill opacity=0.25, forget plot]
        fill between[of=U1 and L1];

      % GILS
      \addplot[blue, mark=square, thick]
        table[x=Mt, y=GILS, col sep=comma]{GCN_different_M.csv};
      \addlegendentry{GILS}
      \addplot[name path=U2, draw=none, forget plot]
        table[x=Mt, y expr=\thisrow{GILS}+\thisrow{GILS-std}, col sep=comma]{GCN_different_M.csv};
      \addplot[name path=L2, draw=none, forget plot]
        table[x=Mt, y expr=\thisrow{GILS}-\thisrow{GILS-std}, col sep=comma]{GCN_different_M.csv};
      \addplot[fill=blue!60, fill opacity=0.25, forget plot]
        fill between[of=U2 and L2];

      % GIP
      \addplot[red, mark=triangle, thick]
        table[x=Mt, y=GIP, col sep=comma]{GCN_different_M.csv};
      \addlegendentry{GIP}
      \addplot[name path=U3, draw=none, forget plot]
        table[x=Mt, y expr=\thisrow{GIP}+\thisrow{GIP-std}, col sep=comma]{GCN_different_M.csv};
      \addplot[name path=L3, draw=none, forget plot]
        table[x=Mt, y expr=\thisrow{GIP}-\thisrow{GIP-std}, col sep=comma]{GCN_different_M.csv};
      \addplot[fill=red!60, fill opacity=0.25, forget plot]
        fill between[of=U3 and L3];

    \nextgroupplot[
      title={(b) Benchmark policies},
      title style={at={(axis description cs:0.5, -0.35)}, anchor=north},
      ylabel={},
      ymin=0.45, ymax=1
    ]

    \addlegendimage{orange, mark=o, thick}
      \addlegendentry{GI}
      \addlegendimage{blue, mark=square, thick}
      \addlegendentry{GILS}
      \addlegendimage{red, mark=triangle, thick}
      \addlegendentry{GIP}

      % LS
      \addplot[magenta, mark=diamond, thick]
        table[x=M, y=LS, col sep=comma]{different_M.csv};
      \addlegendentry{LS}
      \addplot[name path=U4, draw=none, forget plot]
        table[x=M, y expr=\thisrow{LS}+\thisrow{LS-std}, col sep=comma]{different_M.csv};
      \addplot[name path=L4, draw=none, forget plot]
        table[x=M, y expr=\thisrow{LS}-\thisrow{LS-std}, col sep=comma]{different_M.csv};
      \addplot[fill=magenta!60, fill opacity=0.25, forget plot]
        fill between[of=U4 and L4];

      % ROLS
      \addplot[cyan, mark=star, thick]
        table[x=M, y=ROLS, col sep=comma]{different_M.csv};
      \addlegendentry{ROLS}
      \addplot[name path=U5, draw=none, forget plot]
        table[x=M, y expr=\thisrow{ROLS}+\thisrow{ROLS-std}, col sep=comma]{different_M.csv};
      \addplot[name path=L5, draw=none, forget plot]
        table[x=M, y expr=\thisrow{ROLS}-\thisrow{ROLS-std}, col sep=comma]{different_M.csv};
      \addplot[fill=cyan!60, fill opacity=0.25, forget plot]
        fill between[of=U5 and L5];

      % RP
      \addplot[violet, mark=pentagon, thick]
        table[x=M, y=RP, col sep=comma]{different_M.csv};
      \addlegendentry{RP}
      \addplot[name path=U6, draw=none, forget plot]
        table[x=M, y expr=\thisrow{RP}+\thisrow{RP-std}, col sep=comma]{different_M.csv};
      \addplot[name path=L6, draw=none, forget plot]
        table[x=M, y expr=\thisrow{RP}-\thisrow{RP-std}, col sep=comma]{different_M.csv};
      \addplot[fill=violet!60, fill opacity=0.25, forget plot]
        fill between[of=U6 and L6];

      % RO
      % \addplot[green, mark=otimes, thick]
      %   table[x=M, y=RO, col sep=comma]{different_M.csv};
      % \addlegendentry{RO}
      % \addplot[name path=U7, draw=none, forget plot]
      %   table[x=M, y expr=\thisrow{RO}+\thisrow{RO-std}, col sep=comma]{different_M.csv};
      % \addplot[name path=L7, draw=none, forget plot]
      %   table[x=M, y expr=\thisrow{RO}-\thisrow{RO-std}, col sep=comma]{different_M.csv};
      % \addplot[fill=green!60, fill opacity=0.25, forget plot]
      %   fill between[of=U7 and L7];

  \end{groupplot}

  \node at ([yshift=-5mm]current bounding box.south)
    {\pgfplotslegendfromname{seglegend}};

\end{tikzpicture}
    \caption{Average competitive ratios of different policies as functions of the number of constraints $M_\test$ when $K_\train=K_\test=10$, $M_\train=10$, $N_\train=20$, and $N_\test=500$. The shaded areas represent the standard deviation.}
    \label{fig:constraint_num}
    \end{figure}

\subsection{Time Limit for Policies}\label{subsec:time_limit}

To better balance computational efficiency and policy performance, we next 
examine the effect of imposing time limits on policy execution. We focus on 
the GILS, LS, ROLS, and CONIC policies, as the remaining methods complete 
within a few seconds and thus require no time restriction. 
Figure~\ref{fig:algorithm_with_limit} presents the average competitive ratios 
of these policies under different time limits.

\begin{figure}[ht]
\centering
\begin{tikzpicture}

  \begin{groupplot}[
    group style={group size=2 by 1, horizontal sep=1.6cm},
    width=0.48\textwidth, height=6cm,
    xmin=0, 
    scaled x ticks=false,
    xticklabel style={/pgf/number format/fixed,/pgf/number format/precision=2},
    axis background/.style={fill=white},
    xlabel={Time Limit},
    tick align=outside,
    yticklabel={\pgfmathparse{\tick*100}\pgfmathprintnumber{\pgfmathresult}\%},
    legend to name=seglegend,          
    legend columns=4,
    legend style={/tikz/every even column/.append style={column sep=0.8em}}
  ]

    \nextgroupplot[
      title={(a) GILS, LS, and ROLS},
      title style={at={(axis description cs:0.5, -0.35)}, anchor=north},
      ylabel={Competitive Ratio},
      xmax=180,
      ymin=0,
      ymax=1
    ]
                \addplot[blue, mark=square,thick, every nth point=10] table[x=time, y=GILS, col sep=comma]{new_GCN_time_limit.csv};
                \addlegendentry{GILS}
                \addplot[name path=U2, draw=none]table[x=time, y expr=\thisrow{GILS}+\thisrow{GILS-std}, col sep=comma, forget plot]{new_GCN_time_limit.csv};
                \addplot[name path=L2, draw=none, forget plot]table[x=time, y expr=\thisrow{GILS}-\thisrow{GILS-std}, col sep=comma]{new_GCN_time_limit.csv};
                \addplot[fill=blue!60, fill opacity=0.25, forget plot] fill between[of=U2 and L2];
                
                \addplot[magenta, mark=diamond,thick, every nth point=10] table[x=time, y=LS, col sep=comma]{time_limit.csv};
                \addlegendentry{LS}
                \addplot[name path=U5, draw=none]table[x=time, y expr=\thisrow{LS}+\thisrow{LS-std}, col sep=comma, forget plot]{time_limit.csv};
                \addplot[name path=L5, draw=none, forget plot]table[x=time, y expr=\thisrow{LS}-\thisrow{LS-std}, col sep=comma]{time_limit.csv};
                \addplot[fill=magenta!60, fill opacity=0.25, forget plot] fill between[of=U5 and L5];

                \addplot[cyan, mark=star, thick, every nth point=10]
                table[x=time, y=ROLS, col sep=comma]{time_limit.csv};
              \addlegendentry{ROLS}
              \addplot[name path=U6, draw=none, forget plot]
                table[x=time, y expr=\thisrow{ROLS}+\thisrow{ROLS-std}, col sep=comma]{time_limit.csv};
              \addplot[name path=L6, draw=none, forget plot]
                table[x=time, y expr=\thisrow{ROLS}-\thisrow{ROLS-std}, col sep=comma]{time_limit.csv};
              \addplot[fill=cyan!60, fill opacity=0.25, forget plot]
                fill between[of=U6 and L6];

    \nextgroupplot[
      title={(b) CONIC},
      title style={at={(axis description cs:0.5, -0.35)}, anchor=north},
      ylabel={},
      ytick={},
      xmax=500,
      ymin=0, ymax=1
    ]

      \addlegendimage{blue, mark=square, thick}
      \addlegendentry{GILS}
      \addlegendimage{magenta, mark=diamond, thick}
      \addlegendentry{LS}
      \addlegendimage{cyan, mark=star, thick}
      \addlegendentry{ROLS}

      \addplot[black, mark=oplus,thick] table[x=time, y=CONIC, col sep=comma, every nth point=3]{CONIC_time_limit.csv};
                \addplot[name path=U, draw=none]table[x=time, y expr=\thisrow{CONIC}+\thisrow{CONIC-std}, col sep=comma]{CONIC_time_limit.csv};
                \addplot[name path=L, draw=none]table[x=time, y expr=\thisrow{CONIC}-\thisrow{CONIC-std}, col sep=comma]{CONIC_time_limit.csv};
                \addplot[fill=black!60, fill opacity=0.25] fill between[of=U and L];
                \addlegendentry{CONIC}

  \end{groupplot}

  \node at ([yshift=-5mm]current bounding box.south)
    {\pgfplotslegendfromname{seglegend}};

\end{tikzpicture}

\caption{Average competitive ratios of different policies as functions of time limit when $K_\train=K_\test=10$, $M_\train=M_\test=10$, $N_\train=20$ and $N_\test=2000$. The shaded areas represent the standard deviation.}
\label{fig:algorithm_with_limit}
\end{figure}

According to Figure~\ref{fig:algorithm_with_limit}a, setting a time limit of 
15~seconds for the GILS policy yields a competitive ratio of approximately 
91\%. For the LS policy, a limit of 25~seconds results in a competitive ratio 
around 78\%. The ROLS policy requires substantially longer computation to 
reach high performance; a time limit of 180~seconds achieves roughly 83\%. 
Overall, when time limits are imposed, the GILS policy outperforms both the 
LS and ROLS policies in terms of the trade-off between solution quality and 
computational efficiency.  

Figure~\ref{fig:algorithm_with_limit}b further compares the CONIC benchmark. 
To attain the same level of performance as the proposed GCN-based policies, 
the CONIC method requires several minutes of computation. Specifically, 
achieving competitive ratios of 85\%, 90\%, and 95\% takes approximately 
318, 351, and 382~seconds, respectively. In contrast, the proposed GCN-based 
approaches reach comparable performance within only a few seconds, 
highlighting their substantial computational advantage.

\subsection{Learning from Suboptimal Solutions}

In this subsection, we explore whether suboptimal solutions can be used to 
train the GCN effectively, allowing for larger training instances without 
incurring excessive computational cost. As shown in 
Figure~\ref{fig:training_N}, increasing the number of products 
$N_{\train}$ in the training data improves the performance of the proposed 
policies, albeit at the expense of substantially longer computation times for 
obtaining optimal solutions. To address this trade-off, we train the GCN using 
suboptimal solutions as labels for large-scale instances. Specifically, during 
sample generation, we terminate the CONIC policy early once the MIPGap falls 
below a chosen threshold $\text{gap}_{\train}$ (e.g., 10\%) instead of the 
strict 0.1\% optimality requirement. Although the resulting labels are 
suboptimal, they may still capture sufficient structural information for the 
GCN to learn useful patterns in predicting product inclusion probabilities.  

In the numerical experiments, we generate 2{,}000 instances with 
$K_{\train}=10$ customer types, $N_{\train}=200$ products, and 
$M_{\train}=10$ constraints. For each MIPGap threshold 
$\text{gap}_{\train} \in \{1\%, 10\%, 20\%, 30\%, 40\%\}$, the instances are 
solved using the CONIC policy, with computation terminated once the MIPGap 
criterion is met. The resulting dataset is augmented to 10,000 samples, and then divided into an 80\% training set 
and a 20\% validation set. We then train the GCN (average training time: 
251~seconds) and evaluate it on 100 testing instances with 
$N_{\test}=500$ products, consistent with those used in 
Table~\ref{tab:comparison_general}. The numerical results are summarized in 
Table~\ref{tab:suboptimal}. Because the MIPGap represents an upper bound on 
the true optimality gap, we also report the corresponding average realized 
optimality gaps.

\begin{table}[ht]
    \centering
    \caption{Performance of proposed policies with GCN trained on 2,000 suboptimal samples with different stopping criterion $\mbox{gap}_\train$ when $K_\train=K_\test=10$, $M_\train=M_\test=10$, $N_\train=200$ and $N_\test=500$.}
    \label{tab:suboptimal} 
    \resizebox{\textwidth}{!}{
    \begin{tabular}{@{} *{9}{c} @{}}
        \toprule
        \multicolumn{1}{c}{\multirow{2}{*}{$\mbox{gap}_\train$}} & \multicolumn{1}{c}{\multirow{2}{*}{Avg. Gap}} & \multicolumn{1}{c}{\multirow{2}{*}{\makecell{Sample Generation \\ Time (s)}}} & \multicolumn{3}{c@{}}{Avg. Ratio (std)} & \multicolumn{3}{c@{}}{Runtime (s)}\\
        \cmidrule(l){4-6}  \cmidrule(l){7-9} 
        & & & GI & GILS & GIP & GI & GILS & GIP\\
        \midrule 
        1\% & 0.6\% & 25691 & 91.7\% (4.8\%) & 97.3\% (1.4\%) & 98.0\% (1.0\%) & 0.02 & 2.92 & 2.78\\
        10\% & 2.7\% & 8704 & 91.5\% (4.3\%) & 97.3\% (1.3\%) & 98.2\% (0.9\%) & 0.02 & 3.14 & 2.75\\
        20\% & 8.8\% & 6774 & 90.7\% (4.2\%) & 97.6\% (1.2\%) & 97.3\% (1.3\%) & 0.02 & 4.15 & 2.82\\
        30\% & 14.2\% & 5988 & 88.3\% (4.9\%) & 97.1\% (1.2\%) & 96.3\% (1.3\%) & 0.02 & 5.49 & 2.80\\
        40\% & 18.0\% & 5976 & 87.8\% (4.6\%) & 97.1\% (1.2\%) & 96.4\% (1.3\%) & 0.02 & 5.69 & 2.79\\
        \bottomrule
    \end{tabular}}
\end{table}

According to Table~\ref{tab:suboptimal}, even when trained on large-scale 
samples with suboptimal labels, the GCN continues to learn meaningful 
structural patterns and delivers strong performance. For instance, when the 
training samples have an average optimality gap of 8.8\%, the GI policy 
achieves a competitive ratio of approximately 90\%, exceeding the performance 
reported in Table~\ref{tab:comparison_general}. This result indicates that 
most suboptimal solutions still preserve key features of the true optima, 
allowing the GCN to extract and generalize useful patterns. Consequently, the 
proposed policies can leverage these learned representations to produce 
high-quality solutions efficiently, even when trained on imperfect data.

\section{Other Choice Models}\label{sec:other_choice}

In this section, we extend our framework to assortment optimization problems 
under several alternative choice models, including the nested logit (NL), 
exponomial, and Markov chain (MC) models.  

\vspace{1mm}
\textbf{Nested Logit (NL) model (\citealt{mcfadden1980econometric}).} 
In the NL model, products are organized into \emph{nests}, and customers make 
decisions in two sequential stages. In the first stage, a customer chooses one 
of the nests (or opts for no purchase). Conditional on choosing a nest, the 
customer then selects one product within that nest. Similar to 
Section~\ref{sec:model}, we consider $K$ latent customer types with the 
proportion vector $\bs{\alpha}$, where the standard NL model corresponds to 
the special case $K=1$. Let $v_{kj} := \exp(u_{kj} - \eta r_j)$ denote the attraction value of product 
$j$ for a type-$k$ customer, where $u_{kj}$ is the base utility and 
$\eta > 0$ is the price-sensitivity parameter. Each product belongs to exactly 
one nest, and we denote by $S_{\ell}$ the set of products in nest~$\ell$.  
Given $L$ nests $\{S_{\ell}\}_{\ell=1}^{L}$, the type-$k$ customer first 
chooses nest~$S_h$ with probability $Q_k(S_h \mid \bs{x})=\left( \sum_{j \in S_h} x_j v_{kj} \right)^{\gamma_h}/\left(1 + \sum_{\ell=1}^{L}
           \left( \sum_{j \in S_{\ell}} x_j v_{kj} \right)^{\gamma_{\ell}}\right)$
% \begin{align*}
%     Q_k(S_h \mid \bs{x}) 
%     = \frac{\left( \sum_{j \in S_h} x_j v_{kj} \right)^{\gamma_h}}
%            {1 + \sum_{\ell=1}^{L}
%            \left( \sum_{j \in S_{\ell}} x_j v_{kj} \right)^{\gamma_{\ell}}},
% \end{align*}
where $\gamma_{\ell}$ is the dissimilarity parameter for nest~$S_{\ell}$.  
Conditional on selecting nest~$S_h$, the probability of purchasing product 
$j \in S_h$ is $q_k(j \mid S_h, \bs{x}) 
    = x_j v_{kj}/\left(
           \sum_{j' \in S_h} x_{j'} v_{kj'}\right)$.
% \begin{align*}
%     q_k(j \mid S_h, \bs{x}) 
%     = \frac{x_j v_{kj}}
%            {\sum_{j' \in S_h} x_{j'} v_{kj'}}.
% \end{align*}
Combining the two stages, the probability that a type-$k$ customer purchases 
product~$j$ is  
\begin{align*}
    \psi_{kj}^{\mathrm{NL}}(\bs{x})
    &= \sum_{\ell=1}^{L} 
       Q_k(S_{\ell} \mid \bs{x}) \cdot \1\{j \in S_{\ell}\} 
       \cdot q_k(j \mid S_{\ell}, \bs{x}) = \sum_{\ell=1}^{L}
       \frac{\left( \sum_{j' \in S_{\ell}} x_{j'} v_{kj'} \right)^{\gamma_{\ell}}}
            {1 + \sum_{\ell'=1}^{L} 
              \left( \sum_{j'' \in S_{\ell'}} x_{j''} v_{kj''} 
              \right)^{\gamma_{\ell'}}}
       \cdot
       \frac{x_j v_{kj} \, \1\{j \in S_{\ell}\}}
            {\sum_{j' \in S_{\ell}} x_{j'} v_{kj'}}.
\end{align*}
The overall choice probability of product~$j$ across all customer types is then  
$\phi_j^{\mathrm{NL}}(\bs{x}) = \sum_{k=1}^{K} \alpha_k \cdot \psi_{kj}^{\mathrm{NL}}(\bs{x})$.  

\begin{figure}[ht]
    \centering
    \subfloat[NL]{
        \includegraphics[width=0.45\textwidth]{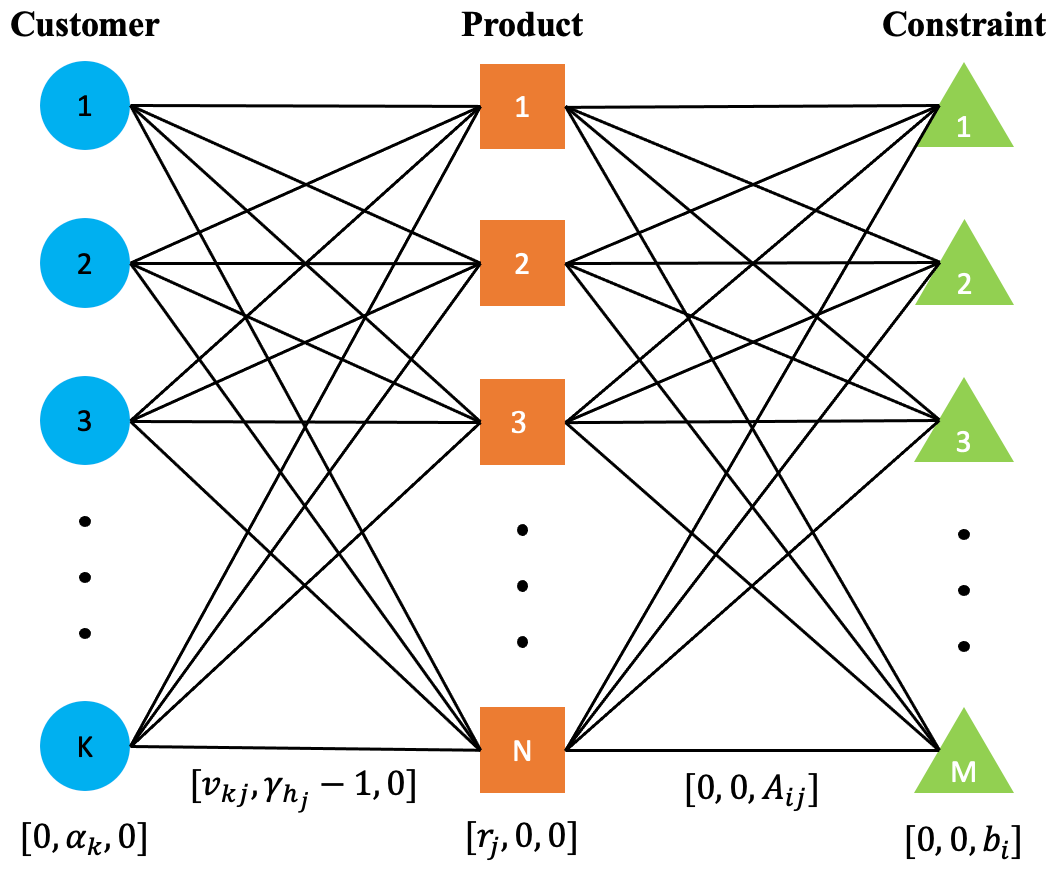}
        \label{fig:NL_graph}
    }
    \qquad 
    \subfloat[Exponomial]{
        \includegraphics[width=0.45\textwidth]{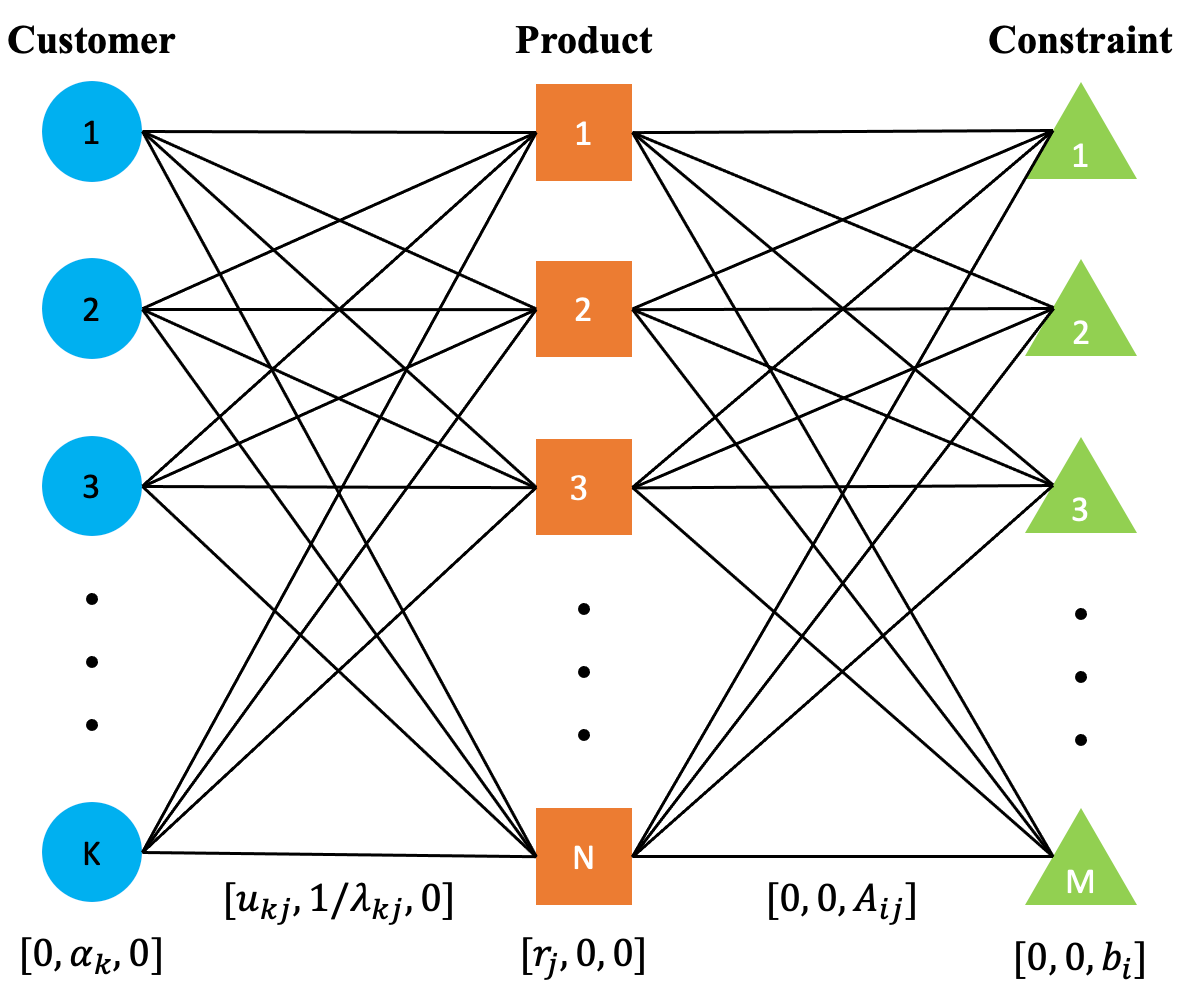}
        \label{fig:EXP_graph}
    }
    \caption{Graph representation of constrained assortment problem under NL and exponomial choice models and linear constraints $\bs{A}\bs{x}\le \bs{b}$.}
    \label{fig:NL_EXP_graphs}
\end{figure}

As shown by \citet{gallego2014constrained}, the assortment optimization 
problem under the NL model with capacity constraints is NP-hard. 
Figure~\ref{fig:NL_graph} illustrates the corresponding graph representation, 
which closely resembles that of the MMNL model in 
Figure~\ref{fig:MMNL_graph}. The feature vectors are defined as follows:  
the customer node~$k$ has features $[0,\, \alpha_k,\, 0]$; the product 
node~$j$ has features $[r_j,\, 0,\, 0]$; and the constraint node~$i$ has 
features $[0,\, 0,\, b_i]$.  
For edges, the feature vector between customer node~$k$ and product node~$j$ 
is $[v_{kj},\, \gamma_{h_j} - 1,\, 0]$, where $h_j$ is the index of the nest 
to which product~$j$ belongs. The edge between product node~$j$ and constraint 
node~$i$ has feature $[0,\, 0,\, A_{ij}]$.  

% \vspace{1mm}
\textbf{Exponomial model (\citealt{alptekinouglu2016exponomial}).} 
As in Section~\ref{sec:model}, we consider $K$ latent customer types with 
proportion vector $\bs{\alpha}$. For type-$k$ customers, the random utility of 
product~$j$ is given by  
$
    U_{kj} = u_{kj} - \eta r_j - \xi_{kj},
$
where $\eta$ is the price-sensitivity parameter and $\xi_{kj}$ are 
independent exponential random variables with rate $\lambda_{kj}$. 
We normalize $u_{k0} = 0$, $r_0 = 0$, and $U_{k0} = -\xi_{k0}$.  Given an assortment represented by $\bs{x}$, the choice probability that a 
type-$k$ customer purchases product~$j$ is  
\begin{align*}
    \psi_{kj}^{\mathrm{EXP}}(\bs{x})
    &= \frac{\lambda_{kj}}{L_{k,\sigma_k^{-1}(j)}}
       \exp\!\Bigg[
       -\!\sum_{i=\sigma_k^{-1}(j)}^{N+1}
         \lambda_{\sigma_k(i)}
         \big(u_{k,\sigma_k(i)} - \eta r_{\sigma_k(i)}
           - u_{kj} + \eta r_j \big)
       \Bigg] \\
    &\quad - \sum_{l=1}^{\sigma_k^{-1}(j)-1}
       \frac{\lambda_{k,\sigma_k(l)} \lambda_{kj}}
            {L_{k,\sigma_k(l)} L_{k,\sigma_k(l+1)}}
       \exp\!\Bigg[
       -\!\sum_{i=l}^{N+1}
         \lambda_{\sigma_k(i)}
         \big(u_{k,\sigma_k(i)} - \eta r_{\sigma_k(i)}
           - u_{k,\sigma_k(l)} + \eta r_{\sigma_k(l)} \big)
       \Bigg],
\end{align*}
where $\sigma_k$ denotes the ascending order of products (and the no-purchase 
option) according to $(u_{kj} - \eta r_j)$, and 
$L_{k,i} = \sum_{j=i}^{N+1} \lambda_{\sigma_k(j)}$. 

As shown by \citet{aouad2023exponomial}, the assortment optimization problem 
under the exponomial model with capacity constraints is APX-hard, while the 
complexity of the unconstrained case remains an open question. Figure~\ref{fig:EXP_graph} illustrates the graph representation under the EXP 
model. The overall structure parallels that of the MMNL representation in 
Figure~\ref{fig:MMNL_graph}, but the feature vectors differ. Specifically, the 
customer node~$k$ has feature vector $[0,\, \alpha_k,\, 0]$; the product 
node~$j$ has $[r_j,\, 0,\, 0]$; and the constraint node~$i$ has 
$[0,\, 0,\, b_i]$.  
For edges, the feature vector between customer node~$k$ and product node~$j$ 
is $[u_{kj},\, 1/\lambda_{kj},\, 0]$, while that between product node~$j$ and 
constraint node~$i$ is $[0,\, 0,\, A_{ij}]$.  
As in Figure~\ref{fig:MMNL_graph}, additional latent customer types can be 
modeled by introducing extra customer nodes.

\vspace{1mm}
\textbf{Markov Chain (MC) model (\citealt{blanchet2016markov}).} 
In the MC model, a fraction $\beta_j$ of customers initially prefer product~$j$, 
with $\sum_{j=1}^N \beta_j = 1$. If a customer whose most preferred product 
$j_1$ is unavailable, they switch to another product 
$j_2 \in [N] \setminus \{j_1\}$ with probability $\rho_{j_1 j_2}$ or choose the 
no-purchase option with probability 
$1 - \sum_{j' \in [N]} \rho_{j_1 j'}$.  
If the new product $j_2$ is also unavailable, the customer again transitions 
to another product $j_3 \in [N] \setminus \{j_2\}$ with probability 
$\rho_{j_2 j_3}$, or exits the market with probability 
$1 - \sum_{j' \in [N]} \rho_{j_2 j'}$.  
This process continues until the customer either switches to an available 
product—resulting in a purchase—or reaches the no-purchase option, thereby 
leaving the market.  Since each transition depends only on the most recent unavailable product, 
the switching process follows a Markov chain.  
Let $\bs{\beta}_S$ denote the vector of customer proportions whose preferred 
products lie in $S$, and let $\bs{\rho}_S$ denote the transition matrix 
$\bs{\rho}$ with all rows indexed by $S$ replaced by zeros.  
Given an assortment $\bs{x}$, the purchase probability of product~$j$ is 
obtained by solving the following system of linear equations:
%\begin{align*}
 $   \phi_j^{\mathrm{MC}}(\bs{x})
    = x_j \cdot \bs{e}_j^{\mathsf{T}}
      ( \bs{I} - \bs{\rho}_{S_{\bs{x}}}^{\mathsf{T}})^{-1}
      \bs{\beta} $,
%\end{align*}
where $\bs{e}_j$ is the $j$-th standard basis vector in $\mathbb{R}^N$.  

\begin{figure}[h]
    \centering
    \includegraphics[width=0.45\textwidth]{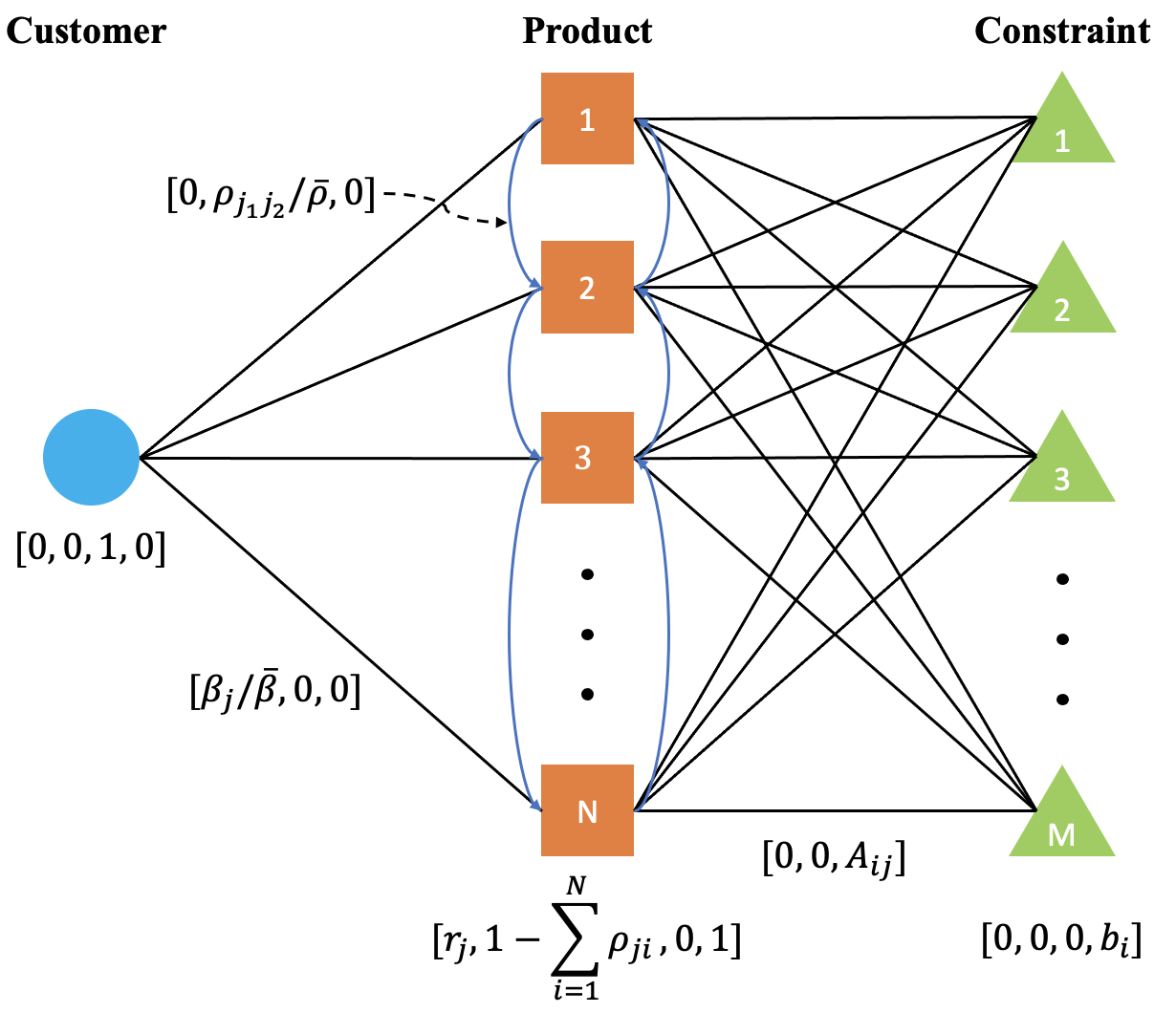}    
    \caption{Graph representation of constrained assortment problem under MC choice model and linear constraints $\bs{A}\bs{x}\le \bs{b}$.}
    \label{fig:MC_graph}
\end{figure}

As shown by \citet{desir2020constrained}, under capacity constraints, the 
assortment optimization problem with the MC model is APX-hard.  Figure~\ref{fig:MC_graph} presents the graph representation for the MC model.  
Relative to Figure~\ref{fig:MMNL_graph}, additional product-to-product edges 
(blue directed edges in Figure~\ref{fig:MC_graph}) are introduced to capture 
transition probabilities between products.  
In constructing the graph, we normalize the parameters by 
$\bar{\rho} := \max_{j_1, j_2} \rho_{j_1 j_2}$ and 
$\bar{\beta} := \max_j \beta_j$.  
The feature vectors are specified as follows:  
the single customer node has feature vector $[0,\, 0,\, 1,\, 0]$;  
each product node~$j$ has 
$[r_j,\, 1 - \sum_{i=1}^N \rho_{ji},\, 0,\, 0]$;  
and each constraint node~$i$ has $[0,\, 0,\, 0,\, b_i]$.  
For the edges, the feature vector from the customer node to product node~$j$ 
is $[\beta_j / \bar{\beta},\, 0,\, 0]$;  
the directed edge from product~$j_1$ to product~$j_2$ has 
$[0,\, \rho_{j_1 j_2} / \bar{\rho},\, 0]$;  
and the edge between product node~$j$ and constraint node~$i$ is 
$[0,\, 0,\, A_{ij}]$.

\vspace{1mm}
\textbf{Numerical Results.} 
In Appendix~\ref{append:other_choice}, we report the numerical experiments under these choice models. Since exactly solving large-scale problems is prohibitive, we utilize the ROLS policy as the performance baseline. Table~\ref{tab:other_choice} presents representative results. Note that the benchmark in Table~\ref{tab:other_choice} is the ROLS policy, as no efficient exact algorithm is available for these problem instances. According to Table~\ref{tab:other_choice}, we find that the GCN-based policies can still derive high-quality solutions for these choice models within seconds.

\begin{table}[ht]
    \centering
    \caption{Performance of different policies under other choice models (see Appendix~\ref{append:other_choice} for details).}
    \label{tab:other_choice}
    \begin{tabular}{@{} *{7}{c} @{}}
    \toprule
       \multicolumn{1}{c}{\multirow{2}{*}{Policies}} & \multicolumn{2}{c@{}}{NL ($N_\test=2,000$)} & \multicolumn{2}{c@{}}{Exponomial ($N_\test=2,000$)} & \multicolumn{2}{c@{}}{MC ($N_\test=100$)} \\
       \cmidrule(l){2-3}  \cmidrule(l){4-5} \cmidrule(l){6-7} 
         & Avg. Ratio (std) & Avg. Time (s) & Avg. Ratio (std) & Avg. Time (s) &  Avg. Ratio (std) & Avg. Time (s) \\
        \midrule 
        GI & 84.5\% (6.0\%) & 0.21 & 94.5\% (2.6\%) & 0.27 & 72.1\% (20.2\%) & 0.10\\
        GILS & 101.5\% (2.3\%) & 93.84 & 100.1\% (0.7\%) & 65.10 & 100.5\% (4.2\%) & 88.55\\
        GIP & 103.9\% (3.3\%) & 7.17 & 79.4\% (5.4\%) & 6.47 & 102.5\% (3.8\%) & 0.84\\
        RO & 23.1\% (9.0\%) & 0.14 & 8.8\% (0.4\%) & 0.18 & 64.3\% (18.0\%) & 0.06\\
        LS & 91.6\% (2.4\%) & 11.23 & 99.4\% (0.9\%) & 18.24 & 95.3\% (0.8\%) & 32.21\\
        RP & 74.9\% (11.7\%) & 4.82 & 44.0\% (8.7\%) & 4.99 & 103.6\% (3.0\%) & 1.37\\
        ROLS & 100\% (0.0\%) & 91.55 & 100.0\% (0.0\%) & 153.76 &100\% (0.0\%) & 83.46\\
        \bottomrule
    \end{tabular}
    
\end{table}

\section{Unknown-Choice-Model Case}\label{sec:model-free}

In the preceding sections, we assumed that the structure of the 
choice model was known and used a GCN to solve the corresponding assortment 
optimization problem. In many practical applications, however, the seller has 
access only to transaction data, while the underlying choice mechanism remains 
unobserved. In this section, we examine the setting where the underlying choice model is unknown, and each product is described by contextual features such as price 
and quality. We assume that the parameters of this latent choice model 
are stable over time. 

Let $\bs{e}_j$ denote a standard basis vector with~1 in the $j$-th position 
and~0 elsewhere, and let $S^+ = S \cup \{0\}$ include the no-purchase option.  
Given an assortment $S$, contextual feature matrix 
$\bs{I} \in \mathbb{R}^{|S| \times F}$, and price vector 
$\bs{r} \in \mathbb{R}^{|S| \times 1}$, we denote by 
$\Phi(S, \bs{I}, \bs{r}, j)$ the (unknown) probability that a customer 
chooses product $j \in S^+$. In practice, we observe a large number of 
historical transactions, each consisting of the tuple 
$(S, \bs{I}, \bs{r}, \bs{X})$, where $\bs{X} = \bs{e}_\ell \in \{0, 1\}^{|S|+1}$ represents the 
observed customer choice, with $\ell$ drawn from~$S^+$ according to the 
probabilities $\{\Phi(S, \bs{I}, \bs{r}, j)\}_{j \in S^+}$. To address this setting, we proceed in two stages.  
First, we train a \emph{choice-GCN} to approximate the underlying choice 
probabilities. The choice-GCN naturally accommodates assortments of varying 
sizes and mitigates errors arising from model misspecification.  
Second, using the learned choice-GCN, we generate a large number of 
small-scale constrained assortment instances and compute their approximate 
optimal assortments by evaluating the predicted revenues of all feasible 
assortments. These derived instances form the training set for a second GCN, 
referred to as the \emph{solution-GCN}, which learns to map problem parameters 
to optimal decisions.  
Once trained, the solution-GCN can be used to output assortment decisions for 
large-scale instances, optionally followed by a local search to refine the 
results.  
The detailed procedure is described in the following subsections.

\subsection{Choice-GCN}

Using the transaction data, we first train the \emph{choice-GCN} to approximate 
the underlying choice probabilities. Each transaction record 
$\ell$ is represented as 
$(S_{\ell}, \bs{I}_{\ell}, \bs{r}_{\ell}, \bs{X}_{\ell})$, 
where $S_{\ell}$ denotes the recommended assortment, 
$\bs{I}_{\ell}$ and $\bs{r}_{\ell}$ represent the corresponding contextual 
features and prices, and $\bs{X}_{\ell}$ is the observed customer choice.  

\begin{figure}[ht]
    \centering
    \subfloat[Choice-GCN]{
        \includegraphics[height=6cm]{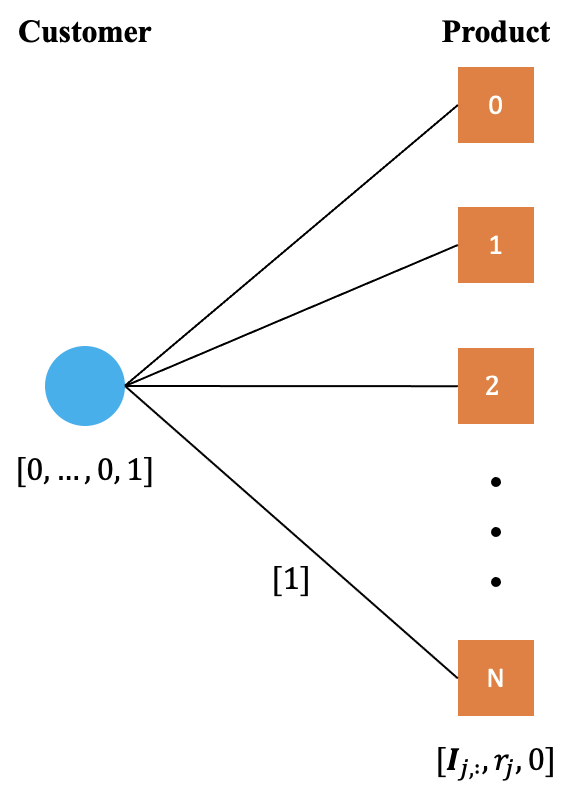}
        \label{fig:choice_graph}
    }
    \qquad \qquad \qquad \qquad
    \subfloat[Solution-GCN]{
        \includegraphics[height=6cm]{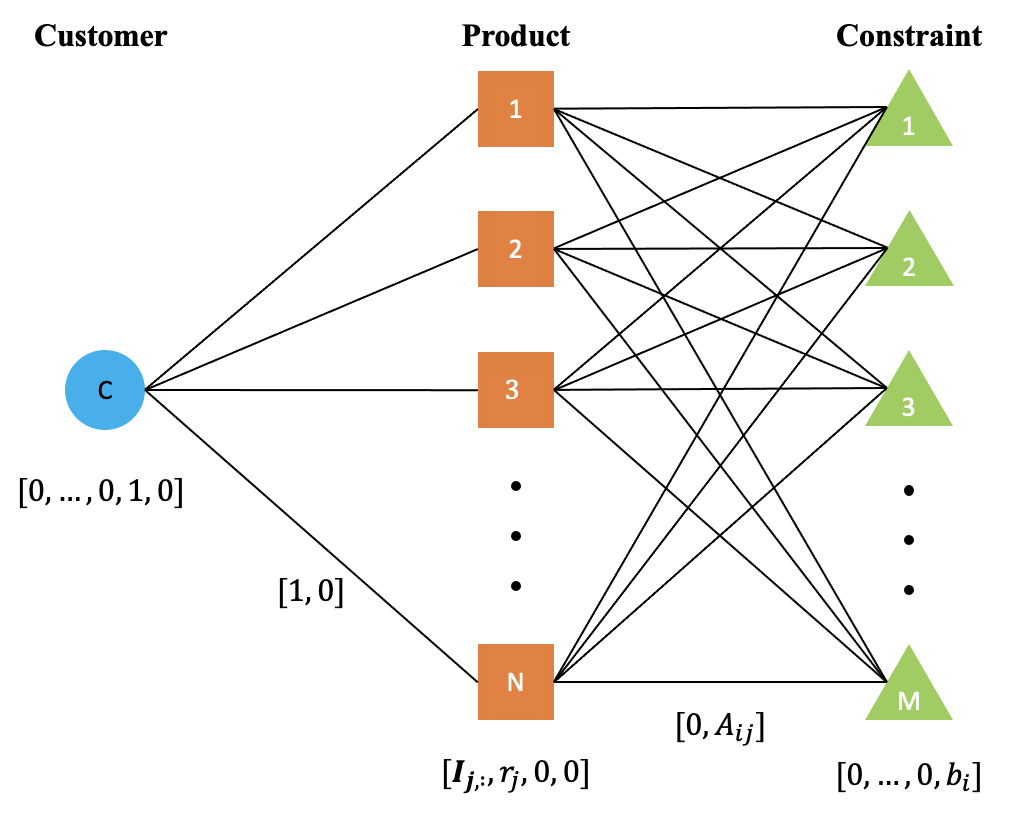}
        \label{fig:assortment_graph}
    }
    \caption{Graph representations of choice-GCN and solution-GCN.}
    \label{fig:model_free_graphs}
\end{figure}

Following the construction in Section~\ref{subsec:graph_represent}, 
we depict the graph architecture for the choice-GCN in 
Figure~\ref{fig:choice_graph}.  
Unlike the previous setting, here we introduce only a single customer node, 
as the underlying structure of the choice model is unknown but fixed.

According to Figure~\ref{fig:choice_graph}, given $N$ products, the graph 
contains one customer node and $N{+}1$ product nodes.  
The feature vectors are specified as follows.  
Each product node $j \in [N]$ has feature vector 
$[\bs{I}_{j,:},\, r_j,\, 0]$, where $\bs{I}_{j,:}$ denotes the contextual 
attributes of product~$j$.  
The product node labeled~0 corresponds to the no-purchase option and has an 
all-zero feature vector.  
The customer node has feature vector $[0, \ldots, 0, 1]$, and each edge has 
a scalar feature $[1]$.  
Although we introduce only a single customer node due to the lack of explicit 
knowledge of the underlying choice structure, the choice-GCN can still 
approximate the choice probabilities effectively because the latent choice 
model is assumed to be fixed.  

Following the same architecture described in Section~\ref{sec:gcn_policy}, 
the choice-GCN adopts the general GCN framework introduced in 
Section~\ref{sec:GCN}.  
Given an input $(S, \bs{I}, \bs{r})$, the network outputs a scalar score 
$w_i(S, \bs{I}, \bs{r}, \bs{\theta})$ for each product node~$i$, where 
$\bs{\theta}$ denotes the set of learnable parameters.  
The predicted choice probability for product $j \in S^+$ is computed as 
\[
\hat{\Phi}(S, \bs{I}, \bs{r}, \bs{\theta}, j)
= \frac{\exp\!\left(w_j(S, \bs{I}, \bs{r}, \bs{\theta})\right)}
       {\sum_{j' \in S^+}
        \exp\!\left(w_{j'}(S, \bs{I}, \bs{r}, \bs{\theta})\right)}.
\]

With $H$ observed data, the choice-GCN is trained by minimizing the loss function $\mathcal{L}(\bs{\theta})
    = -\frac{1}{H} \sum_{\ell=1}^H
      \sum_{j \in S_\ell^+}
      X_{\ell, j} \,
      \log \hat{\Phi}(S_\ell, \bs{I}_\ell, \bs{r}_\ell, \bs{\theta}, j)$,
% \begin{align}\label{eq:model_free_prob}
%     \mathcal{L}(\bs{\theta})
%     = -\frac{1}{H} \sum_{\ell=1}^H
%       \sum_{j \in S_\ell^+}
%       X_{\ell, j} \,
%       \log \hat{\Phi}(S_\ell, \bs{I}_\ell, \bs{r}_\ell, \bs{\theta}, j),
% \end{align}
which is the unweighted version of~\eqref{eq:CE_loss} and corresponds to the 
standard maximum-likelihood estimation loss.  
Before training, the transaction data are randomly split into an 80\% training 
set and a 20\% validation set.  
We employ the Adam optimizer to update the model weights and determine the 
optimal stopping epoch and final model based on the validation loss.  

Although the resulting probability formula resembles that of the MNL model, 
the choice-GCN does not rely on the independence-of-irrelevant-alternatives 
(IIA) assumption.  
Because $\hat{\Phi}(S, \bs{I}, \bs{r}, \bs{\theta}, j)$ explicitly depends on 
the characteristics of all products in~$S$, the model can capture complex and 
non-regular choice behaviors beyond the classical MNL structure.

\subsection{Solution Sample Generation}\label{subsec:sol_generation}

To construct training samples for the solution-GCN, we first generate a set of 
small-scale constrained assortment optimization instances characterized by 
contextual information $(\bs{I}, \bs{r})$ and constraint information 
$(\bs{A}, \bs{b})$.  
For each instance, the choice probabilities of all products are approximated 
using the trained choice-GCN, allowing us to compute the corresponding 
expected revenues.  
We then enumerate all feasible assortments to identify the one with the 
highest approximated expected revenue, denoted by $\bs{x}^*$.  
Each resulting training sample $\ell$ is thus represented as 
$(\bs{I}_\ell, \bs{r}_\ell, \bs{A}_\ell, \bs{b}_\ell, \bs{x}^*_\ell)$.  

\begin{remark}
    In order to speed up computation, we can aggregate inputs into several batches and then compute the revenues of feasible assortments batch by batch. With the help of GPU, the computation of each instance can be finished in seconds. 
\end{remark}

\subsection{Solution-GCN}

Using the training samples generated in Section~\ref{subsec:sol_generation}, 
we train the \emph{solution-GCN} to learn the mapping from problem parameters 
to optimal assortments.  
The corresponding graph representation of the constrained assortment 
optimization problem is illustrated in 
Figure~\ref{fig:assortment_graph}.  
As shown in the figure, each instance with $N$ products and $M$ constraints 
consists of one customer node, $N$ product nodes, and $M$ constraint nodes.  

The feature vectors are defined as follows.  
Each product node~$j$ has feature vector 
$[\bs{I}_{j,:},\, r_j,\, 0,\, 0]$.  
The single customer node has feature vector $[0, \ldots, 0, 1, 0]$,  
and each constraint node~$i$ has $[0, 0, \ldots, 0, 0, b_i]$.  
For edges, the feature vector connecting the customer node to product node~$j$ 
is $[1, 0]$, while that between product node~$j$ and constraint node~$i$ is 
$[0, A_{ij}]$.  
Similar to the base model, we train the solution-GCN using the 
weighted cross-entropy loss in~\eqref{eq:CE_loss}, based on the 
samples constructed in Section~\ref{subsec:sol_generation}.  

The inference step follows the same policies as in the base model—GI, GILS, 
and GIP—with a minor modification.  
Since the underlying choice model is unknown, we use the choice-GCN to 
approximate the expected revenue of each candidate assortment and select the 
one with the highest predicted value.  

It is worth noting that the choice-GCN can be replaced by other existing 
methods for approximating choice probabilities, such as those proposed by 
\citet{aouad2022representing}, \citet{wang2023transformer}, and 
\citet{liu2025beyond}, provided they yield accurate predictions across 
assortments of varying sizes.  
By following the same three-step framework—choice-model estimation, 
sample generation, and solution-GCN training—the resulting policies are 
expected to achieve similarly strong performance. Moreover, if we replace the choice-GCN with some existing choice models (e.g., MMNL), then the framework reduces to that in Section~\ref{sec:gcn_policy}. 

\subsection{Numerical Experiments}\label{subsec:model_free_num}

In the numerical experiments, although our framework can accommodate fairly general choice models, we use the MMNL model as the underlying model because ground-truth benchmarks are difficult to obtain for other choice models. Consider the MMNL model with $K=10$ customer types as the underlying choice model and fix the contextual feature dimension at $F=5$. In this setting, the true choice probability of product~$j$ in 
assortment~$S$ is given by
\begin{align*}
    \Phi(S, \bs{I}, \bs{r}, j)
    = \sum_{k=1}^{K} \alpha_k
      \frac{
        \exp\!\left(
            \sum_{f=1}^{F} c_{k,f} I_{j,f} - \eta r_j
        \right)
      }{
        1 + \sum_{j' \in S}
        \exp\!\left(
            \sum_{f=1}^{F} c_{k,f} I_{j',f} - \eta r_{j'}
        \right)
      },
\end{align*}
where $\eta = 3$ is the price-sensitivity parameter, 
$\bs{\alpha}$ is the market-share vector across customer types, and 
$\bs{C}$ is the type-specific context-weight matrix.  

We randomly generate $\bs{\alpha}$ and $\bs{C}$ as follows:
\begin{itemize}
    \item We sample $K$ values $z_k \sim \mathcal{U}[0,1]$ and set
    $\alpha_k = z_k / \sum_{k'=1}^{K} z_{k'}$.

    \item For each customer type~$k$, we draw $F$ values 
    $c_{k,f} \sim \mathcal{U}[0,1]$ and compute
    $C_{k,f} = c_{k,f} / \sum_{f'=1}^{F} c_{k,f'}$.
\end{itemize}

\vspace{1mm}
Both $\bs{\alpha}$ and $\bs{C}$ remain fixed across all numerical experiments.  

\vspace{1mm}
\tb{Transaction data generation.} 
To simulate a realistic environment, we first generate 20{,}000 transaction 
records with $N_{\trans} = 15$ products.  
Following the setup in Section~\ref{sec:numerical}, for each transaction we 
draw prices and contextual features independently as 
$r_j \sim \mathcal{U}[1, 2]$ and $I_{j,f} \sim \mathcal{U}[0, 1]$.  
Given the assortment $S$, we then sample the customer's one-shot choice 
$\bs{X}$ according to the true choice probabilities 
$\{\Phi(S, \bs{I}, \bs{r}, j)\}_{j \in S^+}$.  

\vspace{1mm}
\tb{Choice-GCN training.} 
The network architecture follows the framework in 
Figure~\ref{fig:GCN_network}, with the hidden-layer dimension set to 
$d_{\text{mid}} = 64$.  
The transaction dataset is randomly split into 80\% training and 20\% validation sets.  
Training terminates when the validation loss fails to improve for 20~epochs 
or reaches 100~epochs in total.  
We employ the Adam optimizer with a learning rate of $10^{-4}$ and batch size 
2{,}048.  
Ten independent trials are conducted to account for randomness; the average 
training time across trials is 346~seconds.  

We evaluate the trained models on 100~testing instances with 
$N_{\test} = 20$.  
To assess estimation accuracy, we compute the root-mean-squared error (RMSE) 
and mean absolute percentage error (MAPE) (see Appendix~\ref{append:def_rmse} for formulas). Across the ten trials, the average RMSE is 0.010 (standard deviation~0.005) 
and the average MAPE is 20.5\% (standard deviation~3.9\%).  
For the subsequent analysis, we select the model from one representative trial, 
which yields an RMSE of 0.011 and a MAPE of 24.0\%.  

\vspace{1mm}
\tb{Solution sample generation.} 
In this experiment, we generate 2{,}000 solution samples for each product size 
$N_{\train} \in \{16, 17, 18, 19, 20\}$, each with 
$M_{\train} = 10$ constraints, resulting in a total of 10{,}000 samples.  
For each pair $(N, M)$, the samples are generated as follows:
\begin{enumerate}
    \item We first generate revenues and context vectors as $r_{j} \sim \mathcal{U}[1, 2]$ and $I_{j, f} \sim \mathcal{U}[0, 1]$. Then, we generate two types of constraints:
    \begin{itemize}
        \item \emph{Capacity constraints:}  
        For $\lfloor M/2 \rfloor$ constraints, sample 
        $A_{ij} \sim \mathcal{U}[0, 1]$ for all $j \in [N]$ and 
        $b_i \sim \mathcal{U}[5, 10]$.  
        \item \emph{Precedence constraints:}  
        For $\lceil M/2 \rceil$ constraints, set $b_i = 0$, then draw 
        $j_1, j_2 \sim \mathcal{U_D}[N]$.  
        If $j_1 \neq j_2$, assign $A_{i j_1} = 1$, $A_{i j_2} = -1$, and 
        $A_{i j'} = 0$ for all 
        $j' \in [N] \setminus \{j_1, j_2\}$; otherwise, set 
        $A_{ij} = 0$ for all $j \in [N]$.  
    \end{itemize}

    \item Lastly, using the trained choice-GCN, we evaluate the expected revenues of all feasible assortments and select the one with the highest revenue as the 
    approximated optimal assortment.  
\end{enumerate}

\vspace{1mm}
The total generation time for the 10{,}000 samples is 87~minutes.    
The resulting optimality ratios and generation times are summarized in 
Table~\ref{tab:soltion_quality} in Appendix~\ref{append:generation_time}.

\begin{table}[ht]
    \centering
    \caption{Performance of proposed policies and existing policies when $S_\trans=20,000$, $N_\train=20$ and $N_\test\in \{500, 2,000\}$; unknown-choice-model case. }
    \label{tab:comparison_free} 
    \begin{tabular}{@{} *{5}{c} @{}}
    \toprule
       \multicolumn{1}{c}{\multirow{2}{*}{Policies}} & \multicolumn{2}{c@{}}{$N_\test=500$} & \multicolumn{2}{c@{}}{$N_\test=2,000$} \\
       \cmidrule(l){2-3}  \cmidrule(l){4-5} 
         & Avg. Ratio (std) & Avg. Time (s) & Avg. Ratio (std) & Avg. Time (s)\\
        \midrule 
        GI & 91.6\% (3.1\%) & 0.06 & 88.6\% (4.6\%) & 0.27\\
        GILS & 95.1\% (1.8\%) & 8.22 & 93.2\% (2.2\%) & 69.63\\
        GIP & 96.1\% (1.3\%) & 2.56 & 95.9\% (1.6\%) & 6.89\\
        RO & 13.0\% (2.0\%) & 0.04 & 11.5\% (2.0\%) & 0.22\\
        LS & 77.3\% (3.7\%) & 14.85 & 76.4\% (5.2\%) & 184.60\\
        ROLS & 89.9\% (2.3\%) & 30.05 & 87.0\% (1.9\%) & 206.30\\
        RP & 64.1\% (5.3\%) & 2.52 & 62.7\% (3.9\%) & 5.96\\
        \bottomrule
    \end{tabular}
\end{table}

\vspace{1mm}
\tb{Solution-GCN training.} 
Following the same procedure as in the base model, we train the 
solution-GCN using the generated solution samples and apply the proposed 
policies to infer the optimal assortments.  
When calling the choice-GCN, we employ batch 
operations and GPU acceleration to expedite computation.  For benchmarking, we use the CONIC policy—implemented with full knowledge of 
the underlying choice model—as the performance baseline.  
Competitive ratios are defined relative to this benchmark.  
When computing these ratios, the ground-truth revenues of the assortments 
produced by each policy are evaluated using the true underlying model.  
The corresponding numerical results are summarized in 
Table~\ref{tab:comparison_free}.

According to Table~\ref{tab:comparison_free}, the proposed policies achieve 
competitive ratios exceeding 88\% within only a few seconds.  
Although the underlying choice model is unknown, it remains fixed across 
instances and can be effectively approximated by the choice-GCN. Thus, the performance of the proposed policies is comparable to that in Table~\ref{tab:comparison_general}, where the choice model is known but varies across instances.

\section{Concluding Remarks}\label{sec:conclusion}
In this work, we consider the scenario where the decision maker needs to frequently solve constrained assortment problems. In this case, the optimal solutions possess some latent features, which can be learned by a neural network. Then, we can use the trained neural network to rapidly generate prediction of each product's probability being in the optimal assortment. Based on these predicted probabilities, we propose several efficient inference policies to derive high-quality solutions. In order to address the scalability challenge, we adopt the graph convolutional network that possesses a weight-sharing mechanism and hence can handle inputs with different sizes. 

We start with the known-choice-model case, where both the choice model and its parameters are known. We propose several policies based on the GCN's output. In the numerical experiments, even though the GCN is trained on small-scale instances (e.g., with 20 products), the proposed policies can achieve superior performance on large-scale instances with up to 2,000 products in a few seconds. Then, we consider the unknown-choice-model case, where the underlying choice model is unknown. In this case, we propose a three-step framework to train a GCN for the assortment optimization problem, and the adapted policies also possess the ``from small to large'' generalization ability.

To the best of our knowledge, this work presents the first learning-to-optimize framework for solving constrained assortment optimization problems. In this work, we theoretically and numerically show that the mapping from problem parameters to optimal solution has some latent patterns that can be learned by a GCN. Moreover, the mapping learned from small-scale samples can be generalized to large-scale samples through the GCN. The resulting policies obtain high-quality solutions within seconds, demonstrating both effectiveness and computational efficiency. Our work illustrates the potential of GCN in solving constrained assortment optimization problems, and we believe it can be adopted to solve other revenue management problems.

Lastly, we discuss limitations and future directions. Since the proposed policy learns patterns from training samples, if the testing sample is significantly different from all training samples, the proposed policy may not work well. Thus, practitioners may train separate GCNs for different
scenarios (e.g., aggregated by categories). There are several directions to be explored, such as design of alternative inference policies and unsupervised learning approaches.

\newpage 
\bibliographystyle{informs2014}
\bibliography{reference}

@article{hornik1989multilayer,
  title={Multilayer feedforward networks are universal approximators},
  author={Hornik, Kurt and Stinchcombe, Maxwell and White, Halbert},
  journal={Neural Networks},
  volume={2},
  number={5},
  pages={359--366},
  year={1989},
  publisher={Elsevier}
}

@article{gallego2024efficient,
  title={Efficient Local-Search Heuristics for Online and Offline Assortment Optimization},
  author={Gallego, Guillermo and Jagabathula, Srikanth and Lu, Wentao},
  journal={Available at SSRN 4828069},
  year={2024}
}

@inproceedings{gasse2019exact,
  title={Exact combinatorial optimization with graph convolutional neural networks},
  author={Gasse, Maxime and Ch{\'e}telat, Didier and Ferroni, Nicola and Charlin, Laurent and Lodi, Andrea},
  booktitle={Proceedings of the 33rd Conference on Neural Information Processing Systems},
  pages={15580--15592},
  year={2019}
}

@article{aouad2023exponomial,
  title={The exponomial choice model for assortment optimization: An alternative to the {MNL} model?},
  author={Aouad, Ali and Feldman, Jacob and Segev, Danny},
  journal={Management Science},
  volume={69},
  number={5},
  pages={2814--2832},
  year={2023},
  publisher={INFORMS}
}

@article{liu2025beyond,
  title={Beyond Complements and Substitutes: A Graph Neural Network Approach for Collaborative Retail Sales Forecasting},
  author={Liu, Jing and Wang, Gang and Zhao, Huimin and Lu, Mingfeng and Huang, Lihua and Chen, Gang},
  journal={Information Systems Research},
  year={2025},
  volume={36},
  number={4},
  pages={1993--2016},
  publisher={INFORMS}
}

@article{alptekinouglu2016exponomial,
  title={The exponomial choice model: A new alternative for assortment and price optimization},
  author={Alptekino{\u{g}}lu, Ayd{\i}n and Semple, John H},
  journal={Operations Research},
  volume={64},
  number={1},
  pages={79--93},
  year={2016},
  publisher={INFORMS}
}

@article{csen2018conic,
  title={A conic integer optimization approach to the constrained assortment problem under the mixed multinomial logit model},
  author={{\c{S}}en, Alper and Atamt{\"u}rk, Alper and Kaminsky, Philip},
  journal={Operations Research},
  volume={66},
  number={4},
  pages={994--1003},
  year={2018},
  publisher={INFORMS}
}

@article{feldman2022customer,
  title={Customer choice models vs. machine learning: Finding optimal product displays on {Alibaba}},
  author={Feldman, Jacob and Zhang, Dennis J and Liu, Xiaofei and Zhang, Nannan},
  journal={Operations Research},
  volume={70},
  number={1},
  pages={309--328},
  year={2022},
  publisher={INFORMS}
}

@inproceedings{fan2023smart,
  title={Smart initial basis selection for linear programs},
  author={Fan, Zhenan and Wang, Xinglu and Yakovenko, Oleksandr and Sivas, Abdullah Ali and Ren, Owen and Zhang, Yong and Zhou, Zirui},
  booktitle={Proceedings of the 40th International Conference on Machine Learning},
  pages={9650--9664},
  year={2023}
}

@inproceedings{liu2024learning,
  title={Learning to Pivot as a Smart Expert},
  author={Liu, Tianhao and Pu, Shanwen and Ge, Dongdong and Ye, Yinyu},
  booktitle={Proceedings of the 38th AAAI Conference on Artificial Intelligence},
  pages={8073--8081},
  year={2024}
}

@article{talluri2004revenue,
  title={Revenue management under a general discrete choice model of consumer behavior},
  author={Talluri, Kalyan and Van Ryzin, Garrett},
  journal={Management Science},
  volume={50},
  number={1},
  pages={15--33},
  year={2004},
  publisher={INFORMS}
}

@article{rusmevichientong2014assortment,
  title={Assortment optimization under the multinomial logit model with random choice parameters},
  author={Rusmevichientong, Paat and Shmoys, David and Tong, Chaoxu and Topaloglu, Huseyin},
  journal={Production and Operations Management},
  volume={23},
  number={11},
  pages={2023--2039},
  year={2014},
  publisher={SAGE Publications Sage CA: Los Angeles, CA}
}

@article{wang2024neural,
  title={Neural-Network Mixed Logit Choice Model: Statistical and Optimality Guarantees},
  author={Wang, Zhi and Gao, Rui and Li, Shuang},
  journal={Available at SSRN 5118033},
  year={2024}
}

@article{mcfadden2000mixed,
  title={Mixed {MNL} models for discrete response},
  author={McFadden, Daniel and Train, Kenneth},
  journal={Journal of Applied Econometrics},
  volume={15},
  number={5},
  pages={447--470},
  year={2000},
  publisher={Wiley Online Library}
}

@article{li2020deepergcn,
  title={{DeeperGCN}: {All} you need to train deeper {GCN}s},
  author={Li, Guohao and Xiong, Chenxin and Thabet, Ali and Ghanem, Bernard},
  journal={arXiv preprint arXiv:2006.07739},
  year={2020}
}

@article{blanchet2016markov,
  title={A {Markov} chain approximation to choice modeling},
  author={Blanchet, Jose and Gallego, Guillermo and Goyal, Vineet},
  journal={Operations Research},
  volume={64},
  number={4},
  pages={886--905},
  year={2016},
  publisher={INFORMS}
}

@article{wang2023transformer,
  title={Transformer choice net: A transformer neural network for choice prediction},
  author={Wang, Hanzhao and Li, Xiaocheng and Talluri, Kalyan},
  journal={arXiv preprint arXiv:2310.08716},
  year={2023}
}

@inproceedings{rosenfeld2020predicting,
  title={Predicting choice with set-dependent aggregation},
  author={Rosenfeld, Nir and Oshiba, Kojin and Singer, Yaron},
  booktitle={Proceedings of the 37th International Conference on Machine Learning},
  pages={8220--8229},
  year={2020}
}

@article{mcfadden1972conditional,
  title={Conditional logit analysis of qualitative choice behavior},
  author={McFadden, Daniel},
  year={1972},
  journal={Frontiers in Econometrics},
  pages={105--142}
}

@article{wang2023neural,
  title={A neural network based choice model for assortment optimization},
  author={Wang, Hanzhao and Cai, Zhongze and Li, Xiaocheng and Talluri, Kalyan},
  journal={arXiv preprint arXiv:2308.05617},
  year={2023}
}

@article{chen2023machine,
  title={Machine learning methods for data-driven demand estimation and assortment planning considering cross-selling and substitutions},
  author={Chen, Zhen-Yu and Fan, Zhi-Ping and Sun, Minghe},
  journal={INFORMS Journal on Computing},
  volume={35},
  number={1},
  pages={158--177},
  year={2023},
  publisher={INFORMS}
}

@article{desir2022capacitated,
  title={Capacitated assortment optimization: Hardness and approximation},
  author={D{\'e}sir, Antoine and Goyal, Vineet and Zhang, Jiawei},
  journal={Operations Research},
  volume={70},
  number={2},
  pages={893--904},
  year={2022},
  publisher={INFORMS}
}

@article{cappart2023combinatorial,
  title={Combinatorial optimization and reasoning with graph neural networks},
  author={Cappart, Quentin and Ch{\'e}telat, Didier and Khalil, Elias B and Lodi, Andrea and Morris, Christopher and Veli{\v{c}}kovi{\'c}, Petar},
  journal={Journal of Machine Learning Research},
  volume={24},
  number={130},
  pages={1--61},
  year={2023}
}

@article{mendez2014branch,
  title={A branch-and-cut algorithm for the latent-class logit assortment problem},
  author={M{\'e}ndez-D{\'\i}az, Isabel and Miranda-Bront, Juan Jos{\'e} and Vulcano, Gustavo and Zabala, Paula},
  journal={Discrete Applied Mathematics},
  volume={164},
  pages={246--263},
  year={2014},
  publisher={Elsevier}
}

@article{berbeglia2020assortment,
  title={Assortment optimisation under a general discrete choice model: A tight analysis of revenue-ordered assortments},
  author={Berbeglia, Gerardo and Joret, Gwena{\"e}l},
  journal={Algorithmica},
  volume={82},
  number={4},
  pages={681--720},
  year={2020},
  publisher={Springer}
}

@article{wang2017consumer,
  title={Consumer choice models with endogenous network effects},
  author={Wang, Ruxian and Wang, Zizhuo},
  journal={Management Science},
  volume={63},
  number={11},
  pages={3944--3960},
  year={2017},
  publisher={INFORMS}
}

@article{rusmevichientong2010dynamic,
  title={Dynamic assortment optimization with a multinomial logit choice model and capacity constraint},
  author={Rusmevichientong, Paat and Shen, Zuo-Jun Max and Shmoys, David B},
  journal={Operations Research},
  volume={58},
  number={6},
  pages={1666--1680},
  year={2010},
  publisher={INFORMS}
}

@article{gallego2015general,
  title={A general attraction model and sales-based linear program for network revenue management under customer choice},
  author={Gallego, Guillermo and Ratliff, Richard and Shebalov, Sergey},
  journal={Operations Research},
  volume={63},
  number={1},
  pages={212--232},
  year={2015},
  publisher={INFORMS}
}

@article{gallego2021bounds,
  title={Bounds, heuristics, and prophet inequalities for assortment optimization},
  author={Gallego, Guillermo and Berbeglia, Gerardo},
  journal={Available at SSRN 3934096},
  year={2021}
}

@article{gallego2024assortment,
  title={Assortment Optimization with Downward Feasibility: Efficient Heuristics Based on Independent Demands},
  author={Gallego, Guillermo and Gao, Pin and Wang, Shaoyu and Berbeglia, Gerardo},
  journal={Available at SSRN},
  year={2024}
}

@article{mcfadden1980econometric,
  title={Econometric models for probabilistic choice among products},
  author={McFadden, Daniel},
  journal={The Journal of Business},
  volume={53},
  number={3},
  pages={S13--S29},
  year={1980},
  publisher={JSTOR}
}

@book{gallego2019revenue,
  title={Revenue Management and Pricing Analytics},
  author={Gallego, Guillermo and Topaloglu, Huseyin},
  year={2019},
  publisher={Springer}
}

@incollection{wang2021discrete,
  title={Discrete choice models and applications in operations management},
  author={Wang, Ruxian},
  booktitle={Tutorials in Operations Research: Emerging Optimization Methods and Modeling Techniques with Applications},
  publisher={INFORMS},
  pages={199--226},
  year={2021},
}

@article{aouad2022representing,
  title={Representing random utility choice models with neural networks},
  author={Aouad, Ali and D{\'e}sir, Antoine},
  journal={arXiv preprint arXiv:2207.12877},
  year={2022}
}

@inproceedings{nazari2018reinforcement,
  title={Reinforcement learning for solving the vehicle routing problem},
  author={Nazari, Mohammadreza and Oroojlooy, Afshin and Snyder, Lawrence and Tak{\'a}c, Martin},
  booktitle={Proceedings of the 32nd Conference on Neural Information Processing Systems},
  year={2018},
  pages={9861--9871}
}

@inproceedings{khalil2016learning,
  title={Learning to branch in mixed integer programming},
  author={Khalil, Elias and Le Bodic, Pierre and Song, Le and Nemhauser, George and Dilkina, Bistra},
  booktitle={Proceedings of the 30th AAAI Conference on Artificial Intelligence},
  pages={724--731},
  year={2016}
}

@inproceedings{li2019graph,
  title={Graph matching networks for learning the similarity of graph structured objects},
  author={Li, Yujia and Gu, Chenjie and Dullien, Thomas and Vinyals, Oriol and Kohli, Pushmeet},
  booktitle={Proceedings of the 36th International Conference on Machine Learning},
  pages={3835--3845},
  year={2019},
}

@article{hausman1984specification,
  title={Specification tests for the multinomial logit model},
  author={Hausman, Jerry and McFadden, Daniel},
  journal={Econometrica},
  volume={52},
  number={5},
  pages={1219--1240},
  year={1984},
  publisher={JSTOR}
}

@article{kingma2014adam,
  title={Adam: A method for stochastic optimization},
  author={Kingma, Diederik P and Ba, Jimmy},
  journal={arXiv preprint arXiv:1412.6980},
  year={2014}
}

@article{srivastava2014dropout,
  title={Dropout: A simple way to prevent neural networks from overfitting},
  author={Srivastava, Nitish and Hinton, Geoffrey and Krizhevsky, Alex and Sutskever, Ilya and Salakhutdinov, Ruslan},
  journal={Journal of Machine Learning Research},
  volume={15},
  number={1},
  pages={1929--1958},
  year={2014}
}

@book{wolsey1999integer,
  title={Integer and Combinatorial Optimization},
  author={Wolsey, Laurence A and Nemhauser, George L},
  year={1999},
  publisher={John Wiley \& Sons}
}

@article{desir2020constrained,
  title={Constrained assortment optimization under the Markov chain--based choice model},
  author={D{\'e}sir, Antoine and Goyal, Vineet and Segev, Danny and Ye, Chun},
  journal={Management Science},
  volume={66},
  number={2},
  pages={698--721},
  year={2020},
  publisher={INFORMS}
}

@article{gallego2014constrained,
  title={Constrained assortment optimization for the nested logit model},
  author={Gallego, Guillermo and Topaloglu, Huseyin},
  journal={Management Science},
  volume={60},
  number={10},
  pages={2583--2601},
  year={2014},
  publisher={INFORMS}
}

@inproceedings{glorot2011deep,
  title={Deep sparse rectifier neural networks},
  author={Glorot, Xavier and Bordes, Antoine and Bengio, Yoshua},
  booktitle={Proceedings of the 14th International Conference on Artificial Intelligence and Statistics},
  pages={315--323},
  year={2011},
}

@inproceedings{lin2017focal,
  title={Focal loss for dense object detection},
  author={Lin, Tsung-Yi and Goyal, Priya and Girshick, Ross and He, Kaiming and Doll{\'a}r, Piotr},
  booktitle={Proceedings of 2017 IEEE International Conference on Computer Vision},
  pages={2980--2988},
  year={2017}
}

@article{zhang2024modeling,
  title={Modeling Reference-dependent Choices with Graph Neural Networks},
  author={Zhang, Liang and Liu, Guannan and Wu, Junjie and Tan, Yong},
  journal={arXiv preprint arXiv:2408.11302},
  year={2024}
}

@inproceedings{
chen2023on,
title={On Representing Mixed-Integer Linear Programs by Graph Neural Networks},
author={Ziang Chen and Jialin Liu and Xinshang Wang and Wotao Yin},
booktitle={Proceedings of the 11th International Conference on Learning Representations },
year={2023},
}

@inproceedings{
azizian2021expressive,
title={Expressive Power of Invariant and Equivariant Graph Neural Networks},
author={Waiss Azizian and Marc Lelarge},
booktitle={Proceedings of the 9th International Conference on Learning Representations},
year={2021},
}

@article{cybenko1989approximation,
  title={Approximation by superpositions of a sigmoidal function},
  author={Cybenko, George},
  journal={Mathematics of Control, Signals and Systems},
  volume={2},
  number={4},
  pages={303--314},
  year={1989},
  publisher={Springer}
}

@article{weisfeiler1968reduction,
  title={The reduction of a graph to canonical form and the algebra which appears therein},
  author={Weisfeiler, Boris and Leman, Andrei},
  journal={NTI, Series},
  volume={2},
  number={9},
  pages={12--16},
  year={1968}
}

@article{berkholz2017tight,
  title={Tight lower and upper bounds for the complexity of canonical colour refinement},
  author={Berkholz, Christoph and Bonsma, Paul and Grohe, Martin},
  journal={Theory of Computing Systems},
  volume={60},
  number={4},
  pages={581--614},
  year={2017},
  publisher={Springer}
}

@inproceedings{
chen2023gnn-lp,
title={On Representing Linear Programs by Graph Neural Networks},
author={Ziang Chen and Jialin Liu and Xinshang Wang and Jianfeng Lu and Wotao Yin},
booktitle={Proceedings of the 11th International Conference on Learning Representations},
year={2023}
}

@article{zaheer2017deep,
  title={Deep sets},
  author={Zaheer, Manzil and Kottur, Satwik and Ravanbakhsh, Siamak and Poczos, Barnabas and Salakhutdinov, Russ R and Smola, Alexander J},
  journal={Proceedings of the 31st Advances in neural information processing systems},
  year={2017}
}

@article{jegelka2022theory,
      title={Theory of Graph Neural Networks: Representation and Learning}, 
      author={Jegelka, Stefanie},
      year={2022},
      journal={arXiv preprint arXiv:2204.07697},
}

@article{zhang2024expressive,
  title={The expressive power of graph neural networks: A survey},
  author={Zhang, Bingxu and Fan, Changjun and Liu, Shixuan and Huang, Kuihua and Zhao, Xiang and Huang, Jincai and Liu, Zhong},
  journal={IEEE Transactions on Knowledge and Data Engineering},
  volume={37},
  number={3},
  pages={1455--1474},
  year={2024},
  publisher={IEEE}
}

@inproceedings{
xu2018how,
title={How Powerful are Graph Neural Networks?},
author={Keyulu Xu and Weihua Hu and Jure Leskovec and Stefanie Jegelka},
booktitle={Proceedings of the 7th International Conference on Learning Representations},
year={2019},
}

@inproceedings{
chen2025expressive,
title={Expressive Power of Graph Neural Networks for (Mixed-Integer) Quadratic Programs},
author={Ziang Chen and Xiaohan Chen and Jialin Liu and Xinshang Wang and Wotao Yin},
booktitle={Proceedings of the 42nd International Conference on Machine Learning},
year={2025},
pages={7880 -- 7911},
}

@inproceedings{
chen2024rethinking,
title={Rethinking the Capacity of Graph Neural Networks for Branching Strategy},
author={Ziang Chen and Jialin Liu and Xiaohan Chen and Xinshang Wang and Wotao Yin},
booktitle={Proceedings of the 38th Annual Conference on Neural Information Processing Systems},
year={2024},
pages={123991 -- 124024}
}

@inproceedings{yehudai2021local,
  title={From local structures to size generalization in graph neural networks},
  author={Yehudai, Gilad and Fetaya, Ethan and Meirom, Eli and Chechik, Gal and Maron, Haggai},
  booktitle={Proceedings of the 38th International Conference on Machine Learning},
  pages={11975--11986},
  year={2021},
}

@article{ruiz2020graphon,
  title={Graphon neural networks and the transferability of graph neural networks},
  author={Ruiz, Luana and Chamon, Luiz and Ribeiro, Alejandro},
  journal={Proceedings of the 34th Annual Conference on Neural Information Processing Systems},
  volume={33},
  pages={1702--1712},
  year={2020}
}

@inproceedings{
maskey2022generalization,
title={Generalization Analysis of Message Passing Neural Networks on Large Random Graphs},
author={Sohir Maskey and Ron Levie and Yunseok Lee and Gitta Kutyniok},
booktitle={Proceedings of the 36th Annual Conference on Neural Information Processing Systems},
year={2022},
}

@article{guo2025solving,
  title={Solving Assortment Optimization with First-Order Methods and Neural Networks: A Computational Framework and Public Benchmark},
  author={Guo, Qing and Lagzi, Saman and Wang, Chenhao and Chen, Ningyuan and Gallego, Guillermo and Kunnumkal, Sumit and Wang, Yao and Yu, Li},
  journal={Available at SSRN 5671592},
  year={2025}
}
  % References here (outcomment the appropriate case)

\newpage
\begin{appendices}
\begin{center}
    \textbf{\Large Appendix}
\end{center}

\renewcommand{\theHsection}{A\arabic{section}}

\section{Distribution Shift for Base Simulation}\label{append:shift}

In this subsection, we evaluate the robustness of the proposed policies under 
a distribution shift in the testing data. Specifically, when generating the 
100~testing instances with $N_{\test}=500$, we modify the distributions of 
the revenues $r_j$ and utilities $u_{kj}$ as follows:  
$
r_j \sim \mathrm{Beta}(0.5,\,0.5;\,1,\,2)$ and 
$
u_{kj} \sim \mathrm{Beta}(0.5,\,0.5;\,0,\,1)
$
where $\mathrm{Beta}(\alpha,\beta;\,a,b)$ denotes the Beta distribution over 
the interval $[a,b]$ with shape parameters $(\alpha,\beta)$.  
The resulting performance of the proposed policies under this shifted 
distribution is summarized in Table~\ref{tab:comparison_shift}.

\begin{table}[ht]
    \centering
    \caption{Performance of proposed policies and existing policies when $K_\train=K_\test=10$, $M_\train=M_\test=10$, $N_\train=20$ and $N_\test=500$; distribution shift.}
    \label{tab:comparison_shift} 
    \begin{tabular}{@{} *{5}{c} @{}}
    \toprule
         Policies & Avg. Ratio (std) & Avg. Time (s) \\
        \midrule 
        GI &  89.8\% (4.0\%) & 0.02 \\
        GILS &  96.3\% (1.4\%) & 8.60\\
        GIP & 96.4\% (1.7\%) & 2.54 \\
        RO & 11.9\% (3.6\%) & 0.04 \\
        LS & 84.5\% (3.9\%) & 5.71 \\
        ROLS & 93.5\% (2.7\%) & 47.00 \\
        RP & 53.2\% (7.0\%) & 2.19 \\
        CONIC & 100.0\% (0.0\%) & 275.00\\
        \bottomrule
    \end{tabular}
\end{table}

According to Table~\ref{tab:comparison_shift}, the proposed policies exhibit 
performance comparable to that reported in 
Table~\ref{tab:comparison_general}. The intuition is straightforward: since 
the training data are generated via uniform sampling over bounded intervals 
and the observed standard deviations are small, the GCN effectively learns the 
underlying structural patterns of the assortment problems within these 
intervals. Consequently, when the testing data follow different but 
similarly supported distributions, such as the Beta distributions considered 
here, the learned representations remain valid and the average competitive 
ratios change only marginally.  

\section{Numerical Experiments under Other Choice Models}\label{append:other_choice}
In this section, we present the numerical experiments for other choice models mentioned in Section~\ref{sec:other_choice}.

\tb{NL model.} In this case, given $(K, N, M, L)$, each (training or testing) instance is generated as follows:
\begin{enumerate}
    \item To generate the proportions of different customer types, we first sample $K$ values $z_k$'s from $\mathcal{U}[0, 1]$, and then choose $\bs{\alpha}=(\alpha_1, \dots, \alpha_k)$ where $\alpha_k = z_k/\left(\sum_{k'=1}^K z_{k'}\right)$. 
    \item We generate revenues and utilities as $r_{j} \sim \mathcal{U}[1, 2]$ and $u_{kj} \sim \mathcal{U}[0, 1]$, and compute $v_{kj} = \exp(u_{kj} - \eta r_j)$ with $\eta=3$ for $k\in [K]$ and $j\in [N]$. 
    \item For each nest $\ell$, we randomly generate its dissimilarity factor as $\gamma_\ell \sim \mathcal{U}[0.5, 1.5]$. For each product $j$, we generate its nest index as $h_j\sim \mathcal{U_D}[L]$. 
    \item Then, we generate two types of constraints:
    \begin{itemize}
        \item $\lfloor M/2\rfloor$ capacity constraints: For constraint $i$, we sample $A_{ij} \sim \mathcal{U}[0, 1]$ for any $j\in [N]$ and $ b_i \sim \mathcal{U}[5, 10]$.
        \item $\lceil M/2\rceil$ precedence constraints: For constraint $i$, we set $b_i=0$ and then sample $j_1\sim \mathcal{U_D}[N]$ and $j_2\sim \mathcal{U_D}[N]$. If $j_1\neq j_2$, we have $A_{ij_1}=1$, $A_{ij_2}=-1$, and $A_{ij'}=0$ for all $j'[N]\setminus\{j_1, j_2\}$; otherwise, $A_{ij}=0$ for all $j\in [N]$. 
    \end{itemize} 
    \item Lastly, for testing samples, we \textit{enumerate} all feasible assortments to derive the optimal assortment $\bs{x}^*$, which serves as the label.
\end{enumerate}

We first randomly generate 2,000 instances with $K_\train=10$ customer types, $N_\train=20$ products, $M_\train=10$ constraints and $L_\train=5$ nests, and then augment the dataset to 10,000 instances by Algorithm~\ref{alg:data_augment}. The total generation time is 246.0 minutes. The training process is the same as that in Section~\ref{sec:numerical}, and we also run 10 independent trials. The average training loss is 0.366, the average validation loss is 0.366, and the average running time is 104 seconds. 

Then, we test the performance of the different policies on 100 instances with $K_\test=10$ customer types, $N_\test=500 \text{ or } 2,000$ products, $M_\test=10$ constraints, and $L_\test=5$ nests. Since solving the optimal solution is prohibitive, we use the objective of the ROLS policy as the baseline. The performance of different policies is summarized in Table~\ref{tab:comparison_nl}.

\begin{table}[ht]
    \centering
    \caption{Performance of proposed policies and existing policies when $L_\train=L_\test=5$,  $K_\train=K_\test=10$, $M_\train=M_\test=10$, $N_\train=20$ and $N_\test=500\mbox{ or } 2,000$; NL model.}
    \label{tab:comparison_nl} 
    \begin{tabular}{@{} *{5}{c} @{}}
    \toprule
       \multicolumn{1}{c}{\multirow{2}{*}{Policies}} & \multicolumn{2}{c@{}}{$N_\test=500$} & \multicolumn{2}{c@{}}{$N_\test=2,000$} \\
       \cmidrule(l){2-3}  \cmidrule(l){4-5} 
         & Avg. Ratio (std) & Avg. Time (s) & Avg. Ratio (std) & Avg. Time (s)\\
        \midrule 
        GI & 88.1\% (5.1\%) & 0.04 & 84.5\% (6.0\%) & 0.21\\
        GILS & 100.8\% (2.5\%) & 11.31 & 101.5\% (2.3\%) & 93.84\\
        GIP & 100.3\% (3.4\%) & 2.78 & 103.9\% (3.3\%)& 7.17\\
        RO & 23.8\% (13.2\%) & 0.02 & 23.1\% (9.0\%) & 0.14\\
        LS & 93.7\% (6.0\%) & 3.55 & 91.6\% (2.4\%) & 11.23\\
        RP & 71.7\% (14.8\%) & 2.66 & 74.9\% (11.7\%) & 4.82\\
        ROLS & 100.0\% (0.0\%) & 15.68 & 100.0\% (0.0\%) & 91.55\\
        \bottomrule
    \end{tabular}
\end{table}

According to Table~\ref{tab:comparison_nl}, under the NL model, the proposed policies can still derive high-quality solutions within seconds, implying that the ``from-small-to-large'' idea still works in this case. 

\vspace{0.1in}

\tb{Exponomial model.} In this case, given $(K, M, N)$, each (training or testing) sample is generated as follows:
\begin{enumerate}
    \item To generate the proportions of different customer types, we first sample $K$ values $z_k$'s from $\mathcal{U}[0, 1]$, and then choose $\bs{\alpha}=(\alpha_1, \dots, \alpha_k)$ where $\alpha_k = z_k/\left(\sum_{k'=1}^K z_{k'}\right)$. 
    \item We generate revenues, utilities and distribution parameters as $r_{j} \sim \mathcal{U}[1, 2]$, $u_{kj} \sim \mathcal{U}[0, 1]$ and $\lambda_{kj}\sim\mathcal{U}[0.5, 1.5]$. We still set $\eta=3$. 
    \item Then, we generate two types of constraints:
    \begin{itemize}
        \item $\lfloor M/2\rfloor$ capacity constraints: For constraint $i$, we sample $A_{ij} \sim \mathcal{U}[0, 1]$ for any $j\in [N]$ and $ b_i \sim \mathcal{U}[5, 10]$.
        \item $\lceil M/2\rceil$ precedence constraints: For constraint $i$, we set $b_i=0$ and then sample $j_1\sim \mathcal{U_D}[N]$ and $j_2\sim \mathcal{U_D}[N]$. If $j_1\neq j_2$, we have $A_{ij_1}=1$, $A_{ij_2}=-1$, and $A_{ij'}=0$ for all $j'[N]\setminus\{j_1, j_2\}$; otherwise, $A_{ij}=0$ for all $j\in [N]$. 
    \end{itemize} 
    \item Lastly, for testing samples, we enumerate all feasible assortments to derive the optimal assortment $\bs{x}^*$, which serves as the label. 
\end{enumerate}

We first randomly generate 2,000 instances with $K_\test=10$ customer types, $N_\train=20$ products and $M_\train=10$ constraints, and then augment the dataset to 10,000 instances by Algorithm~\ref{alg:data_augment}.
The total sample generation time is 422 minutes. The training process is the same as that in Section~\ref{sec:numerical}, and we also run 10 independent trials. The average training loss is 0.364, the average validation loss is 0.351, and the average running time is 101 seconds. 

Then, we test the performance of the different policies on 100 instances with $K_\test=10$ customer types, $N_\test=500 \text{ or } 2,000$ products and $M_\test=10$ constraints. Similarly, we choose the ROLS policy as the baseline. The performance of different policies are summarized in Table~\ref{tab:comparison_exponomial}. 

\begin{table}[ht]
    \centering
    \caption{Performance of proposed policies and existing policies when $K_\train=K_\test=10$, $M_\train=M_\test=10$, $N_\train=20$ and $N_\test=500\mbox{ or } 2,000$; exponomial model.}
    \label{tab:comparison_exponomial} 
    \begin{tabular}{@{} *{5}{c} @{}}
    \toprule
       \multicolumn{1}{c}{\multirow{2}{*}{Policies}} & \multicolumn{2}{c@{}}{$N_\test=500$} & \multicolumn{2}{c@{}}{$N_\test=2,000$} \\
       \cmidrule(l){2-3}  \cmidrule(l){4-5} 
         & Avg. Ratio (std) & Avg. Time (s) & Avg. Ratio (std) & Avg. Time (s)\\
        \midrule 
        GI & 96.3\% (2.8\%) & 0.04 & 94.5\% (2.6\%) & 0.27\\
        GILS & 100.5\% (1.2\%) & 8.70 & 100.1\% (0.7\%) & 65.10\\
        GIP & 83.5\% (4.2\%) & 2.84 & 79.4\% (5.4\%) & 6.47\\
        RO & 9.9\% (0.8\%) & 0.02 & 8.8\% (0.4\%) & 0.18\\
        LS & 100.2\% (1.2\%) & 3.18 & 99.4\% (0.9\%) & 18.24\\
        RP & 44.5\% (8.2\%) & 2.67 & 44.0\% (8.7\%) & 4.99\\
        ROLS & 100.0\% (0.0\%) & 30.64 & 100.0\% (0.0\%) & 153.76\\
        \bottomrule
    \end{tabular}
\end{table}

According to Table~\ref{tab:comparison_exponomial}, under the exponomial model, the GI and the GILS policies derive high-quality solutions within seconds. Moreover, according to Section~\ref{subsec:time_limit}, we can choose an appropriate time limit to obtain high-quality solutions with a significantly shorter running time. 
The performance of the GIP policy is not excellent, but it still significantly outperforms the RP policy that shares the same policy structure with the GIP policy. 

\vspace{0.1in}

\tb{MC model.} In this case, given $(M, N)$, each (training or testing) sample is generated as follows:
\begin{enumerate}
    \item We generate revenues as $r_{j} \sim \mathcal{U}[1, 2]$.
    \item To generate the arrival probabilities $\bs{\beta}$, we first sample $N$ values $z_j$'s from $\mathcal{U}[0, 1]$, and then choose $\bs{\beta}=(\beta_1, \dots, \beta_N)$ where $\beta_j = z_j/\left(\sum_{j'=1}^N z_{j'}\right)$.
    \item To generate the transition matrix $\bs{\rho}$, for product $j$, we first sample $N$ values $z_{j\ell}$'s from $\mathcal{U}[0, 1]$, and set $z_{jj}=0$. Then, we choose $\bs{\rho}_{j, :}=(\rho_{j, 1}, \dots, \rho_{j, N})$ where $\rho_{j\ell} = 0.8\cdot z_{j \ell}/\left(\sum_{\ell'=1}^N z_{j\ell'}\right)$.
    \item Then, we generate two types of constraints:
    \begin{itemize}
        \item $\lfloor M/2\rfloor$ capacity constraints: For constraint $i$, we sample $A_{ij} \sim \mathcal{U}[0, 1]$ for any $j\in [N]$ and $ b_i \sim \mathcal{U}[5, 10]$.
        \item $\lceil M/2\rceil$ precedence constraints: For constraint $i$, we set $b_i=0$ and then sample $j_1\sim \mathcal{U_D}[N]$ and $j_2\sim \mathcal{U_D}[N]$. If $j_1\neq j_2$, we have $A_{ij_1}=1$, $A_{ij_2}=-1$, and $A_{ij'}=0$ for all $j'[N]\setminus\{j_1, j_2\}$; otherwise, $A_{ij}=0$ for all $j\in [N]$. 
    \end{itemize} 
    \item Lastly, we solve an integer linear program as follows:
    \begin{align*}
        \max_{\bs{x}, \bs{y}, \bs{z}} \quad & \sum_{j=1}^N y_j r_j \\
        \mbox{s.t.}\quad & y_j + z_j - \sum_{i=1}^N \rho_{ij} z_i = \beta_j & \forall j\in [N]\\
        & x_j \ge y_j & \forall j \in [N]\\
        & \bs{A} \bs{x} \le \bs{b}\\
        & x_j \in \{0, 1\}, \; y_j\ge 0, \; z_j\ge 0&\forall j\in [N],
    \end{align*}
    to derive the optimal assortment $\bs{x}^*$, which serves as the label. 
\end{enumerate}

For the MC model, we cannot use the augmentation approach in Algorithm~\ref{alg:data_augment}. Thus, in this case, we generate 2,000 solution samples with $M_\train=10$ constraints for each number of products ($N_\train =16, 17, 18, 19, 20$), resulting in a dataset with 10,000 samples. The total data generation time is 14 minutes.

The training process is the same as that in Section~\ref{sec:numerical}, and we also run 10 independent trials. The average training loss is 0.313, the average validation loss is 0.315, and the average running time is 120 seconds. Since computing the revenue of each assortment is time-consuming due to the matrix inversion operation, we only test the performance of the different policies on 100 instances with $N_\test=100$ products and $M_\test=10$ constraints. Similarly, we choose the ROLS policy as the baseline. The performance of different policies is summarized in Table~\ref{tab:comparison_mc}. 

\begin{table}[ht]
    \centering
    \caption{Performance of proposed policies and existing policies when $M_\train=M_\test=10$, $N_\train=20$ and $N_\test=100$; MC model.}
    \label{tab:comparison_mc} 
    \begin{tabular}{ccc}
    \toprule
         Policies & Avg. Ratio (std) & Avg. Time (s)\\
         \midrule 
        GI & 72.1\% (20.2\%) & 0.10 \\
        GILS & 100.5\% (4.2\%) & 88.55\\
        GIP & 102.5\% (3.8\%) & 0.84 \\
        RO & 64.3\% (18.0\%) & 0.06 \\
        LS & 95.3\% (0.8\%) & 32.21 \\
        RP & 103.6\% (3.0\%) & 1.37 \\
        ROLS & 100.0\% (0.0\%) & 83.46 \\
        \bottomrule
    \end{tabular}
\end{table}

According to Table~\ref{tab:comparison_mc}, under the MC model, the proposed policies can derive high-quality solutions within seconds. In order to further improve their performance, practitioners may enhance the GCN framework (see Remark~\ref{remark:improvement}). 

\section{Omitted Details in Section~\ref{subsec:model_free_num}}
\subsection{RMSE and MAPE}\label{append:def_rmse}
\begin{align*}
    \mathrm{RMSE}
    &= \sqrt{
        \frac{
            \sum_{\ell=1}^{L}
            \sum_{j \in S_\ell^+}
            \left[
                \Phi(S_\ell, \bs{I}_\ell, \bs{r}_\ell, j)
                - \hat{\Phi}(S_\ell, \bs{I}_\ell, \bs{r}_\ell, j)
            \right]^2
        }{
            \sum_{\ell=1}^{L} (|S_\ell| + 1)
        }
      }, \\[2mm]
    \mathrm{MAPE}
    &= \frac{1}{
        \sum_{\ell=1}^{L} (|S_\ell| + 1)
      }
      \sum_{\ell=1}^{L}
      \sum_{j \in S_\ell^+}
      \left|
        \frac{
            \Phi(S_\ell, \bs{I}_\ell, \bs{r}_\ell, j)
            - \hat{\Phi}(S_\ell, \bs{I}_\ell, \bs{r}_\ell, j)
        }{
            \Phi(S_\ell, \bs{I}_\ell, \bs{r}_\ell, j)
        }
      \right|.
\end{align*}

\subsection{Sample Generation}\label{append:generation_time}
We compute the optimality gaps of these approximated solutions relative 
to the true optimal solutions obtained under the known choice model.
\begin{table}[ht]
    \centering
    \begin{tabular}{cccc}
    \toprule
         $N_\train$ & \# of Samples & Optimality Ratio (std) & Avg. Generation Time (s)\\
         \midrule
         16 & 2,000 & 99.6\% (1.0\%) & 0.27\\
         17 & 2,000 & 99.5\% (1.0\%) & 0.32\\
         18 & 2,000 & 99.5\% (1.1\%) & 0.46\\
         19 & 2,000 & 99.4\% (1.1\%) & 0.66\\
         20 & 2,000 & 99.2\% (1.1\%) & 0.88\\
         \bottomrule
    \end{tabular}
    \caption{Optimality ratio and generation time of generated solution samples. }
    \label{tab:soltion_quality}
\end{table}

\section{Supplementary Algorithm Details}\label{appendix:conic}
In this section, we provide the local search algorithm in \cite{gallego2024efficient} and the conic integer programming algorithm in \cite{csen2018conic} for the constrained assortment problem under the MMNL model. 

\vspace{4mm}
\begin{breakablealgorithm}
    \caption{Local search policy}\label{alg:local_search}
    \small
    \begin{algorithmic}
    \State Input: Initial solution $\bs{x}$. By default, we have $\bs{x}=\bs{0}$. 
    \State Initialize: $\mathcal{T}(j)\gets 0$ for each $j\in [N]$.
    \While{$\mathcal{T}(j)<1$ for some $j\in [N]$}
        \State Set available product set $\mathcal{A}\gets \{j\in [N]\mid \mathcal{T}(j)<1\}$ and the assortment $S=\{j\in [N]\mid x_j=1\}$.
        \State Set $\mbox{ind}_A\gets \argmax_{j\in \mathcal{A}\cap\bar{S}} \mathcal{R}(\bs{x}+\bs{e}_j)$ and $\bs{x}_A\gets \bs{x}+\bs{e}_{\mbox{ind}_A}$. \Comment{Try addition.}
        \If{$\mathcal{R}(\bs{x}_A) \ge 1.001\times\mathcal{R}(\bs{x})$}
            \State Set $\bs{x}\gets \bs{x}_A$.
        \Else 
            \State Set $\mbox{ind}_D\gets \argmax_{j\in S} \mathcal{R}(\bs{x}-\bs{e}_j)$ and $\bs{x}_D \gets \bs{x}-\bs{e}_{\mbox{ind}_D}$; \Comment{Try deletion.}
            \State Set $(\mbox{ind}_X^1, \mbox{ind}_X^2) \gets \argmax_{(i, j)\in S\times (\bar{S}\cap \mathcal{A})} \mathcal{R}\left(\bs{x}-\bs{e}_{i}+\bs{e}_{j}\right)$ and $\bs{x}_X\gets \bs{x}-\bs{e}_{\mbox{ind}_X^1}+\bs{e}_{\mbox{ind}_X^2}$. \Comment{Try swap.}
            \If{$\max\{ \mathcal{R}(\bs{x}_D), \mathcal{R}(\bs{x}_X)\}\ge 1.001\times \mathcal{R}(\bs{x})$}
            \If{$\mathcal{R}(\bs{x}_D)\ge \mathcal{R}(\bs{x}_X)$}
            \State Set $\bs{x}\gets \bs{x}_D$ and $\mathcal{T}(\mbox{ind}_D)\gets \mathcal{T}(\mbox{ind}_D)+1$.
            \Else 
            \State Set $\bs{x}\gets \bs{x}_X$ and $\mathcal{T}(\mbox{ind}_X^1)\gets \mathcal{T}(\mbox{ind}_X^1)+1$.
            \EndIf
            \Else
            \State Break;
            \EndIf
        \EndIf
    \EndWhile
    \State Return: $\bs{x}$. 
    \end{algorithmic}
\end{breakablealgorithm}

\vspace{4mm}
\begin{breakablealgorithm}
    \caption{Revenue-order with local search (ROLS) policy}\label{alg:rols}
    \small
    \begin{algorithmic}
    \State Implement the index policy in Algorithm~\ref{alg:index_policy} by plugging in $r_j$'s. The output is denoted by $\bs{x}_1$. 
    \State Implement the LS policy in Algorithm~\ref{alg:local_search} by plugging in $\bs{x}_1$. The output is denoted by $\bs{x}_2$
    \State Return: $\bs{x}_2$. 
    \end{algorithmic}
\end{breakablealgorithm}

\vspace{4mm}
\begin{breakablealgorithm}
    \caption{Revenue-based integer programming (RP) policy}\label{alg:rp}
    \small
    \begin{algorithmic}
    \State Solve the following integer linear programming
    \begin{align*}
        \max_{\bs{x}\in \{0, 1\}^N} &\quad \sum_{j\in [N]} r_j x_j \\
        \mbox{s.t.} &\quad \bs{A} \bs{x} \le \bs{b},
    \end{align*}
    and obtain the optimal solution $\bs{x}^*$.
    \State Return: $\bs{x}^*$. 
    \end{algorithmic}
\end{breakablealgorithm}

\vspace{4mm}
\begin{breakablealgorithm}
    \caption{Conic integer programming algorithm in \cite{csen2018conic}}\label{alg:conic}
    \small
    \begin{algorithmic}
    \State Input: $\bs{\alpha}$, $\bs{r}$, $\bs{V}$, $\bs{A}$, $\bs{b}$, $K$, $M$, $N$.
    \State Use Algorithm~\ref{alg:mcCormick} to compute McCormick terms $f_{k\mid x_j=\xi}$'s for $k\in [K]$, $j\in [N]$ and $\xi\in \{0, 1\}$. 
    \State Set $\bar{r}\gets \max_{j\in [N]} \; r_j$.
    \State Solve the following conic integer programming:
    \begin{align*}
        \max_{\bs{x}, \bs{z}, \bs{y}} \quad & \bar{r} - \sum_{k\in [K]} \alpha_k \bar{r}y_k -\sum_{k\in [K]}\sum_{j\in [N]} \alpha_k v_{kj} (\bar{r} - r_j)z_{kj} \\
        \mbox{s.t.}\quad & \bs{A} \bs{x} \le \bs{b}\\
        & w_k = 1 + \sum_{j\in [N]} v_{kj} x_j & \forall k\in [K]\\
        & z_{kj} w_k \ge x_j^2 & \forall k\in [K],\; j\in [N]\\
        & y_k w_k\ge 1 & \forall k\in [K]\\
        & y_k + \sum_{j\in [N]} v_{kj}z_{kj} \ge 1 & \forall k\in [K]\\
        & z_{kj} \le \frac{x_j}{1+v_{kj}} & \forall k\in [K],\; j\in [N]\\
        & z_{kj} \ge \frac{x_j}{1+f_{k\mid x_j=1}} & \forall k\in [K],\; j\in [N]\\
        & z_{kj} \le y_k - \frac{1-x_j}{1+f_{k\mid x_j=0}} & \forall k\in [K],\; j\in [N]\\
        & z_{kj} \ge y_k - (1-x_j) & \forall k\in [K],\; j\in [N]\\
        & x_j\in \{0, 1\} & \forall j\in [N]\\
        & z_{kj}\ge 0 & \forall k\in [K],\; j\in [N]\\
        & y_k\ge 0 & \forall k\in [K]
    \end{align*}
    \State Return: $\bs{x}^*$. 
    \end{algorithmic}
\end{breakablealgorithm}
\vspace{4mm}

\vspace{4mm}
\begin{breakablealgorithm}
    \caption{McCormick estimators}\label{alg:mcCormick}
    \small
    \begin{algorithmic}
    \State Input: $\bs{V}$, $\bs{A}$, $\bs{b}$, $K$, $M$, $N$.
    \For{$k=1, 2, \dots, K$}
    \For{$j=1, 2, \dots, N$}
    \For{$\xi = 0, 1$}
    \State Solve the following linear program:
    \begin{align*}
        f_{k\mid x_j=\xi} := \max_{\bs{x}\in [0, 1]^N} \; \sum_{j\in [N]} v_{kj} x_j
        \qquad\mbox{s.t.} \;  \bs{A}\bs{x}\le \bs{b}, \quad x_j = \xi.
    \end{align*}
    \EndFor
    \EndFor
    \EndFor    
    \State Return: $f_{k\mid x_j=\xi}$'s.  
    \end{algorithmic}
\end{breakablealgorithm}
\vspace{4mm}

\section{Independent Trials in Section~\ref{sec:numerical}}\label{append:ind_trial}
In this section, we present the results under 10 independent trials as the random seed ranging from 1 to 10. 

\begin{table}[ht]
    \centering
    \caption{Summary of ten independent trials and the competitive ratios and standard deviations of the proposed policies.}
    \label{tab:ind_trial} 
    \resizebox{\textwidth}{!}{
    \begin{tabular}{@{} *{10}{c} @{}}
    \toprule
       \multicolumn{1}{c}{\multirow{2}{*}{Seed}} & \multicolumn{1}{c}{\multirow{2}{*}{Training Loss}} & \multicolumn{1}{c}{\multirow{2}{*}{Validation Loss}} & \multicolumn{1}{c}{\multirow{2}{*}{Training Time(s)}} & \multicolumn{3}{c@{}}{$N_\test=500$} & \multicolumn{3}{c@{}}{ $N_\test=2,000$} \\
       \cmidrule(l){5-7}  \cmidrule(l){8-10} 
        & & &  & GI & GILS & GIP & GI & GILS & GIP\\
        \midrule 
        1 & 0.340 & 0.345 & 126 & 88.0\% (5.9\%) & 94.8\% (1.9\%)& 94.7\% (2.1\%)& 87.1\% (3.9\%) & 93.6\% (1.7\%) & 94.7\% (1.7\%)\\
        2 & 0.348 & 0.346 & 126 & 86.9\% (4.5\%) & 94.7\% (1.7\%)& 96.4\% (1.3\%) & 82.8\% (7.2\%) & 93.3\% (2.5\%) & 96.6\% (1.2\%)\\
        3 & 0.416 & 0.416 & 75 & 86.3\% (4.6\%) & 95.0\% (1.7\%)& 97.7\% (1.0\%) & 84.4\% (5.1\%) & 94.2\% (1.7\%) & 97.9\% (0.8\%)\\
        4 & 0.350 & 0.350 & 126 & 87.0\% (6.2\%) & 94.6\% (2.0\%) & 94.5\% (1.9\%) & 86.4\% (6.1\%) & 93.3\% (2.1\%) & 94.3\% (1.6\%) \\
        5 & 0.372 & 0.381 & 122 & 88.1\% (4.2\%) & 94.6\% (2.3\%) & 96.7\% (1.2\%) & 87.1\% (4.6\%) & 93.7\% (2.1\%) & 96.8\% (1.1\%)\\
        6 & 0.415 & 0.405 & 73 & 88.7\% (4.3\%) & 96.4\% (1.2\%) & 97.8\% (0.9\%) & 86.5\% (4.4\%) & 95.2\% (1.6\%) & 97.9\% (0.9\%) \\
        7 & 0.363 & 0.408 & 114 & 89.9\% (3.5\%) & 96.3\% (1.5\%) & 96.5\% (1.2\%) & 90.0\% (2.9\%) & 95.3\% (1.4\%) & 96.1\% (1.2\%) \\
        8 & 0.345 & 0.346 & 126 & 87.4\% (6.9\%) & 94.7\% (2.2\%) & 95.0\% (1.9\%) & 86.3\% (5.0\%) & 93.4\% (2.0\%) & 94.6\% (1.8\%)\\
        9 & 0.352 & 0.346 & 126 & 86.9\% (4.6\%) & 94.2\% (2.0\%) & 95.8\% (1.7\%) & 86.5\% (3.6\%) & 93.3\% (1.8\%) & 95.8\% (1.7\%)\\
        10 & 0.363 & 0.364 & 126 & 88.9\% (4.9\%) & 95.6\% (1.4\%) & 95.7\% (1.4\%)& 87.0\% (7.5\%) & 94.1\% (2.1\%) & 95.9\% (1.2\%)\\
        \bottomrule
    \end{tabular}}
\end{table}

\section{Hash Functions in Algorithm~\ref{alg:WL_test}}\label{append:hash_function}
In this section, we formally define admissible hash functions in Algorithm~\ref{alg:WL_test}.
Fix a finite collection of graph inputs $\mathcal{D} \subseteq \mathcal{G}_{\mathrm{GCN}}$
and a finite depth $L \in \mathbb{N}$. For each $\bs{G} \in \mathcal{D}$, initialize the
colors by
\[
C^{S,0}_k(\bs{G})=Y^S_{k,:}(\bs{G}), \qquad
C^{P,0}_j(\bs{G})=Y^P_{j,:}(\bs{G}), \qquad
C^{C,0}_i(\bs{G})=Y^C_{i,:}(\bs{G}).
\]
A family of hash functions
\[
\left\{\mathrm{HASH}_{\tau,\ell}: \tau \in \{S,P,C\},\ \ell \in [L]\right\}
\quad \text{and} \quad
\left\{\mathrm{HASH}^{E}_{\rho,\ell}: \rho \in \{PS,SP,CP,PC\},\ \ell \in [L]\right\}
\]
is called admissible with respect to $(\mathcal{D},L)$ if the following conditions
hold recursively for every $\ell \in [L]$.

First, we define the realized edge-input domains
\begin{align*}
\mathcal{H}^{E}_{PS,\ell}
&=
\left\{
\left(\bs{Z}^S_{j,k,:}(\bs{G}), C^{P,\ell-1}_j(\bs{G})\right):
\bs{G} \in \mathcal{D},\ k \in \mathcal{N}^{S}(\bs{G}),\ j \in \mathcal{N}^{P}(\bs{G})
\right\},\\
\mathcal{H}^{E}_{SP,\ell}
&=
\left\{
\left(\bs{Z}^S_{j,k,:}(\bs{G}), C^{S,\ell-1}_k(\bs{G})\right):
\bs{G} \in \mathcal{D},\ k \in \mathcal{N}^{S}(\bs{G}),\ j \in \mathcal{N}^{P}(\bs{G})
\right\},\\
\mathcal{H}^{E}_{CP,\ell}
&=
\left\{
\left(\bs{Z}^C_{i,j,:}(\bs{G}), C^{C,\ell-1}_i(\bs{G})\right):
\bs{G} \in \mathcal{D},\ i \in \mathcal{N}^{C}(\bs{G}),\ j \in \mathcal{N}^{P}(\bs{G})
\right\},\\
\mathcal{H}^{E}_{PC,\ell}
&=
\left\{
\left(\bs{Z}^C_{i,j,:}(\bs{G}), C^{P,\ell-1}_j(\bs{G})\right):
\bs{G} \in \mathcal{D},\ i \in \mathcal{N}^{C}(\bs{G}),\ j \in \mathcal{N}^{P}(\bs{G})
\right\}.
\end{align*}
For each $\rho \in \{PS,SP,CP,PC\}$, the edge hash function
$\mathrm{HASH}^{E}_{\rho,\ell}$ is injective on $\mathcal{H}^{E}_{\rho,\ell}$.
Moreover, the following sets of edge colors are linearly independent:
\begin{align*}
\mathrm{HASH}^{E}_{PS,\ell}\left(\mathcal{H}^{E}_{PS,\ell}\right), \qquad \mathrm{HASH}^{E}_{SP,\ell}\left(\mathcal{H}^{E}_{SP,\ell}\right)
\cup
\mathrm{HASH}^{E}_{CP,\ell}\left(\mathcal{H}^{E}_{CP,\ell}\right), \qquad \mathrm{HASH}^{E}_{PC,\ell}\left(\mathcal{H}^{E}_{PC,\ell}\right).
\end{align*}
This condition ensures that the summation of edge colors uniquely represents the
corresponding finite multiset of incoming edge-color pairs.

Second, given the edge hash functions, define the aggregated neighborhood colors by
\begin{align*}
A^{S,\ell}_k(\bs{G})
&=
\sum_{j \in \mathcal{N}^{P}(\bs{G})}
\mathrm{HASH}^{E}_{PS,\ell}
\left(\bs{Z}^S_{j,k,:}(\bs{G}), C^{P,\ell-1}_j(\bs{G})\right),\\
A^{P,\ell}_j(\bs{G})
&=
\sum_{k \in \mathcal{N}^{S}(\bs{G})}
\mathrm{HASH}^{E}_{SP,\ell}
\left(\bs{Z}^S_{j,k,:}(\bs{G}), C^{S,\ell-1}_k(\bs{G})\right)
+
\sum_{i \in \mathcal{N}^{C}(\bs{G})}
\mathrm{HASH}^{E}_{CP,\ell}
\left(\bs{Z}^C_{i,j,:}(\bs{G}), C^{C,\ell-1}_i(\bs{G})\right),\\
A^{C,\ell}_i(\bs{G})
&=
\sum_{j \in \mathcal{N}^{P}(\bs{G})}
\mathrm{HASH}^{E}_{PC,\ell}
\left(\bs{Z}^C_{i,j,:}(\bs{G}), C^{P,\ell-1}_j(\bs{G})\right).
\end{align*}
For each node type $\tau \in \{S,P,C\}$, define the realized node-input domain
\[
\mathcal{H}_{\tau,\ell}
=
\left\{
\left(C^{\tau,\ell-1}_u(\bs{G}), A^{\tau,\ell}_u(\bs{G})\right):
\bs{G} \in \mathcal{D},\ u \in \mathcal{N}^{\tau}(\bs{G})
\right\}.
\]
The node hash function $\mathrm{HASH}_{\tau,\ell}$ is injective on
$\mathcal{H}_{\tau,\ell}$, and the updated colors are defined as
\[
C^{\tau,\ell}_u(\bs{G})
=
\mathrm{HASH}_{\tau,\ell}
\left(C^{\tau,\ell-1}_u(\bs{G}), A^{\tau,\ell}_u(\bs{G})\right),
\qquad
\tau \in \{S,P,C\}.
\]

\section{Proof of Theorem~\ref{thm:finite_represent}}
We first provide several useful results, and then prove the theorem. 

\subsection{WL–isomorphism of Tripartite Graph}
\begin{lemma}\label{lem:WL_iso}
    For any $\bs{G}, \bs{G}'\in \mathcal{G}_{\text{GCN}}^U$, $\bs{G}\sim \bs{G}'$ if and only if there exists a unique $\gamma\in \mathfrak{G}_S\times \mathfrak{G}_P\times \mathfrak{G}_C$ with $\bs{G}'=\gamma\cdot \bs{G}$. 
\end{lemma}

\noindent \textit{Proof.} First, if there exists $\gamma\in \mathfrak{G}_S\times \mathfrak{G}_P\times \mathfrak{G}_C$ with $\bs{G}'=\gamma\cdot \bs{G}$, due to the permutation-invariant property of the WL test in Algorithm~\ref{alg:WL_test}, we have $\text{WL}_{\text{MMNL}}(\bs{G}) = \text{WL}_{\text{MMNL}}(\bs{G}')$ for any choice of hash functions and any depth $L$, implying that $\bs{G}\sim \bs{G}'$. 

For the other direction, we fix $\bs{G}, \bs{G}'\in \mathcal{G}_{\text{GCN}}^U$ with $\bs{G}\sim \bs{G}'$ and prove the statement in five steps. 

\vspace{0.1in}
\noindent \textbf{Step 1: Joint discreteness.} By Definition~\ref{def:unfoldable}, for each of $\bs{G}$ and $\bs{G}'$ separately, there exist an admissible choice of hash functions and a depth $L_G, L_{G'}\in \mathbb{N}$ such that the WL coloring is discrete on all three node types of the respective graph. We build a joint admissible encoder valid for both graphs layer by layer simultaneously: Let $\mathcal{Q}_l$ denote the (finite) set of [edge feature, color] pairs appearing in $\bs{G}$ or $\bs{G}'$ at iteration $l$. Consider a real linear space $V_l$ of dimension greater than $|\mathcal{Q}_l|$ and choose hash functions that assign each element of $\mathcal{Q}_l$ to a basis of $V_l$. This results in admissible edge hash functions, $\text{HASH}^E_{\rho, l}$'s, for both $\bs{G}$ and $\bs{G}'$. Similarly, we choose node hash functions, $\text{HASH}_{\tau, l}$'s, that are injective on the union of inputs across both graphs. Set the depth as $L^*=\max\{L_G, L_{G'}\}+1$. Then, at iteration $L^*$, both $\bs{G}$ and $\bs{G}'$ have discrete colorings on every node type. 

\vspace{0.1in}
\noindent \textbf{Step 2: Color-matching bijections.} Given the constructed hash functions and depth, each color appears in each multiset only once. Since $\bs{G}\sim \bs{G}'$, we have $\text{WL}_{\text{MMNL}}(\bs{G})=\text{WL}_{\text{MMNL}}(\bs{G}')$, which implies the existence of unique bijections $\pi_S\in \mathfrak{G}_S$, $\pi_P\in \mathfrak{G}_P$ and $\pi_C\in \mathfrak{G}_C$ such that 
\begin{align}\label{eq:color-matching}
    C_{\pi_S(k)}'^{S, L^*} = C_{k}^{S, L^*}, \qquad C_{\pi_P(j)}'^{P, L^*} = C_{j}^{P, L^*}, \qquad 
    C_{\pi_C(i)}'^{C, L^*} = C_{i}^{C, L^*},
\end{align}
for all $k\in [K]$, $j\in [N]$, $i\in [M]$. 

\vspace{0.1in}
\noindent \textbf{Step 3: Backward propagation of color-matching.} We prove by induction that for $l\in \{0\}\cup [L^*]$: 
\begin{align}\label{eq:color-matching_l}
    C_{\pi_S(k)}'^{S, l} = C_{k}^{S, l}, \qquad C_{\pi_P(j)}'^{P, l} = C_{j}^{P, l}, \qquad 
    C_{\pi_C(i)}'^{C, l} = C_{i}^{C, l},
\end{align}
for all $k\in [K]$, $j\in [N]$, $i\in [M]$. 

When $l=L^*$, the statement holds due to \eqref{eq:color-matching}. Assume that \eqref{eq:color-matching_l} holds for $l+1$. We prove the statement for $l$. By Algorithm~\ref{alg:WL_test}, we have 
\begin{align*}
    &C_k^{S, l+1}=\text{HASH}_{S, l+1}\left(C_k^{S, l},\; \sum_{j=1}^N \text{HASH}^E_{PS, l}\left(\bs{Z}_{j, k, :}^S, C_j^{P, l}\right) \right),\\
    &C_{\pi_S(k)}'^{S, l+1}=\text{HASH}_{S, l+1}\left(C_{\pi_S(k)}'^{S, l},\; \sum_{j=1}^N \text{HASH}^E_{PS, l}\left(\bs{Z}_{j, {\pi_S(k)}, :}'^S, C_j'^{P, l}\right) \right).
\end{align*}
By the inductive hypothesis, we have $C_k^{S, l+1}=C_{\pi_S(k)}'^{S, l+1}$. Since $\text{HASH}_{S, l+1}$ is injective, we have 
\begin{align}\label{eq:hash_res}
    C_k^{S, l} = C_{\pi_S(k)}'^{S, l}, \qquad \sum_{j=1}^N \text{HASH}^E_{PS, l}\left(\bs{Z}_{j, k, :}^S, C_j^{P, l}\right)=\sum_{j=1}^N \text{HASH}^E_{PS, l}\left(\bs{Z}_{j, {\pi_S(k)}, :}'^S, C_j'^{P, l}\right),
\end{align}
where the first equation establishes the statement for customer nodes at iteration $l$. Similarly, we can prove the statement for product and constraint nodes. Therefore, the statement holds for $l$, implying that \eqref{eq:color-matching_l} holds for $l\in \{0\}\cup [L^*]$. 

\vspace{0.1in}
\noindent \textbf{Step 4: Matching node features.} By the statement in Step 3, we have for $l=0$, $C_{\pi_\tau(v)}'^{\tau, 0}=C_{v}^{\tau, 0}$ for $\tau\in \{S, P, C\}$. The injectivity of $\text{HASH}_{\tau, 0}$'s implies that 
\begin{align*}
    \bs{Y}_{k, :}^S = \bs{Y}_{\pi_S(k), :}'^S, \qquad \bs{Y}_{j, :}^P = \bs{Y}_{\pi_P(j), :}'^P, \qquad 
    \bs{Y}_{i, :}^C = \bs{Y}_{\pi_C(i), :}'^C,
\end{align*}
for all $k\in [K]$, $j\in [N]$, $i\in [M]$. 

\vspace{0.1in}
\noindent \textbf{Step 5: Matching edge features.} Consider $l=L^*-1$. By \eqref{eq:hash_res}, we have 
\begin{align*}
    \sum_{j=1}^N \text{HASH}^E_{PS, L^*-1}\left(\bs{Z}_{j, k, :}^S, C_j^{P, L^*-1}\right)=\sum_{j=1}^N \text{HASH}^E_{PS, L^*-1}\left(\bs{Z}_{j, {\pi_S(k)}, :}'^S, C_j'^{P, L^*-1}\right).
\end{align*}
Reindex the right-hand side by $j\to \pi_P(j)$ and substitute $C_{\pi_P(j)}'^{P, L^*-1}=C_{j}^{P, L^*-1}$:
\begin{align*}
    \sum_{j=1}^N \text{HASH}^E_{PS, L^*-1}\left(\bs{Z}_{j, k, :}^S, C_j^{P, L^*-1}\right)=\sum_{j=1}^N \text{HASH}^E_{PS, L^*-1}\left(\bs{Z}_{\pi_P(j), {\pi_S(k)}, :}'^S, C_j^{P, L^*-1}\right).
\end{align*}

Since the constructed edge hash functions map each edge-color pair to a basis, we can deduce that 
\begin{align*}
    \bs{Z}_{j, k, :}^S = \bs{Z}_{\pi_P(j), {\pi_S(k)}, :}'^S, \qquad \forall k\in [K], j\in [N].
\end{align*}
We can prove similar results for $\bs{Z}^C$ and $\bs{Z}'^C$. 

\vspace{0.1in}
\noindent \textbf{Conclusion.} Based on Steps 4 and 5, we can deduce that for $\bs{G}, \bs{G}'\in \mathcal{G}_{\text{GCN}}^U$ with $\bs{G}\sim \bs{G}'$, there exists $\gamma\in \mathfrak{G}_S\times \mathfrak{G}_P\times \mathfrak{G}_C$ with $\bs{G}'=\gamma\cdot \bs{G}$. Combining with the other direction, we finish the proof of this lemma. \hfill \qed

\subsection{Universal Approximation of $\mathcal{F}_{\text{GCN}}$}
\begin{lemma}\label{lem:universal_approximation}
    Let $\mathcal{D} \subseteq \mathcal{G}_{\text{GCN}}^U$ be compact and closed under $\Gamma$. Suppose $\Psi:\mathcal{D}\to \mathbb{R}^N$ is continuous and satisfies: 
    \begin{enumerate}
        \item $\Psi(\bs{G})_j=\Psi(\gamma\cdot \bs{G})_{\pi_P(j)}$ for all $\gamma\in \Gamma$, $\bs{G}\in \mathcal{D}$, $j\in [N]$;
        \item if $\bs{G}\sim \bs{G}'$ for $\bs{G}, \bs{G}'\in \mathcal{D}$, then for the unique $\gamma\in \Gamma$ realizing $\bs{G}'=\gamma \cdot \bs{G}$ (Lemma~\ref{lem:WL_iso}), $\Psi(\bs{G})_j = \Psi(\bs{G}')_{\pi_P(j)}$;
        \item if product nodes $j\neq j'$ in $\bs{G}\in \mathcal{D}$ satisfy $C_j^{P, L}=C_{j'}^{P, L}$ for any choice of hash functions and any depth $L$, then $\Psi(\bs{G})_j=\Psi(\bs{G})_{j'}$.
    \end{enumerate}
    Then for any $\eta>0$, there exists $\Phi\in \mathcal{F}_{\text{GCN}}$ with $\sup_{\bs{G}\in \mathcal{D}} \Vert\Phi(\bs{G}) - \Psi(\bs{G}) \Vert_\infty < \eta$.
\end{lemma}
The proof is similar to Theorem~E.1 in \cite{chen2023gnn-lp}. 

\subsection{Canonical Optimal Assortment}\label{append:canonical_assort}
Given any MMNL instance, the feasible set is not empty because $\bs{0}$ is always a feasible solution. 
Similar to \cite{chen2023on}, the optimal assortment may not be unique and hence we need to define the canonical optimal assortment. According to the WL test, given an unfoldable graph, we can derive a unique color for each node. Then, similar to Appendix~C in \cite{chen2023on}, we can generate a unique ranking of product nodes, i.e., $\pi_{\bs{G}}:[N]\to [N]$. 

\begin{definition}[canonical optimal assortment]
    $\bs{x}^*(\bs{G})$ is the canonical optimal assortment if $\bs{x}^*$ satisfies the linear constraints and $(x_{\pi_{\bs{G}}(1)}, x_{\pi_{\bs{G}}(2)}, \dots, x_{\pi_{\bs{G}}(N)})$ is minimal in the sense of lexicographical order. 
\end{definition}

Then we define $\Phi_{\text{assort}}$ as the mapping from $\bs{G}$ to $\bs{x}^*(\bs{G})$. Since $\bs{x}^*(\bs{G})$ is unique, the mapping $\Phi_{\text{assort}}$ is well-defined. Moreover, due to the permutation-invariant property of $\pi_{\bs{G}}$, we have $\Phi_{\text{assort}}(\bs{G})_j=\Phi_{\text{assort}}(\gamma\cdot \bs{G})_{\pi_P(j)}$ for any $\gamma\in \Gamma$. 

\subsection{Proof of Theorem~\ref{thm:finite_represent}}
Define $\mathcal{X}:=\{\gamma\cdot \bs{G}:\gamma\in \Gamma, \bs{G}\in \mathcal{D}\}$. $\mathcal{X}$ is finite and $\Gamma$-closed by construction. Since the unfoldability property is $\Gamma$-invariant, $\mathcal{X}\subseteq \mathcal{G}_{\text{GCN}}^U$. Under the subspace topology inherited from $\mathcal{G}_{\text{GCN}}^U$, it is discrete and compact, so every function on $\mathcal{X}$ is continuous. 

Given any $\epsilon\in (0, 1/2)$, we define the function
\begin{align*}
    \Psi_\epsilon(\bs{G})_j = \begin{cases}
        1+\log(1/\epsilon-1) & \Phi_{\text{assort}}(\bs{G})_j=1,\\
        -1-\log(1/\epsilon-1) & \Phi_{\text{assort}}(\bs{G})_j=0,
    \end{cases}
    \qquad \forall\; \bs{G}\in \mathcal{X}, j\in [N],
\end{align*}
which satisfies the conditions in Lemma~\ref{lem:universal_approximation}. Then, letting $\eta=1$, by Lemma~\ref{lem:universal_approximation}, we have $\sup_{\bs{G}\in \mathcal{X}} \Vert \Phi(\bs{G})-\Psi_\epsilon(\bs{G})\Vert_\infty<1$. Thus, for every $\bs{G}\in \mathcal{X}$ and $j\in [N]$, there exists $\Phi\in \mathcal{F}_{\text{GCN}}$ such that 
\begin{align*}
    \Phi(\bs{G})_j>\log(1/\epsilon-1) \; \text{if $\Phi_{\text{assort}}(\bs{G})_j=1$}, \qquad \Phi(\bs{G})_j<-\log(1/\epsilon-1) \; \text{if $\Phi_{\text{assort}}(\bs{G})_j=0$}.
\end{align*}
Therefore, for the corresponding $\Phi^\sigma$, we have 
\begin{align*}
    \Phi^\sigma(\bs{G})_j>1-\epsilon \quad \text{if $\Phi_{\text{assort}}(\bs{G})_j=1$}, \qquad \Phi^\sigma(\bs{G})_j<\epsilon \quad \text{if $\Phi_{\text{assort}}(\bs{G})_j=0$},
\end{align*}
which finishes the proof. \hfill \qed

\section{Proof of Theorem~\ref{thm:mnl_prob_learning}}
Since $K=1$ and $M=0$, we only have one customer node and $N$ product nodes. Let $S$ denote the number of instances and $\bar{N}:=\max_{i} N_i$ denote the maximal number of products among all instances. It is easy to check that $\mathcal{R}^*(\bs{G})$ is permutation-invariant. According to \cite{zaheer2017deep}, we have the following lemma. 
\begin{lemma}\label{lem:set_approximation}
    There exist functions $\zeta: \mathbb{R}^2\to \mathbb{R}^{S\bar{N}}$ and $\rho: \mathbb{R}^{S\bar{N}}\to \mathbb{R}$, such that for any instance $\bs{G}\in \mathcal{D}$, 
    \[
    \mathcal{R}^*(\bs{G}) = \rho\left(\sum_{i=1}^N \zeta(r_i, v_i)\right).
    \]
\end{lemma}

We choose the hidden layer dimension $d_{\text{mid}}=S\cdot\bar{N}$. For the first layer, we construct the following functions:
\begin{align*}
    &f_{PS}^{(1)} \left([x_1, x_2], [y_1, y_2, y_3]\right):= \zeta(y_1, x_1) \in \mathbb{R}^{1\times d_{\text{mid}}}, \qquad g_S^{(1)}([x_1, x_2, x_3], \bs{y}) := [\rho\left(\bs{y}\right), 0, 0, \dots, 0]\in \mathbb{R}^{1\times d_{\text{mid}}},\\
    &f_{SP}^{(1)} \left([x_1, x_2], [y_1, y_2, y_3]\right):= [0, 0, \dots, 0]\in \mathbb{R}^{1\times d_{\text{mid}}}, \qquad g_P^{(1)}([x_1, x_2, x_3], \bs{y}) = [x_1, 0, \dots, 0]\in \mathbb{R}^{1\times d_{\text{mid}}},
\end{align*}
where $\zeta(\cdot)$ and $\rho(\cdot)$ are the functions in Lemma~\ref{lem:set_approximation}.
Then, we have 
\begin{align*}
    \bs{Y}^{S, 1}_1 = \left[\rho\left(\sum_{j=1}^N  \zeta(r_j, v_j)\right), 0, \dots, 0\right]\in \mathbb{R}^{1\times d_{\text{mid}}}, \qquad \bs{Y}_j^{P, 1} = [r_j, 0, \dots, 0] \in \mathbb{R}^{1\times d_{\text{mid}}}\quad \forall j\in [N].
\end{align*}

For the second layer, we construct the following functions: 
\begin{align*}
    f_{SP}^{(2)}\left([x_1, x_2], \bs{y}\right):=\bs{y}\in \mathbb{R}^{1\times d_{\text{mid}}}, \qquad g_{P}^{(2)}\left(\bs{x}, \bs{y}\right) = [\beta\cdot(x_1-y_1), 0, \dots, 0]\in \mathbb{R}^{1\times d_{\text{mid}}}.
\end{align*}

Then, we have 
\begin{align*}
    \bs{Y}^{P, 2}_j = \left[\beta\cdot \left(r_j - \rho\left(\sum_{j=1}^N  \zeta(r_j, v_j)\right)\right), 0, \dots, 0\right]\in \mathbb{R}^{1\times d_{\text{mid}}},
\end{align*}
Letting $f_{\text{out}}(\bs{x})=x_1$, by Lemma~\ref{lem:set_approximation}, we have
\begin{align*}
    p_j = \sigma\left(\beta\cdot (r_j-\mathcal{R}^*(\bs{G})\right),
\end{align*}

If $\Phi_{assort}(\bs{G})_j=1$, then $r_j> \mathcal{R}^*(\bs{G})$ and hence
\begin{align*}
    p_j = \sigma\left(\beta \cdot (r_j - \mathcal{R}^*(\bs{G}))\right) \ge \sigma\left( \beta\delta\right) = 1-\epsilon. 
\end{align*}

If $\Phi_{assort}(\bs{G})_j=0$, then $r_j<\mathcal{R}^*(\bs{G})$ and hence 
\begin{align*}
    p_j = \sigma\left(\beta \cdot (r_j - \mathcal{R}^*(\bs{G}))\right) \le \sigma\left( -\beta\delta\right) = \epsilon.
\end{align*}

Therefore, for any $\bs{G}\in \mathcal{D}$ and every $j\in [N]$, we have 
\begin{align*}
    |\Phi^\sigma_{GCN}(\bs{G})_j-\Phi_{assort}(\bs{G})_j|\le \epsilon. 
\end{align*}

As $\beta\to \infty$, we can deduce that $|\Phi^\sigma_{GCN}(\bs{G})_j-\Phi_{assort}(\bs{G})_j|\to 0$ and hence the training loss tends to 0. 
\hfill \qed

\end{appendices}

% Acknowledgments here
% \ACKNOWLEDGMENT{}	
		% CASE 1: BiBTeX used to constantly update the references
		%   (while the paper is being written).
		%\bibliographystyle{ormsv080} % outcomment this and next line in Case 1
		%\bibliography{<your bib file(s)>} % if more than one, comma separated
		
		% CASE 2: BiBTeX used to generate mypaper.bbl (to be further fine tuned)
		%\input{mypaper.bbl} % outcomment this line in Case 2
		
		%If you don't use BiBTex, you can manually itemize references as shown below.

%\newpage

		%----------------------------------------------------------------------------------------
		
	\end{document}